\documentclass{article}

\PassOptionsToPackage{round,sort&compress}{natbib}

\usepackage[final]{neurips_2021}

\usepackage[utf8]{inputenc} 
\usepackage[T1]{fontenc}    
\usepackage{hyperref}       
\usepackage{url}            
\usepackage{booktabs}       
\usepackage{amsfonts}       
\usepackage{nicefrac}       
\usepackage{microtype}      
\usepackage{xcolor}         
\usepackage{graphicx}
\usepackage{amsmath}
\usepackage{amsfonts}
\usepackage{multirow}
\usepackage{amsfonts}
\usepackage{tikz}
\usetikzlibrary{calc}
\usepackage{tikz-qtree}
\usetikzlibrary{trees}
\usepackage[edges]{forest}
\usetikzlibrary{shadows,arrows.meta}
\usepackage{makecell}
\usepackage{color}
\usepackage{CJKutf8}
\usepackage{rotating}

\title{Faithfulness in Natural Language Generation: A Systematic Survey of Analysis, Evaluation and Optimization Methods}

\author{%
  Wei Li\textsuperscript{1}, Wenhao Wu\textsuperscript{2}, Moye Chen\textsuperscript{1}, Jiachen Liu\textsuperscript{1}, Xinyan Xiao\textsuperscript{1}, Hua Wu\textsuperscript{1}\\
  \textsuperscript{1}Baidu Inc., Beijing, China\\
  \textsuperscript{2}Key Laboratory of Computational Linguistics, MOE, Peking University \\
  \texttt{\{liwei85,wuwenhao,chenmoye,liujiachen,xiaoxinyan,wu\_hua\}@baidu.com} \\
}

\begin{document}

\maketitle

\begin{abstract}
  Natural Language Generation (NLG) has made great progress in recent years due to the development of deep learning techniques such as pre-trained language models. 
  This advancement has resulted in more fluent, coherent and even properties controllable (e.g. stylistic, sentiment, length etc.) generation, naturally leading to development in downstream tasks such as abstractive summarization, dialogue generation, machine translation, and data-to-text generation.
  However, the faithfulness problem that the generated text usually contains unfaithful or non-factual information has become the biggest challenge, which makes the performance of text generation unsatisfactory for practical applications in many real-world scenarios.
  Many studies on analysis, evaluation, and optimization methods for faithfulness problems have been proposed for various tasks, but have not been organized, compared and discussed in a combined manner.
  In this survey, we provide a systematic overview of the research progress on the faithfulness problem of NLG, including problem analysis, evaluation metrics and optimization methods.
  We organize the evaluation and optimization methods for different tasks into a unified taxonomy to facilitate comparison and learning across tasks.
  Several research trends are discussed further.
\end{abstract}

\begin{figure}[h!]
	\centering
	\includegraphics[width=3.5in]{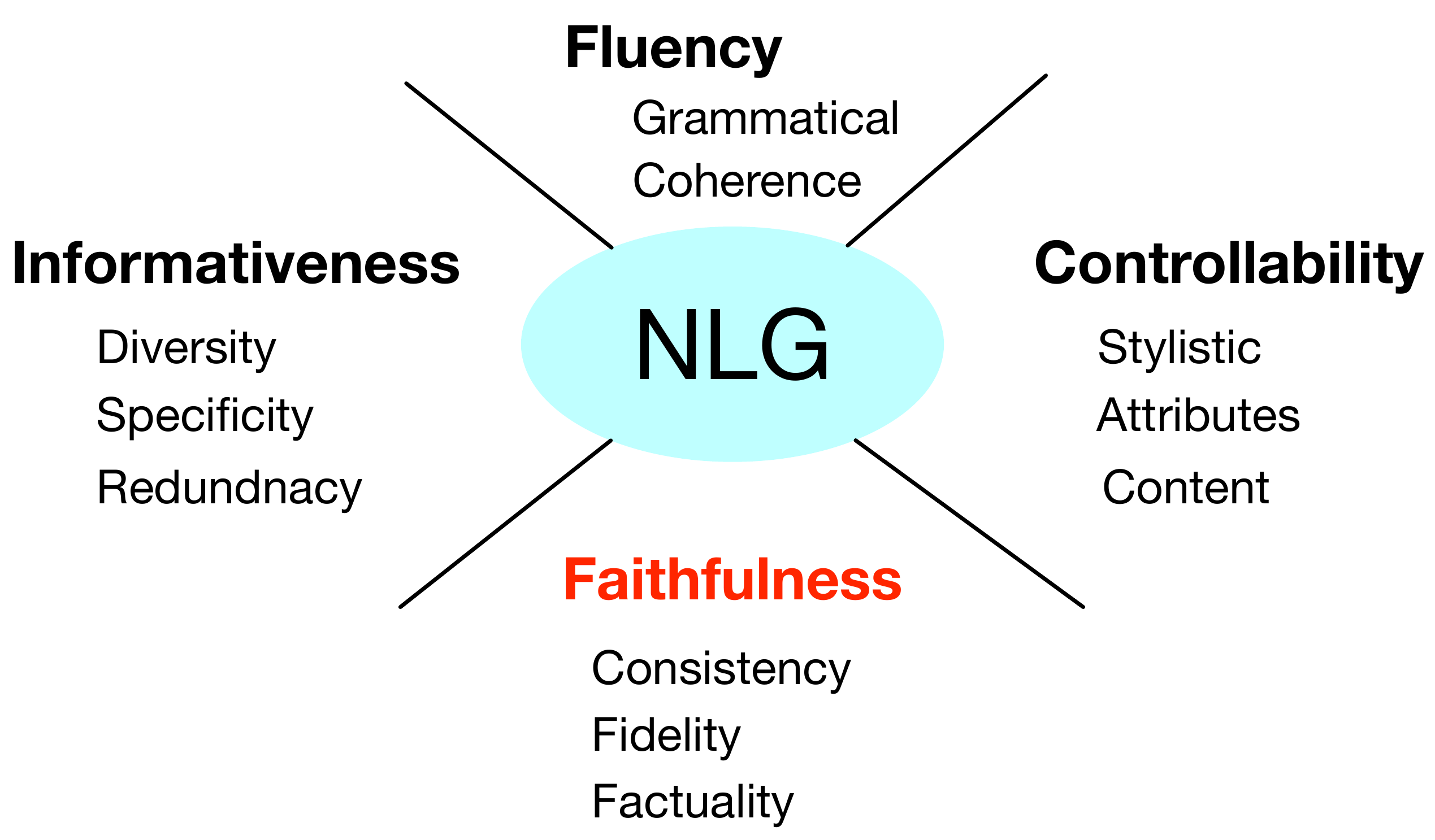}
	\caption{Four aspects of the NLG challenge. Faithfulness has become the biggest challenge in modern natural language generation.}
	\label{nlg_issues}
\end{figure}

\newpage
\tableofcontents
\newpage

\section{Introduction}

Natural Language Generation (NLG) is the process of producing a natural language text from a textual or non-textual input in order to meet specified communicative goals \citep{gatt2018survey}. The input of NLG varies with different task settings, however, the output is always readable natural language text. According to the type of input, the tasks of NLG can be mainly categorized into: text-to-text generation, data-to-text generation, and multimodality-to-text generation.

The text-to-text generation tasks take existing texts as input, and automatically produce a new, coherent text as output. The most common applications include: text summarization \citep{allahyari2017text}, dialogue generation \citep{li2016deep}, machine translation \citep{koehn2009statistical}, question generation \citep{du2017learning}, paraphrase generation \citep{li2017paraphrase} etc.
The data-to-text generation tasks automatically generate text from numerical or structured data such as table, key-value lists, and tuples. The example applications include: table-to-text generation \citep{liu2018table}, KG-to-text generation \citep{ke2021jointgt}, meaning-to-text generation (e.g. AMR-to-text) \citep{song2018graph} etc.
The multimodality-to-text generation tasks transfer the semantics in multimodal input such as images or videos, into natural language texts. Typical tasks include image caption \citep{vinyals2015show}, visual storytelling \citep{huang2016visual}, and video summarization \citep{ma2002user}.

From the perspective of input-output information transformation, the tasks of NLG can be divided into open-ended language generation and non-open-ended language generation. Open-ended language generation tasks refer to tasks that the input is incomplete and the output semantics are not contained by the input. For example, story generation is a classical open-ended language generation task, which tends to generate a complete story based on some leading sentences or keywords. Obviously, the model needs to create new information to completing storyline planning and generating meaningful stories. One of the greatest characteristics of the open-ended language generation tasks is that the information mapping between input and output is usually one-to-many. The same input can produce many outputs with different meanings.

By contrast, for non-open-ended language generation tasks, the input usually provides complete or even more information for the output. Machine translation is one typical non-open-ended language generation task where the input provides complete semantics for the output. Paraphrase generation can be regarded as an equivalent transformation of information, where the input and output semantics are exactly the same, but the language expression is different. In text summarization, input usually provides more information than output, so the summarization model needs to select salient information to produce summary output.

Table~\ref{task_list} lists some common NLG tasks as well as their characteristics.

\begin{table}[h!]
  \caption{Categories of common natural language generation tasks.}
  \label{task_list}
  \centering
  \begin{tabular}{lll}
    \toprule
    Tasks     & Category     & Information Mapping \\
    \hline
    \midrule
    Text Summarization & Text-to-Text  & Non-open-ended     \\
    Machine Translation     & Text-to-Text & Non-open-ended      \\
    Sentence Simplication     & Text-to-Text       & Non-open-ended  \\
    Paraphrase Generation & Text-to-Text  & Non-open-ended     \\
    Dialogue Generation     & Text-to-Text & Open-ended      \\
    Question Generation     & Text-to-Text       & Non-open-ended  \\
    Story Generation & Text-to-Text  & Open-ended     \\
    Essay Generation     & Text-to-Text & Open-ended      \\
    News Generation     & Text-to-Text       & Open-ended  \\
    Poetry Generation     & Text-to-Text       & Open-ended  \\
    Table-to-Text Generation & Data-to-Text  & Non-open-ended     \\
    AMR-to-Text Generation     & Data-to-Text & Non-open-ended      \\
    Image Caption     & Multimodality-to-Text       & Non-open-ended  \\
    Video Caption     & Multimodality-to-Text       & Non-open-ended  \\
    Visual Storytelling     & Multimodality-to-Text       & Open-ended  \\
    \bottomrule
  \end{tabular}
\end{table}

\subsection{Developing of NLG}

\begin{table}
  \renewcommand\arraystretch{3}
  \caption{Four paradigms in natural language generation.}
  \label{nlg_develop}
  \centering
  \begin{tabular}{p{3cm}p{6cm}p{4cm}}
    \toprule
    \textbf{Paradigm}     & \textbf{Engineering}     &  \textbf{Main Problems} \\
    \midrule
    \hline
    Template-based  & \makecell[l]{Manual Rules \\(Content Planning, Sentence Planning, \\Text Realization)}  & Fluency, Informativeness     \\
    \hline
    Statistical-based  &  \makecell[l]{Statistic Language Model \\(e.g. N-gram, Smoothing, Perplexity)}  &  Fluency, Informativeness \\
    \hline
    Neural-based  &  \makecell[l]{Neural Architecture \\(e.g. RNN, LSTM, CNN, Transformer)}  &  Controllability, Faithfulness \\
    \hline
    Pretraining-based  & \makecell[l]{Pretraining Objectives \\ (e.g. BERT, T5, GPT3, BART, CTRL)}  &  Faithfulness  \\
    \bottomrule
  \end{tabular}
\end{table}

The research on NLG has a long history, starting from 1950s. The developing of NLG approaches can be mainly divided into four stages: template-based, statistical-based, neural-based and pretraining-based, as shown in Table~\ref{nlg_develop}.
\begin{itemize}
    \item \textbf{Template-based}. The earliest natural language generation system adopted the method of rules and templates to design different modules for text generation, which reflected the linguistic knowledge of vocabulary, grammar, syntax and even pragmatics designed by many experts. They usually consists of several different components including content planning, sentence planning and text realization, each performing a specified function.
    \item \textbf{Statistical-based}. Statistical language model further proposes a new idea of language modeling from the perspective of probability and statistics, which encodes the dependency between vocabulary and context in conditional probability. N-gram language model is the most popular statistical language model, which is usually coupled with template-based methods for re-ranking and selecting fluent generated texts.
    \item \textbf{Neural-based}. With the development of deep learning, the neural-based end-to-end methods have gradually occupied a dominant position, which can better model the statistical co-occurrence relationship between vocabulary and context through end-to-end training, thus significantly improves the performance of text generation. Various neural architectures have been explored for NLG, such as Recurrent Neural Network (RNN) \citep{graves2013generating, zaremba2014recurrent}, Convolutional Neural Network (CNN) \citep{kalchbrenner2014convolutional} and self-attention Transformer network \citep{vaswani2017attention}.
    \item \textbf{Pretraining-based}. Most recently,  the pre-trained language generation model based on the Transformer architecture can better capture the linguistic knowledge of vocabulary, syntax and semantics, which greatly promotes the development of natural language generation. The rise of pre-trained language models \citep{devlin2018bert,liu2019roberta,brown2020language} has led to strong text generation models for applications including text summarization \citep{zhang2020pegasus,liu2019text,dong2019unified}, dialogue generation \citep{zhang2019dialogpt,DBLP:conf/acl/BaoHWWW20}, data-to-text generation \citep{chen2020kgpt}, and machine translation \citep{liu2020multilingual}. However, while these models generate fluent and grammatical text, they are prone to making factual errors that contradict the input text \citep{cao_faithful_2017}.
\end{itemize}

\paragraph{Challenges and Issues}
NLG faces four main challenges and issues at different development stages:  fluency, informativeness, controllability and faithfulness, as shown in Figure~\ref{nlg_issues}.
\begin{itemize}
    \item The \textbf{fluency} problem refers to whether the generated text is fluent, grammatical, and coherent.
    \item The \textbf{informativeness} problem refers to that the model generates redundant, meaningless, and general content, and the generated text is significantly insufficient in informativeness, diversity, and specificity.
    \item The \textbf{controllability} problem means that the generated text cannot satisfy the pre-given constraints, such as text style, attributes and content.
    \item The \textbf{faithfulness} problem means that the generated content is inconsistent with the input information, has hallucinations or non-factual information.
\end{itemize}

Traditional template-based methods can usually generate reliable and faithful texts, but limited by the diversity and generality of rules, the generated texts usually face the problems of fluency and informativeness. Benefiting from end-to-end training on large corpus, the neural-based methods can generate fluent and informative texts. However, due to the introduction of the probability sampling mechanism, they need to sample from the probability distribution estimated by the model each time. Considering that the vocabulary is very large, generally in the order of $1000\sim 50000$, the probability distribution inevitably contains a large number of long-tail words with low probability of occurrence, coupled with the randomness of probability sampling itself, the controllability and faithfulness of the neural-based NLG model is particularly serious. In the pre-training era, through self-supervised training on large-scale unlabeled corpora, the model generated text is outstanding in terms of fluency, informativeness and even controllability, but it still cannot solve the faithfulness problem.

\subsection{The Faithfulness Problem}
The faithfulness problem has become the biggest challenge in NLG, which largely limits the applicability of NLG algorithms in practical scenarios. 
For example, the researches on abstractive text summarization show that about 30\% of summaries generated by state-of-the-art models have faithfulness issues \citep{cao_faithful_2017,pagnoni_understanding_2021,kryscinski_evaluating_2019,falke_ranking_2019}. This brings serious problems to the credibility and usability of abstractive summarization systems.

\begin{table}[h!]
  \renewcommand\arraystretch{2}
  \caption{Examples of unfaithful errors for several common NLG tasks. \textcolor[rgb]{1,0,0}{Red color} denotes factual errors.}
  \label{unfaithful_exp}
  \centering
  \begin{tabular}{p{2cm}p{8cm}p{3cm}}
    \toprule
    \textbf{Tasks}    &   \textbf{Source}   &   \textbf{Output}    \\
    \midrule
    \hline
    Abstractive Summarization  &  The first vaccine for Ebola was approved by the FDA in 2019 in the US, five years after the initial outbreak in 2014. To produce the vaccine, scientists had to sequence the DNA of Ebola, then identify possible vaccines, and finally show successful clinical trials. Scientists say a vaccine for COVID-19 is unlikely to be ready this year, although clinical trials have already started.  & The first vaccine for Ebola was \textcolor[rgb]{1,0,0}{rejected} in 2019. Scientists say a vaccine for \textcolor[rgb]{1,0,0}{Ebola} is unlikely to be ready this year. \\
    \hline
    Dialogue Generation & \textbf{Persona}: I have two cats. I work as a teacher at a middle school. My favorite color is yellow. I dislike country music. \textbf{Dialogue}: hi, do you have any pets? &  \textcolor[rgb]{1,0,0}{I do not have any pets}. Do you play any sports ? \\
    \hline
    Machine Translation & \begin{CJK}{UTF8}{gbsn} 迈克周四去书店。\end{CJK}(Michael goes to the bookstore on Thursday.) & Mike \textcolor[rgb]{1,0,0}{happily} goes to the bookstore on Thursday \textcolor[rgb]{1,0,0}{with his friend}. \\
    \hline
    Table-to-Text Generation & Name: Frank Lino; Caption: FBI surveillance photo; Birth date: October 30, 1938; Birth place: Gravesend, Brooklyn, New York, United States; & Frank Lino (born October 30, 1938 in Brooklyn) is an American \textcolor[rgb]{1,0,0}{criminal defense attorney}. \\
    \bottomrule
  \end{tabular}
\end{table}

The faithfulness problem is widely existed in nearly all NLG tasks, such as text summarization, dialogue generation, machine translation and table-to-text generation. Unfaithful examples of several popular tasks are shown in Table~\ref{unfaithful_exp}. For the non-open-ended NLG tasks and open-ended NLG tasks, the definitions of faithfulness problem are different.

\paragraph{Faithfulness in Non-open-ended NLG}
For non-open NLG tasks, NLG models generate text based on the input which provides complete or even more information for the output text. The faithfulness problem in non-open NLG tasks is defined as whether the generated content is factually consistent with the input information, often referred to as factual consistency.
For example, faithfulness in text summarization is whether the generated summary is factually consistent with and faithful to the input document. If the summary has hallucinations that not contained by the input document, then it's unfaithful to the input document. Similarly, faithfulness in machine translation is whether the translation is consistent with and faithful to the original language.

\paragraph{Faithfulness in Open-ended NLG}
For Open-ended NLG tasks, the model needs to leverage knowledge in Knowledge Graph or corpus to create new content that not contained by the input. The faithfulness problem in open-ended NLG tasks is defined as whether the generated content is factually consistent with the world knowledge or commonsense, often referred to as factual correctness.
For example, the faithfulness in news article generation is whether the facts in the generated article actually exists or happen in the real world. One relevant research topic is fake news detection \citep{shu2017fake,zhang2020overview}.

\begin{figure}
	\centering
	\includegraphics[width=3.5in]{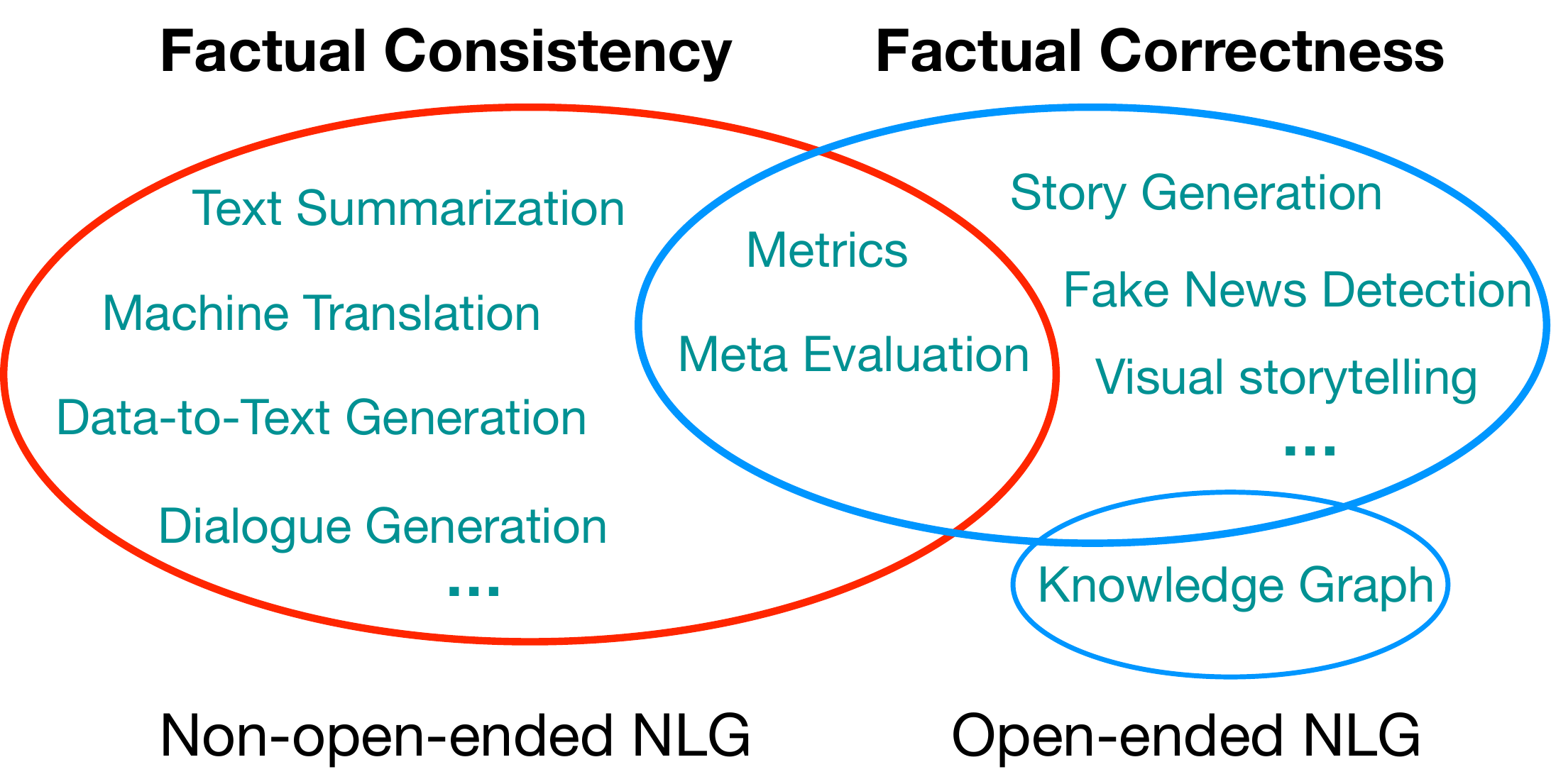}
	\caption{The research framework on the faithfulness problem.}
	\label{research_framework}
\end{figure}

To address the faithfulness problem, a lot of automatic faithfulness evaluation metrics and meta evaluations for these metrics have been proposed. Besides, much effort has been devoted to optimizing faithfulness for different NLG tasks. The framework of existing researches is demonstrated in Figure~\ref{research_framework}. Since the research on faithfulness mainly focuses on non-open-ended tasks, such as text summarization, machine translation, knowledge-grounded dialogue generation and data-to-text generation, \textbf{this paper mainly studies the faithfulness (i.e. factual consistency) in non-open-ended tasks}. We conduct a comprehensive survey of existing researches on faithfulness, including problem analysis, evaluation metrics, and optimization approaches.

\begin{figure}
\centering   
\begin{forest}
  forked edges,
  for tree={
    grow'=0,
    draw,
    align=c,
    font=\footnotesize,
    rounded corners
  },
  [{Faithfulness}, for tree={fill=black!10,draw=black}
    [{Problem\\Analysis \ref{problem_analysis}}, for tree={fill=purple!10,draw=purple,text width=60}
    [{Definition\&\\Categorization}
        [{FRANK[\citenum{pagnoni_understanding_2021}], NPH[\citenum{dziri_neural_2021}], etc.}, for tree={fill=purple!10,draw=purple!10,text width=120}]
      ]
    [{Challenges\\\&Issues}
        [{Model Analysis [\citenum{pagnoni_understanding_2021, cao_cliff_2021}], etc.}, for tree={fill=purple!10,draw=purple!10,text width=120}]
        [{Evaluation Problem [\citenum{kryscinski_evaluating_2019,fabbri_qafacteval_2021}]}, for tree={fill=purple!10,draw=purple!10,text width=120}]
        [{Annotation Problem [\citenum{falke_ranking_2019,pagnoni_understanding_2021}]}, for tree={fill=purple!10,draw=purple!10,text width=120}]
      ]
     [{Cause Analysis}
        [{Data Divergence [\citenum{maynez_faithfulness_2020,dhingra_handling_2019,wiseman2017ChallengesDatatoDocument}]}, for tree={fill=purple!10,draw=purple!10,text width=120}]
      [{Exposure Bais [\citenum{wang_exposure_2020,maynez_faithfulness_2020}], etc.}, for tree={fill=purple!10,draw=purple!10,text width=120}
      ]
      [{Poor Representation [\citenum{longpre2021entity}], etc.}, for tree={fill=purple!10,draw=purple!10,text width=120}
      ]
      ]
    ]
    [{Evaluation\\Metrics \ref{metrics}}, for tree={fill=orange!10,draw=orange}
      [{Meta Evaluation}, for tree={text width=75}
        [{FRANK[\citenum{pagnoni_understanding_2021}], SUMMAc[\citenum{laban_summac_2021}], BEGIN[\citenum{DBLP:journals/corr/abs-2105-00071}], etc.}, for tree={fill=orange!10,draw=orange!10, text width=180}]
      ]
      [{NLI-based Metrics}, for tree={text width=75}
        [{DAE[\citenum{goyal_evaluating_2020}], FactCC[\citenum{hadash2018estimate}], DialogNLI[\citenum{welleck_dialogue_2019}], etc.}, for tree={fill=orange!10,draw=orange!10, text width=180}]
      ]
      [{QA-based Metrics}, for tree={text width=75}
        [{QAGS[\citenum{DBLP:conf/acl/WangCL20}], FEQA[\citenum{hadash2018estimate}], $Q^2$[\citenum{DBLP:conf/emnlp/HonovichCANSA21}], QUALS[\citenum{DBLP:conf/acl/NanSZNMNZWAX20}] etc.}, for tree={fill=orange!10,draw=orange!10, text width=180}]
      ]
      [{Fact-based Metrics}, for tree={text width=75}
        [{Entity}, for tree={fill=orange!10,draw=orange, text width=40}
           [{EntityAlign[\citenum{nan2021entity}], SimAlign[\citenum{sabet2020simalign}]}, for tree={fill=orange!10,draw=orange!10, text width=125}]
        ]
        [{N-gram}, for tree={fill=orange!10,draw=orange, text width=40}
           [{PARENT[\citenum{dhingra_handling_2019}], PARENT-T[\citenum{wang2020FaithfulNeural}]}, for tree={fill=orange!10,draw=orange!10, text width=125}]
        ]
        [{Relation}, for tree={fill=orange!10,draw=orange, text width=40}
           [{TripleAlign[\citenum{hadash2018estimate}], ArcsAlign[\citenum{goyal_evaluating_2020}]}, for tree={fill=orange!10,draw=orange!10, text width=125}]
        ]
      ]
      [{Other Metrics}, for tree={text width=75}
        [{BARTScore[\citenum{yuan2021bartscore}], COCO[\citenum{xie_factual_2021}], TokenCLS[\citenum{zhou_detecting_2021}]}, for tree={fill=orange!10,draw=orange!10, text width=185}]
      ]
    ]
    [{Optimization\\Methods \ref{methods}}, for tree={fill=blue!10,draw=blue}
      [{Factual\\Guidance}, for tree={text width=60}
        [{Abstractive\\Summarization}, for tree={text width=60}
           [{Keyword[\citenum{dou2021gsum}], Sentence[\citenum{song_attractive_2020}],
           Relation[\citenum{cao_faithful_2017}]}, for tree={fill=blue!10,draw=blue!10, text width=160}]
        ]
        [{Dialogue\\Generation}, for tree={text width=60}
           [{Inplicit[\citenum{DBLP:conf/acl/LiGBSGD16}], Extractive[\citenum{DBLP:conf/aaai/GhazvininejadBC18}], Retrived[\citenum{DBLP:conf/iclr/DinanRSFAW19}]}, for tree={fill=blue!10,draw=blue!10, text width=160}]
        ]
        [{Data-to-Text\\Generation}, for tree={text width=65}
            [{SANA [\citenum{wang2021SketchRefine}], Segment[\citenum{shen2020NeuralDatatoText}], Entity [\citenum{liu2021FaithfulnessOpen}]}, for tree={fill=blue!10,draw=blue!10, text width=160}]
        ]
      ]
      [{Auxilary Tasks} , for tree={fill=yellow!10,draw=yellow}
        [{Abstractive\\Summarization}, for tree={text width=60}
          [{Entailment[\citenum{li_ensure_2018,falke_ranking_2019}], QA[\citenum{DBLP:conf/acl/NanSZNMNZWAX20}], Others[\citenum{zhang_optimizing_2020}]}, for tree={fill=yellow!10,draw=yellow!10, text width=160}]
        ]
        [{Dialogue\\Generation}, for tree={text width=60}
           [{NLI-reranking [\citenum{welleck_dialogue_2019}], NLI-RL [\citenum{song_profile_2020, mesgar_improving_2021}]}, for tree={fill=yellow!10,draw=yellow!10, text width=160}]
        ]
        [{Data-to-Text\\Generation}, for tree={text width=65}
            [{EntityMatching [\citenum{wang2020FaithfulNeural}], FocusAttn[\citenum{liu2019ComprehensiveDescription}]}, for tree={fill=yellow!10,draw=yellow!10, text width=160}]
        ]
        [{Machine\\Translation}, for tree={text width=65}
            [{FENMT [\citenum{weng_towards_2020}], WordAlignment[\citenum{zhang2021neural,feng2020modeling,garg2019jointly}]}, for tree={fill=yellow!10,draw=yellow!10, text width=160}]
        ]
      ]
      [{Post-Editing} , for tree={fill=green!10,draw=green}
        [{Abstractive\\Summarization}, for tree={text width=65}
           [{SpanFact[\citenum{dong2020multi}], FactCorrect[\citenum{cao2020factual}], ContrastSel[\citenum{chen_improving_2021}]}, for tree={fill=green!10,draw=green!10, text width=160}]
        ]
        [{Dialogue\\Generation}, for tree={text width=65}
           [{GDR [\citenum{DBLP:conf/acl/SongWZLL20}], NeuralPathHunter [\citenum{dziri_neural_2021}]}, for tree={fill=green!10,draw=green!10, text width=160}]
        ]
      ]
    ]
  ]
\end{forest}
\end{figure}

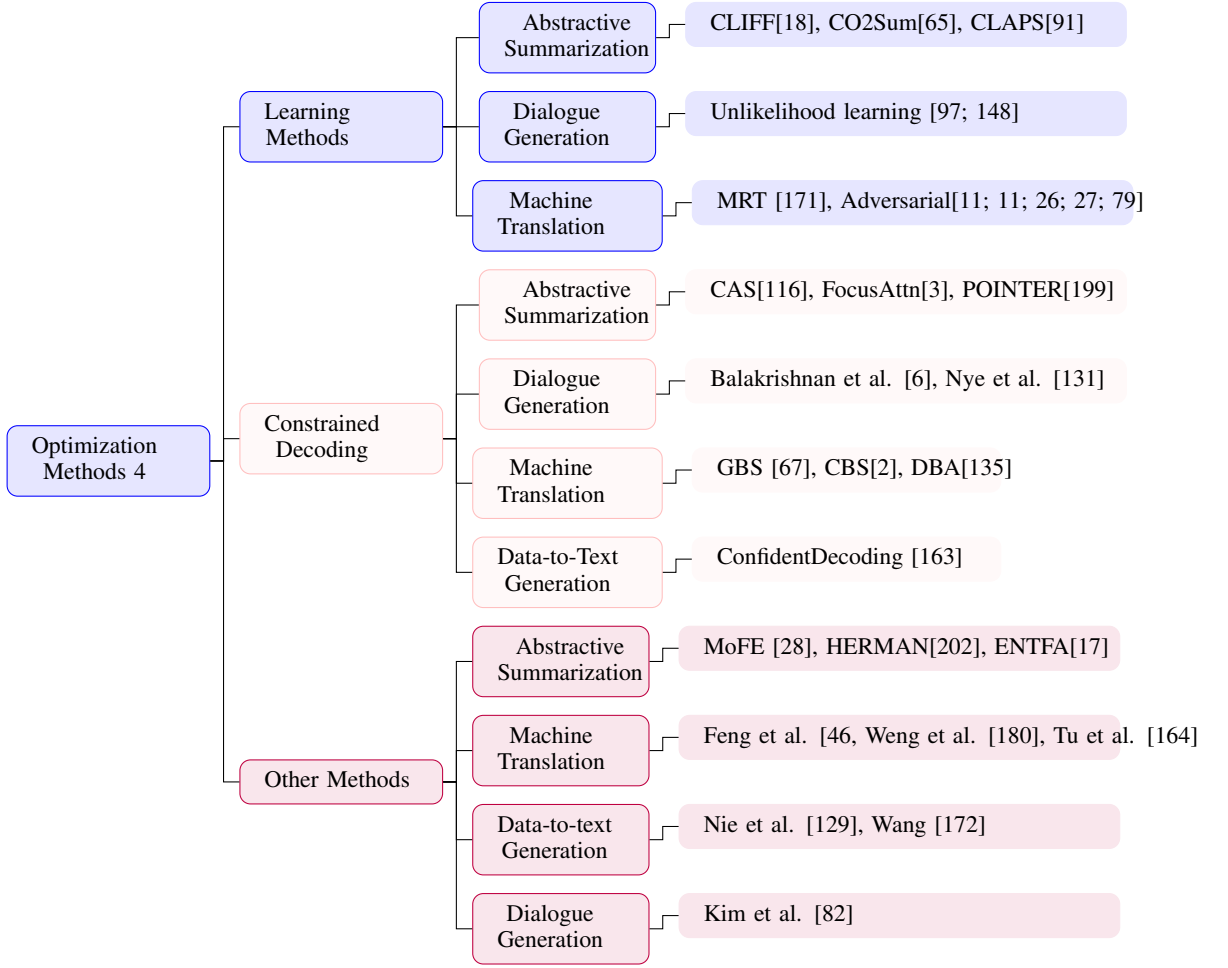
\begin{figure}
\centering
\begin{forest}
  forked edges,
  for tree={
    grow'=0,
    draw,
    align=c,
    font=\footnotesize,
    rounded corners
  },
  highlight/.style={
    thick,
    font=\sffamily\bfseries
  }
  [{Optimization\\Methods \ref{methods}}, for tree={fill=blue!10,draw=blue,text width=70}
  [{Learning\\Methods} , for tree={fill=blue!10,draw=blue}
        [{Abstractive\\Summarization}, for tree={text width=60}
        [{CLIFF[\citenum{cao_cliff_2021}], CO2Sum[\citenum{hadash2018estimate}], CLAPS[\citenum{lee_contrastive_2021}]}, for tree={fill=blue!10,draw=blue!10, text width=160}]
        ]
        [{Dialogue\\Generation}, for tree={text width=60}
            [{Unlikelihood learning [\citenum{DBLP:conf/acl/LiRKWBCW20,DBLP:conf/eacl/RollerDGJWLXOSB21}]}, for tree={fill=blue!10,draw=blue!10, text width=160}]
        ]
        [{Machine\\Translation}, for tree={text width=65}
            [{MRT [\citenum{wang_exposure_2020}], Adversarial[\citenum{belinkov2017synthetic,cheng2018towards,belinkov2017synthetic,karpukhin2019training,cheng_robust_2019}]}, for tree={fill=blue!10,draw=blue!10, text width=160}]
        ]
      ]
      [{Constrained\\Decoding} , for tree={fill=pink!10,draw=pink}
        [{Abstractive\\Summarization}, for tree={text width=60}
            [{CAS[\citenum{mao2020constrained}], FocusAttn[\citenum{DBLP:conf/acl/AralikatteNMRM20}], POINTER[\citenum{zhang_pointer_2020}]}, for tree={fill=pink!10,draw=pink!10, text width=160}]
        ]
        [{Dialogue\\Generation}, for tree={text width=60}
        [{
            Balakrishnan et al. [\citenum{DBLP:conf/acl/BalakrishnanRUW19}], Nye et al. [\citenum{,nye2021improving}]}, for tree={fill=pink!10,draw=pink!10, text width=160}]
        ]
        [{Machine\\Translation}, for tree={text width=65}
            [{GBS [\citenum{hokamp_lexically_2017}], CBS[\citenum{anderson_guided_2017}], DBA[\citenum{post_fast_2018}]}, for tree={fill=pink!10,draw=pink!10, text width=110}]
        ]
        [{Data-to-Text\\Generation}, for tree={text width=65}
            [{ConfidentDecoding [\citenum{tian2019StickingFacts}]}, for tree={fill=pink!10,draw=pink!10, text width=110}]
        ]
      ]
      [{Other Methods} , for tree={fill=purple!10,draw=purple}
        [{Abstractive\\Summarization}, for tree={text width=60}
            [{MoFE [\citenum{choubey_mofe_2021}], HERMAN[\citenum{zhao_reducing_2020}], ENTFA[\citenum{cao_inspecting_2021}]}, for tree={fill=purple!10,draw=purple!10, text width=160}]
        ]
        [{Machine\\Translation}, for tree={text width=60}
            [{Feng et al. [\citenum{feng2020modeling}, Weng et al. [\citenum{weng2020gret}], Tu et al. [\citenum{tu2016modeling}]}, for tree={fill=purple!10,draw=purple!10, text width=160}]
        ]
        [{Data-to-text\\Generation}, for tree={text width=60}
            [{Nie et al. [\citenum{nie2019SimpleRecipe}], Wang [\citenum{wang2019RevisitingChallenges}]}, for tree={fill=purple!10,draw=purple!10, text width=160}]
        ]
        [{Dialogue\\Generation}, for tree={text width=60}
            [{Kim et al. [\citenum{DBLP:conf/emnlp/KimKK20}]}, for tree={fill=purple!10,draw=purple!10, text width=160}]
        ]
    ]
]
\end{forest}
\caption{The content typology of the survey.}
\label{content_structure}
\end{figure}

\subsection{Structure of This Survey}
The content typology of this survey is shown in Figure~\ref{content_structure}.

In Section~\ref{problem_analysis}, we give a systematic analysis on the faithfulness problem in NLG, including categorization of unfaithful errors, manual annotations, challenges for evaluating and optimizing faithfulness, cause analysis, and relations with other aspects.

In Section~\ref{metrics}, we organize the various evaluation metrics proposed for faithfulness evaluation, and combine the meta-evaluations for these metrics to facilitate future research on faithfulness evaluations.

In Section~\ref{methods}, we summarize different optimization methods from both the perspective of tasks and methodology, and detail their relative advantages.

\section{Problem Analysis}
\label{problem_analysis}

In general, the task of natural language generation (NLG) targets at finding an optimal sequence $y_{< T+1} = (y_1, y_2, \ldots, y_T)$ that satisfies:
\begin{equation}
\begin{aligned}
    y_{<T+1} =& \mathop{\arg\max}_{y_{<T+1} \in \mathcal{S}} log P_{\theta} (y_{<T+1} | x) \\
    & \mathop{\arg\max}_{y_{<T+1} \in \mathcal{S}} \sum_{t=1}^{T} log P_{\theta} (y_{t} |y_{<t}, x)
\end{aligned}
\label{eq1}
\end{equation}
\noindent where $T$ represents the number of tokens of the generated sequence, $\mathcal{S}$ represents a set containing all possible sequences, and $P_{\theta} (y_{t} |y_{<t}, x)$ is the conditional probability of the next token $y_t$ based on its previous tokens $y_{<t} = (y_1, y_2, \ldots, y_{t-1})$ and the source sequence $x$ with model parameters $\theta$.

\subsection{Definition and Categorization}
We define the output sequence as being unfaithful if it has a span $y_i,...,y_j$, that is not supported by the input sequence $x$. The faithfulness issues, i.e. factual inconsistent with source sequence, can be divided into two categories:
\begin{itemize}
    \item \textbf{Intrinsic Error}: the fact that is contradicted to the source sequence $x$ due to synthesizing content using information present in $x$, which is also referred to ``intrinsic hallucination'' in \citet{maynez_faithfulness_2020}.
    \item \textbf{Extrinsic Error}: the fact that is neither supported nor contradicted by the source, which is also referred to ``extrinsic hallucination'' in \citet{maynez_faithfulness_2020}. 
\end{itemize}

Frank \citep{pagnoni_understanding_2021} defines a fine-grained typology of factual errors for text summarization, which is theoretically grounded in frame semantics \citep{fillmore1976frame,palmer2005proposition} and linguistic discourse analysis \citep{brown1983discourse}. It can also be applied to other non-open-ended NLG tasks, such as dialogue generation, machine translation and table-to-text generation.

\begin{table}
\renewcommand\arraystretch{1.5}
  \caption{Hierarchical typology of unfaithful errors. Examples of text summarization are shown. The source input is the same as the first line in Table~\ref{unfaithful_exp}.}
  \label{infaithful_categorization}
  \centering
  \begin{tabular}{l|l|p{2cm}|p{7cm}}
    \toprule
    \multicolumn{3}{c|}{Categorization}     & Examples     \\
    \midrule
    \multirow{5}{1.5cm}{Intrinsic Error} & \multirow{3}{1.5cm}{Semantic Frame Errors} & Predicate Error (PredE) & The Ebola vaccine was \textcolor[rgb]{1,0,0}{rejected} by the FDA in 2019. \\
    \cline{3-4}
    & & Entity Error (EntE) & Scientists say a vaccine for \textcolor[rgb]{1,0,0}{Ebola} is unlikely to be ready this year. \\
    \cline{3-4}
    & & Circumstance Error (CircE) & The first vaccine for Ebola was approved by the FDA in \textcolor[rgb]{1,0,0}{2014}. \\
    \cline{2-4}
    & \multirow{2}{1.5cm}{Discourse Errors} & Co-reference Error (CorefE) & The first vaccine for Ebola was approved in 2019. \textcolor[rgb]{1,0,0}{They} say a vaccine for COVID-19 is unlikely to be ready this year. \\
    \cline{3-4}
    & & Discourse Link Error (LinkE) & To produce the vaccine, scientists have to show successful human trials, \textcolor[rgb]{1,0,0}{then} sequence the DNA of the virus. \\
    \hline
    \multirow{2}{1.5cm}{Extrinsic Error} & \multicolumn{2}{c|}{Factual} & \textcolor[rgb]{1,0,0}{China} has already started clinical trials of the COVID-19 vaccine. \\
    \cline{2-4}
    & \multicolumn{2}{c|}{Non-Factual} & \textcolor[rgb]{1,0,0}{China didn't} start clinical trials of the COVID-19 vaccine. \\
    \bottomrule
  \end{tabular}
\end{table}

The fine-grained categories of factual errors mainly include:
\begin{enumerate}
\item \textbf{Semantic Frame Errors} capture factual errors in frame semantic and its core and non-core frame elements, including:
\begin{itemize}
    \item Predicate Error (PredE) denotes errors where the predicate is inconsistent with the source text; 
    \item Entity Error (EntE) denotes errors where the primary arguments (like entities) of the predicate are wrong or have the wrong attributes;
    \item Circumstance Error (CircE) denotes errors where the arguments and predicates interact (e.g. location, time, manner, direction, modality) are wrong.
\end{itemize}
\item \textbf{Discourse Errors} capture erroneous links between discourse segments, including:
\begin{itemize}
    \item Co-reference Error (CorefE) denotes errors where pronouns and other types of references to previously mentioned entities either are incorrect or have no clear antecedents;
    \item Discourse Link Error (LinkE) denotes incorrect discourse link between different statements.
\end{itemize}
\item \textbf{Content Verifiable Errors}
capture erroneous information that cannot be verified against the source text, which are mainly caused by:
\begin{itemize}
    \item Out of Article Error (OutE) denotes information that cannot be deduced by from the original text (the same as extrinsic hallucinations \citep{maynez_faithfulness_2020});
    \item Grammatical Error (GramE) denotes not well formed statements that make their meaning incomprehensible or ambiguous and cannot be verified against the source.
\end{itemize}
\end{enumerate}

Despite all extrinsic errors are assumed incorrect, \citet{cao_inspecting_2021} and \citet{maynez_faithfulness_2020} find that much hallucinated content is factual, namely consistent with world knowledge.
Factual hallucinations refer to content that is verifiable by world knowledge but not inferable from source text.
For example, in text summarization, they find that more than half of the hallucinated entities are factual with respect to the source document and world knowledge. These factual hallucinations can be beneficial in a summary by providing useful background information. Thus, the extrinsic errors or OutEs can be further categorized into factual hallucinations and non-factual hallucinations. 

Combining these definitions and categorization, we define a more thorough hierarchical typology of the faithfulness problems, as shown in Table~\ref{infaithful_categorization}.
This typology provides us with the means to categorize the types of errors made by generation models, helping us gain deeper insights than simply categorizing content as faithful or unfaithful.

\subsection{Challenges and Issues}
\paragraph{Model Analysis}
A lot of researches make annotations with different granularity to analyze the faithfulness performance of existing language generation models. The results show that even the most powerful pertaining models suffer from serious unfaithful problems.
Take the abstractive summarization task for example, the annotation results of the ratio of unfaithful summaries generated by several popular models including T5 \citep{raffel2019exploring}, BART \citep{lewis2019bart} and PEGASUS \citep{zhang2020pegasus}, are shown in Table~\ref{model_analysis}.
The annotation results are combined from \citet{pagnoni_understanding_2021} and \citet{cao_cliff_2021}.

\begin{table}[h!]
  \renewcommand\arraystretch{1.5}
  \caption{The ratio of unfaithful summaries annotated by human for different systems.}
  \label{model_analysis}
  \centering
  \begin{tabular}{p{3cm}p{2cm}p{2cm}}
    \toprule
    \textbf{System}    &   \textbf{XSum}   &   \textbf{CNN/DM}    \\
    \midrule
    \hline
    TransS2S & 96.9\% & 74.8\% \\
    BERTSum & 83.7\% & 27.2\% \\
    T5 & 82.0\% & 26.7\% \\
    BART & 66.7\% & 24.7\% \\
    PEGASUS & 60.7\% & 13.3\% \\
    \bottomrule
  \end{tabular}
\end{table}

The above results show that all systems generate over 60\% unfaithful summaries on the XSum dataset.
Also, on the CNN/DM dataset, T5 and BART generate over 20\% unfaithful summaries.
On the one hand, the above results show the severity of the faithfulness problem of current models, and on the other hand, it also shows that the impact of different datasets is also very large.
We will analyze the influence of dataset in Section~\ref{cause_analysis}.

\paragraph{Evaluation}
Common automatic evaluation metrics for text generation based on n-gram overlap – BLEU, ROUGE, and METEOR \citep{papineni2002bleu,lin2004rouge,banerjee2005meteor} – are insufficient to measure the faithfulness of the generated text. \citet{kryscinski_evaluating_2019} and \citet{fabbri_qafacteval_2021} find that they have low correlation with human judgements of factuality, as shown in Table~\ref{eval_problem}.
So, a lot of new evaluation methods are proposed to evaluate the faithfulness of generated text for different tasks. We will describe them in Section~\ref{metrics}.

\begin{table}[h!]
  \renewcommand\arraystretch{1.5}
  \renewcommand\tabcolsep{4pt}
  \caption{The Person and Spearman correlation between different n-gram based metrics and human annotation of faithfulness on CNN/DM dataset, XSum dataset and their combination.}
  \label{eval_problem}
  \centering
  \begin{tabular}{l|cccc|cccc|cccc}
    \toprule[1pt]
    &
    \multicolumn{4}{c|}{\textbf{All data}}    &   \multicolumn{4}{c|}{\textbf{CNN/DM}}   &   \multicolumn{4}{c}{\textbf{XSum}}    \\
    \midrule[1pt]
    \multirow{2}{*}{Metrics} & \multicolumn{2}{c}{Person} & \multicolumn{2}{c|}{Spearman} & \multicolumn{2}{c}{Person} & \multicolumn{2}{c|}{Spearman} & \multicolumn{2}{c}{Person} & \multicolumn{2}{c}{Spearman} \\
    & $\rho$ & p-val & $\gamma$ & p-val & $\rho$ & p-val & $\gamma$ & p-val & $\rho$ & p-val & $\gamma$ & p-val \\
    \hline
    BLEU & 0.10 & 0.00 & 0.07 & 0.00 & 0.08 & 0.01 & 0.08 & 0.01 & 0.14 & 0.00 & 0.20 & 0.00 \\
    METEOR & 0.14 & 0.00 & 0.11 & 0.00 & 0.12 & 0.00 & 0.10 & 0.00 & 0.15 & 0.00 & 0.10 & 0.00 \\
    Rouge-1 & 0.14 & 0.00 & 0.10 & 0.00 & 0.12 & 0.00 & 0.10 & 0.00 & 0.15 & 0.00 & 0.09 & 0.01 \\
    Rouge-2 & 0.12 & 0.00 & 0.08 & 0.00 & 0.08 & 0.00 & 0.07 & 0.01 & 0.17 & 0.00 & 0.14 & 0.00 \\
    Rouge-L & 0.13 & 0.00 & 0.09 & 0.00 & 0.11 & 0.00 & 0.09 & 0.00 & 0.16 & 0.00 & 0.10 & 0.00\\
    \bottomrule
  \end{tabular}
\end{table}

\paragraph{Annotation}
Faithfulness annotation of NLG models is very difficult.
Most existing work consider faithfulness as a binary concept, annotating generated text as faithful or unfaithful \citep{maynez_faithfulness_2020}.
However, \citet{falke_ranking_2019} showed relatively low crowd–expert agreement, indicating the presence of subjectivity in the annotation process.
\citet{pagnoni_understanding_2021} annotated the faithfulness of summarization systems in a more fine-grained manner, however, the inter-annotator agreement is also low.
They collect human annotations from three independent annotators. 
The inter-annotator agreement in terms of Fleiss Kappa $\kappa$ \citep{fleiss1971measuring} is 0.58 for faithful or not, and 0.39 for specific unfaithful error types shown in Table~\ref{infaithful_categorization}, which all indicate low inter-annotator agreement.
\citet{tang2021investigating} compared the reliability of ranking and rating-based human annotations of faithfulness in summarization models and found that ranking-based Best-Worst Scaling annotations are largely reliable than rating-based annotations.

\subsection{Cause Analysis}
\label{cause_analysis}
Many factors can affect the faithfulness of model-generated results, such as dataset, training method, and model expressiveness.

\paragraph{Data divergence between source and reference.}
The divergence between source and reference is one of the main reason for extrinsic hallucinations during generation.
For example, in text summarization, summaries were usually written by journalists as introductions to the news articles they precede. These summaries, therefore, often have true additional information not found in the document. Such divergence issue between source and target is not uncommon in conditional text generation \citep{kryscinski_evaluating_2019,wiseman2017ChallengesDatatoDocument,dhingra_handling_2019}.
The divergence may be a product of heuristic data collection, or it may be inevitable due to the nature of some NLG tasks, such as table-to-text generation and dialogue generation.

Existing models are usually agnostic to the source-reference divergence, making them vulnerable to hallucinations. Thus, models can generate texts that are not consistent with the input, yet would likely have reasonable model log-likelihood.
This is the main reason why the same model performs differently on different datasets, such as the difference in summarization performance on the XSum dataset and the CNN/DM dataset.
The XSum dataset is collected heuristically by simply taking the introductory sentence prefacing each article as its reference summary, so reference summaries often contain hallucinations.
\citet{maynez_faithfulness_2020} reported that 76.9\% of reference summaries contained unfaithful content.
In contrast, the reference summaries of the CNN/DM datasets are all human-written with less hallucinations.
Therefore, the faithfulness of the summarization model on the CNN/DM dataset is much better than that on the XSum dataset.

\paragraph{Exposure bias between training and inference.}
\citet{wang_exposure_2020} state that exposure bias \citep{ranzato2015sequence}, a discrepancy between training and inference, is partially to blame for hallucinations.
Specifically, the standard teacher-forcing training algorithm \citep{williams1989learning} used by most existing work can lead to a discrepancy between what the model sees during training and test time, resulting in degenerate outputs with factual hallucinations \citep{maynez_faithfulness_2020}.
Furthermore, the model is also only optimized to maximize the log-likelihood of the reference summary at the word-level, which does not necessarily reward models for being faithful.

\paragraph{Poor text representation.}
A model with poor input text representation will fail to do document level inference, often required for abstraction and generation, and will be vulnerable to such errors.
For example, in text summarization, the percentage of system summaries with intrinsic hallucination was much higher than in gold summaries. This phenomenon particularly revealed the models’ tendency to misrepresent information in the document due to the lack of document-level understanding and inference. To improve text representation, it is a common practice to leverage large pre-trained models for downstream NLG tasks.
Pre-training can improve text generation, due to its exposure to vast amount of text through pretraining, allowing it to integrate background knowledge with generation.
However, \citet{longpre2021entity} have discovered that such models usually over-rely on the parametric knowledge learned from large scale corpus over the provided input. Also, the dominant language model usually prompts the decoder generates common words to make sure outputs are fluent.



\section{Automatic Evaluation Metrics}
\label{metrics}
Recently, there has been wide empirical success in text summarization, machine translation, dialogue response generation, and other text generation tasks. For evaluation, these models generally rely on metrics like ROUGE (Recall-Oriented Understudy for Gisting Evaluation) \citep{lin2004rouge}, BLEU (Bilingual Evaluation Understudy) \citep{papineni2002bleu} and METEOR (Metric for Evaluation of Translation with Explicit ORdering) \citep{banerjee2005meteor} that measure locally constrained n-gram overlap. However, these metrics cannnot evaluate the faithfulness of generated text.

Recently, much work focus on evaluating the factual consistency of generated text and propose various new metrics for different NLG tasks. We categorize these metrics into 4 types: Entailment-based, QA-based, Fact-based, and Others, as shown in Table~\ref{eval_metrics}.

\begin{table}
\renewcommand\arraystretch{1.5}
  \caption{Categorization of Evaluation Metrics.}
  \label{eval_metrics}
  \centering
  \begin{tabular}{l|l|l}
    \toprule[1pt]
    \textbf{Categories}     &    \textbf{Methods}    &   \textbf{Target Tasks}     \\
    \midrule[1pt]
    \multirow{8}{*}{Entailment-based} & DAE \citep{goyal_annotating_2021}  & Summarization, Paraphrasing \\
    & RankNLI \citep{falke_ranking_2019} & Summarization \\
    & SummaC \citep{laban_summac_2021} & Summarization \\
    & DialogNLI \citep{welleck_dialogue_2019} & Dialogue Generation \\
    & FactCC \citep{kryscinski_evaluating_2019} & Summarization \\
    & RCDG \citep{song_generating_2020} & Dialogue Generation \\
    & KvPI\citep{song_profile_2020} & Dialogue Generation \\
    & CI-ToD \citep{DBLP:conf/emnlp/QinXHCXC21} & Dialogue Generation \\
    & DECODE \citep{DBLP:conf/acl/NieWBKW20} & Dialogue Generation \\
    & SentenceNLI \citep{mishra_looking_2021} & Summarization \\
   
    \hline
    \multirow{6}{*}{QA-based} & QAGS \citep{DBLP:conf/acl/WangCL20} & Summarization \\
    & FEQA \citep{durmus_feqa_2020} & Summarization \\
    & QAFactEval \citep{fabbri_qafacteval_2021} & Summarization \\
    & QuestEval \citep{scialom_questeval_2021} & Summarization \\
    & $Q^2$ \citep{DBLP:conf/emnlp/HonovichCANSA21} & Dialogue Generation \\
    & QUALS \citep{DBLP:conf/acl/NanSZNMNZWAX20} & Summarization \\
    \hline
    \multirow{5}{*}{Fact-based} & SimAlign \citep{sabet2020simalign} & Machine Translation \\
    & EntityAlign \citep{nan2021entity} & Summarization \\
    & TripleAlign \citep{goodrich_assessing_2019} & Summarization \\
    & PARENT \citep{dhingra_handling_2019} & Table-to-Text \\
    & PARENT-T \citep{wang2020FaithfulNeural} & Table-to-Text \\
    \hline
    \multirow{3}{*}{Others} & COCO \citep{xie_factual_2021} & Summarization \\
    & TokenLevelCLS \citep{zhou_detecting_2021} & Machine Translation, Summarization \\
    & BARTScore \citep{yuan2021bartscore} & 16 NLG tasks \\
    & ShannonScore \citep{egan_play_2021} & Summarization \\
    \bottomrule
  \end{tabular}
\end{table}

\subsection{Entailment-based Metrics}
\label{metric_entail}
\begin{figure}
	\centering
	\includegraphics[width=3in]{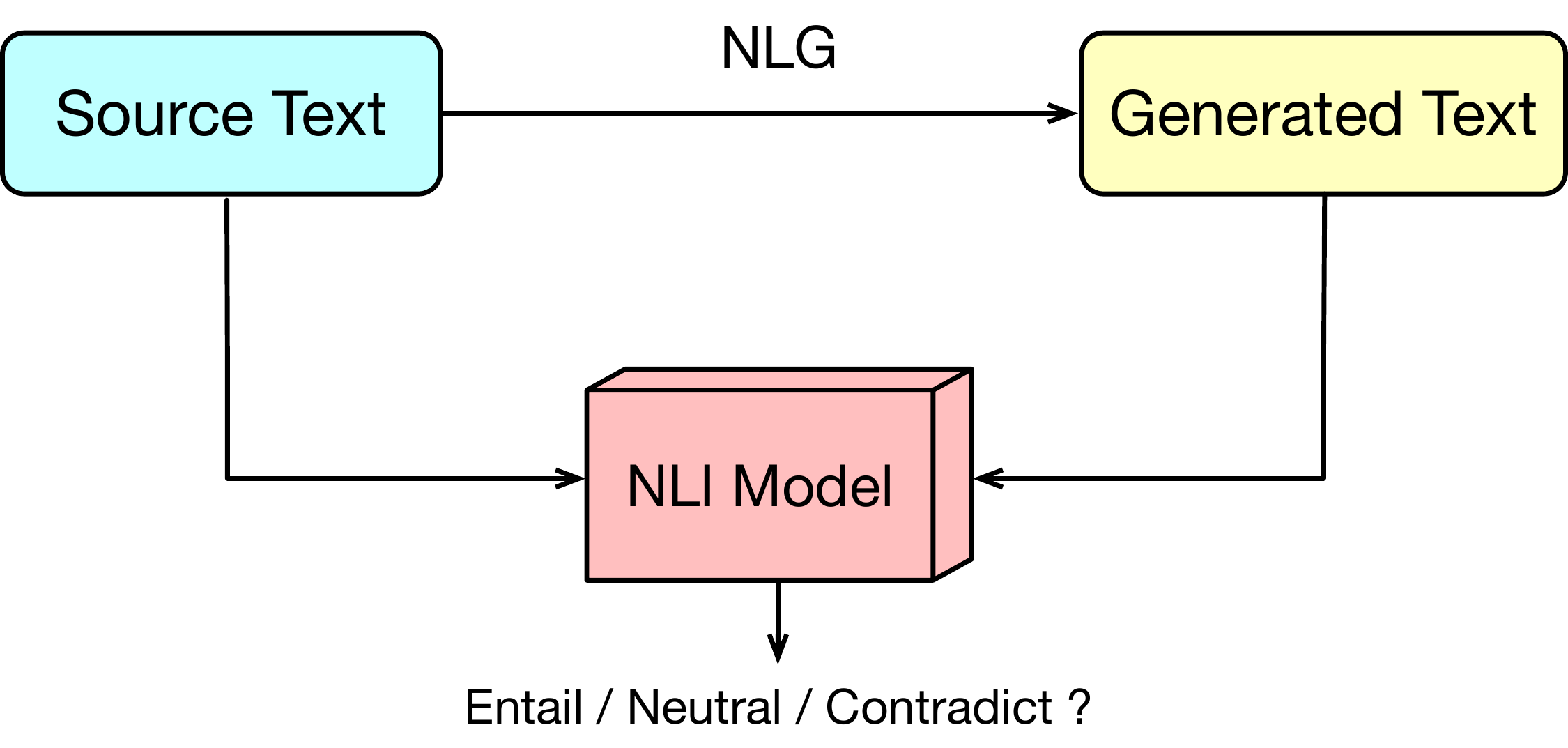}
	\caption{The framework of entailment-based metrics.}
	\label{eval_entail}
\end{figure}

One of the most popular methods is to apply NLI (Natural Language Inference) to access the faithfulness of generated texts, that is whether the generated text is entailed, neutral, or conflicting with a given input, as shown in Figure~\ref{eval_entail}.
The basic hypothesis is that the content of generated texts should be entailed by or at least not conflict with the source text.
Though an NLI model usually predicts three different scores for entailment, neutral, and contradiction, most work only utilize entailment score to evaluate faithfulness.
Formally, given a source text $x$ as a premise, a generated text $y$ as a hypothesis, an NLI model $\mathcal{N}$ predicts the entailment score as $\mathcal{N}(x,y)$.
The larger the $\mathcal{N}(x,y)$ is, the more faithful $y$ given $x$.
For evaluating the proposed metrics, most works report their correlations with human judgements, while some other works, especially entailment-based metrics, also propose ranking-based downstream tasks to demonstrate performances.
We will also introduce these ranking tasks in the following.

\paragraph{Sentence-level NLI}
Traditional NLI tasks predict entailment scores between sentences.
However, in the text generation scenario, the input text $x$ takes various forms and often contains multiple sentences that severely challenge the application of NLI. Earlier attempts directly apply NLI classifiers to access the factual consistency between input text $x$ and output text $y$.
They study how NLI models trained on traditional NLI datasets like  MNLI \citep{DBLP:conf/naacl/WilliamsNB18} perform.
\citet{falke_ranking_2019} proposed to aggregate entailment scores between sentences of $x$ and $y$ to calculate the faithfulness score between $x$ and $y$, namely \textbf{RankNLI}.
Given sentences $s_y \in y$, $s_x \in x$, RankNLI formalizes faithfulness between $y$ and $x$ as:
\begin{equation}
     \frac{1}{|y|}\sum_{s_y \in y} \max_{s_x \in x}\mathcal{N}(s_x,s_y)\label{RankNLI}
\end{equation}
They found that while entailment prediction should help with this problem, out-of-the-box NLI models performed poorly on this task.
\citet{falke_ranking_2019} also proposes a summary re-ranking task to evaluate the performance of RankNLI.
In this task, a better metric should help the summarization model to select  more faithful summaries during the reranking process of beam search.
They further analyze how  different architectures  of the NLI model $\mathcal{N}$, such as ESIM \citep{DBLP:conf/acl/ChenZLWJI17}, BERT \citep{devlin2018bert},  affect the summary ranking task.

\citet{maynez_faithfulness_2020} applied a much simpler strategy by directly using NLI models trained on MNLI to predict entailment score  $\mathcal{N}(x,y)$. 
\citet{DBLP:journals/corr/abs-2005-11739} further found that applying ANLI dataset instead of MNLI dataset in \citet{falke_ranking_2019} to train the  NLI classifier is more suitable for faithfulness evaluation.
\textbf{SummaC} \citep{DBLP:journals/corr/abs-2111-09525} comprehensively revisits sentence-level NLI for accessing faithfulness.
They apply a CNN module to aggregate the entailment score matrix between document and summary sentences, and demonstrate the potential of sentence-level NLI on various benchmarks.

\paragraph{Annotation-based}
The major problem of sentence-level NLI metrics is that they are inconsistent with their downstream tasks, which often require the evaluator to predict paragraph-level entailment scores. 
Some work attempted to directly train an NLI classifier between source text $x$ and target output $y$.
A straightforward solution is to annotate a  certain scale of samples for training the classifier.
In text summarization, \citet{DBLP:conf/acl/AralikatteNMRM20} and \citet{DBLP:journals/corr/abs-2102-01672} finetuned NLI classifier  on hundreds of (around 500) manual annotated samples for faithfulness evaluation and reported a good performance of this simple metric.

In dialog generation,  \citet{DBLP:conf/acl/WelleckWSC19} constructed a Dialog NLI dataset (DialogNLI) for factual consistency evaluation.
To save human labor, they annotate  the relation triples of dialogue sentences instead.
Based on relation triples, they inference the  NLI labels by certain rules.
\citet{DBLP:conf/acl/WelleckWSC19} also propose an utterance ranking task, which is often applied   to evaluate the factual consistency of a dialogue model.
In this task, given history utterances   $u_{<t}$, a dialogue model is asked to select the next utterance $u_{t}$ with the lowest perplexities from a set of candidate utterances $\mathcal{U}$:
\begin{equation}
    u_{t} = \arg\min_{u\in \mathcal{U}} -\log p(u|u_{<t}), \quad where  \quad \mathcal{U}=(\mathcal{U}^+, \mathcal{U}^-, \mathcal{U}^{rand} )
\end{equation}
where the generation probability $p$ is calculated by the dialogue model, $\mathcal{U}^+,\mathcal{U}^-, \mathcal{U}^{rand}$ represent the set of  utterances that are entailed with $u_{<t}$, conflicted with $u_{<t}$ or randomly selected, respectively.
The higher probability the model selects a $u_t$ from $\mathcal{U}^+$, the better factual consistency it has.

\citet{song_profile_2020} proposed a human-annotated dataset, namely Key-value Profile Identification (KvPI), with single-turn conversations and corresponding attribute profiles.
They further labeled NLI relations between each conversation and structured profile.
With the NLI labels, they trained a classifier to predict entailment relations between structured attributes and generated utterances. 
\citet{DBLP:conf/emnlp/QinXHCXC21} proposed a human-annotated dataset CI-ToD, which incorporates  NLI labels between various types of inputs including dialogue history, user query and the corresponding knowledge  base.
They propose a uniform model to assess all the factual consistency relations above.  
\citet{DBLP:conf/acl/NieWBKW20}  proposed a multi-domain dataset DECODE for evaluating consistency of dialogue. 
DECODE balances human-written contradicting dialogues with an equal number of non-contradicting dialogues from several public datasets. 
NLI models trained on DialogNLI or DECODE can be used for assessing the faithfulness of dialogues.
\citet{DBLP:journals/corr/abs-2110-08222}  proposed a benchmark  DialFact to evaluate and check knowledge errors in open domain dialog generation. 
They proposed a fact-checking pipeline for this benchmark, in which an NLI component is applied to check whether the generated utterance is supported by the retrieved evidences.

\paragraph{Weak Supervision}
Because it is difficult to annotate a large-scale document-level NLI dataset, several recent works apply weakly supervised methods for training.
Some work applied data augmentation methods to construct synthetic datasets.
\textbf{FactCC} \citep{kryscinski_evaluating_2019}  constructed positive entailment samples by different heuristic methods for weak supervised training.
For constructing various positive samples, they swap, deletes or insert textual units like entities, pronouns, numbers, etc.
To better explain this metric, FactCC also provides an additional version, FactCCX, which highlights spans of evidence in source documents.
\citet{DBLP:journals/corr/abs-2105-00071}  introduced a new benchmark BEGIN for evaluating factual consistency of knowledge grounded dialogue systems. 
For the evaluation metric, they also proposed a NLI classifier by constructing  adversarial samples similar to FactCC.  
SetenceNLI \citep{mishra_looking_2021} suggested that the major bottleneck in the utility of NLI models is that traditional NLI datasets do not exhibit long premises. 
To solve this problem, they convert multiple-choice reading
comprehension datasets into two-class NLI datasets using  data augmentation methods.
In addition to similar rule-based methods, they also applied text generation models to generate higher-quality samples.

\paragraph{Fine-grained Prediction}
Some works utilize fine-grained features to assess faithfulness. They convert direct predictions of $\mathcal{N}(x,y)$ into sequentially labeled fine-grained features in the generated text.
DAE~\citep{goyal_evaluating_2020} applies a new formulation of entailment that decomposes it at the level of dependency arcs.
Instead of making decisions at the sentence level, DAE sequentially predicts the entailment scores of each dependency arc in the generated sentence and aggregates them to obtain the final faithfulness score.
\citet{goyal_annotating_2021} further explores the difference in error distribution between synthetic and human written summaries, and investigates how fine-grained supervision information can benefit faithfulness evaluation.


\subsection{QA-based Metrics}
\label{qestion_answering_task}
\begin{figure}
	\centering
	\includegraphics[width=3in]{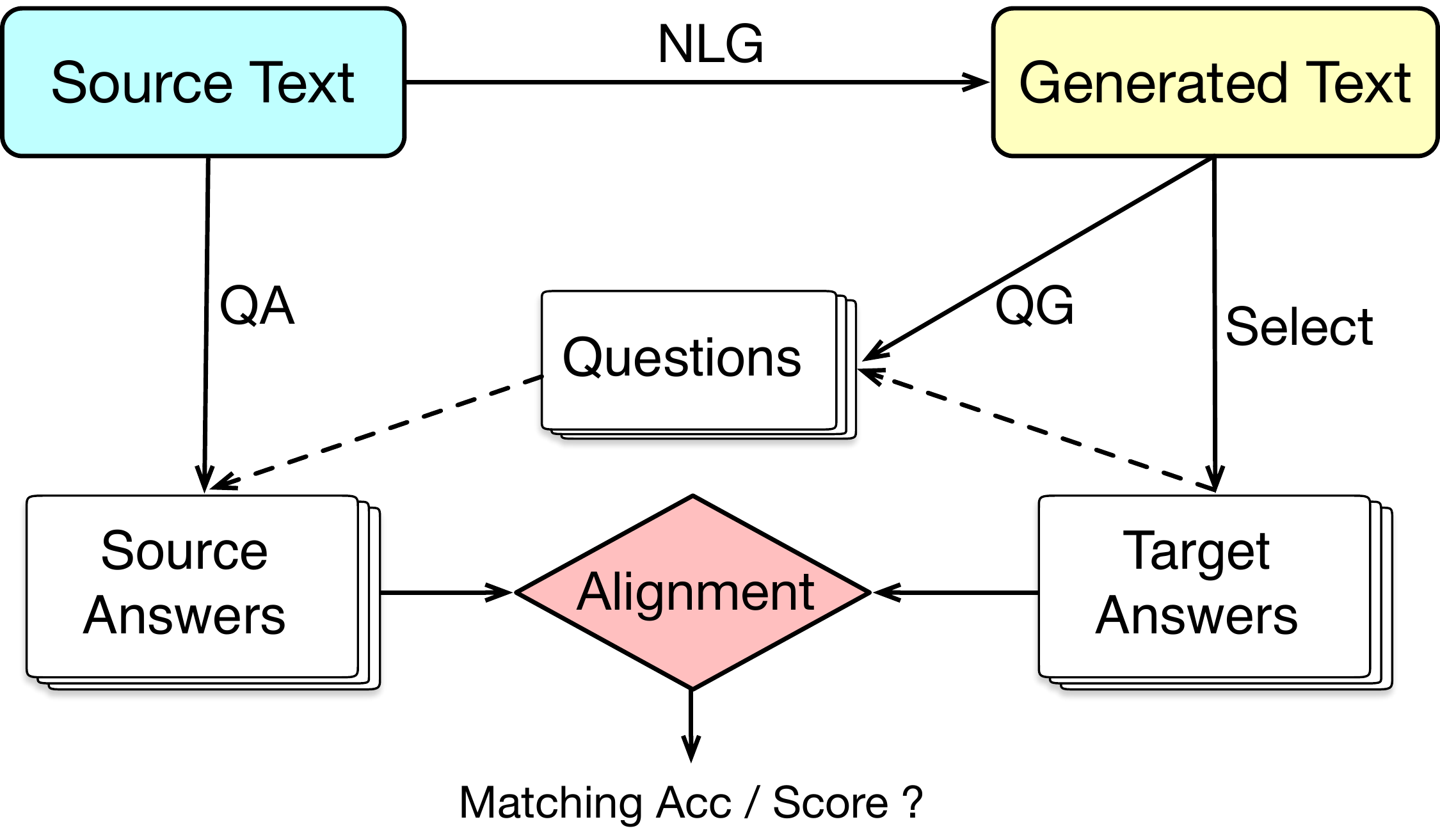}
	\caption{The framework of QA-based metrics.}
	\label{eval_qa}
\end{figure}

Because assessing faithfulness requires logical inference over factual information, it is natural to utilize the reasoning ability of Question Answering (QA) models.
Several recent works proposed QA-based factual evaluation metrics.
As shown in Figure~\ref{eval_qa}, these metrics often include two components: a question generation (QG) module and a QA module.
The core idea of these metrics is to predict the matching score between source answers (key information units from source text) and target answers  (key information units from generated text).
The overall procedure of these metrics are summarized as following:
\begin{enumerate}
\item \textbf{Answer Selection:} Extract information units from the generated text, which is viewed as  target answers.
\item \textbf{Question Generation:} Conditioned upon the selected target answers, the QG module generates questions using the generated text as context.
\item \textbf{Question Answering:} The  QA module answers the questions with the source text as context to retrieve source answers.
\item \textbf{Answer Alignment Evaluation:} Calculate the matching score between source and target answers by a answer alignment metric to output the final evaluation score.
\end{enumerate}
QAGS \citep{DBLP:conf/acl/WangCL20} and FEQA \citep{durmus_feqa_2020} are the earliest QA-based factual evaluation metrics.
These two metrics share similar model architectures and processing procedures introduced above.
In the procedure 1, QAGS extracted n-grams as the information units for target answers while FEQA extracted entities. 
In procedure 2-4, they both applied BERT-based QA modules, BART-based QG modules and token-level F1 as answer alignment metrics.

Several QA-based metrics followed the framework of QAGS and FEQA with moderate modifications.
QuestEval \cite{scialom_questeval_2021} extended this framework by adding an extra  procedure  to measure the recall-oriented performance.
The additional procedure generated  question-answer pairs from the source document and answered the questions from the generated text.
In contrast to QuestEval, QUALS \cite{DBLP:conf/acl/NanSZNMNZWAX20} simplified the above procedure 1-3 by only one neural language model (QAGen).
QUALS employs QAGen as proposed in (Shakeri et al., 2020), to generate both the questions and answers from the generated text.
In particular, given a summary $y$, QAGen outputs
a question-answer (q-a) pair jointly, separated by a special token <a>. Let $LL_{y}(q,a)$ be the average log likelihood of generating the q-a pair from the given summary $y$.
Then given the input document $x$, QUALS simply evaluates the average log likelihood of the QAGen model producing the same q-a pairs, denoted as $LL_{x}(q,a)$.
Formally, given a summary $y$ and input document $x$, QUALS score is computed as follows:
\begin{equation}
\begin{aligned}
    QUALS(x,y)=\frac{1}{M} \sum_{(q,a) \in y} (LL_{x}(q,a) - LL_{y}(q,a))
\end{aligned}
\label{eq_quals}
\end{equation}
\noindent where $M$ is the number of q-a pairs selected on the
summary $y$.
This simplification largely decreases the computational time and memory of the original QAGS.

QAFactEval \cite{fabbri_qafacteval_2021}  conducted extensive comparisons of  QA-based metrics and demonstrated that carefully choosing the components  of  a  QA-based  metric  is  critical to  performance.
The optimized settings of QAFactEval in each procedure are listed in the following:
\begin{enumerate}
    \item Select NP chunks as the textual units as target answers;
    \item Apply BART-QA2D \citep{DBLP:journals/corr/abs-1809-02922} for the  QG module and filter low quality generated questions;
    \item Apply Electra-large \citep{DBLP:conf/iclr/ClarkLLM20} for QA;
    \item Apply LERC \citep{DBLP:conf/emnlp/ChenSSG20} score as the answer alignment metric.
\end{enumerate}
With these carefully selected settings, \citet{fabbri_qafacteval_2021} boosted the performance of QA-based metric to a new level.

In addition to the factual metrics in text summarization listed above, \citet{DBLP:conf/emnlp/HonovichCANSA21} proposed a QA-based metric $Q^2$ for evaluating factual consistency in open-domain dialogue generation.
They utilized the entailment score predicted by an NLI model as the alignment metric for answer spans.

\subsection{Fact-based Metrics}
\begin{figure}
	\centering
	\includegraphics[width=3in]{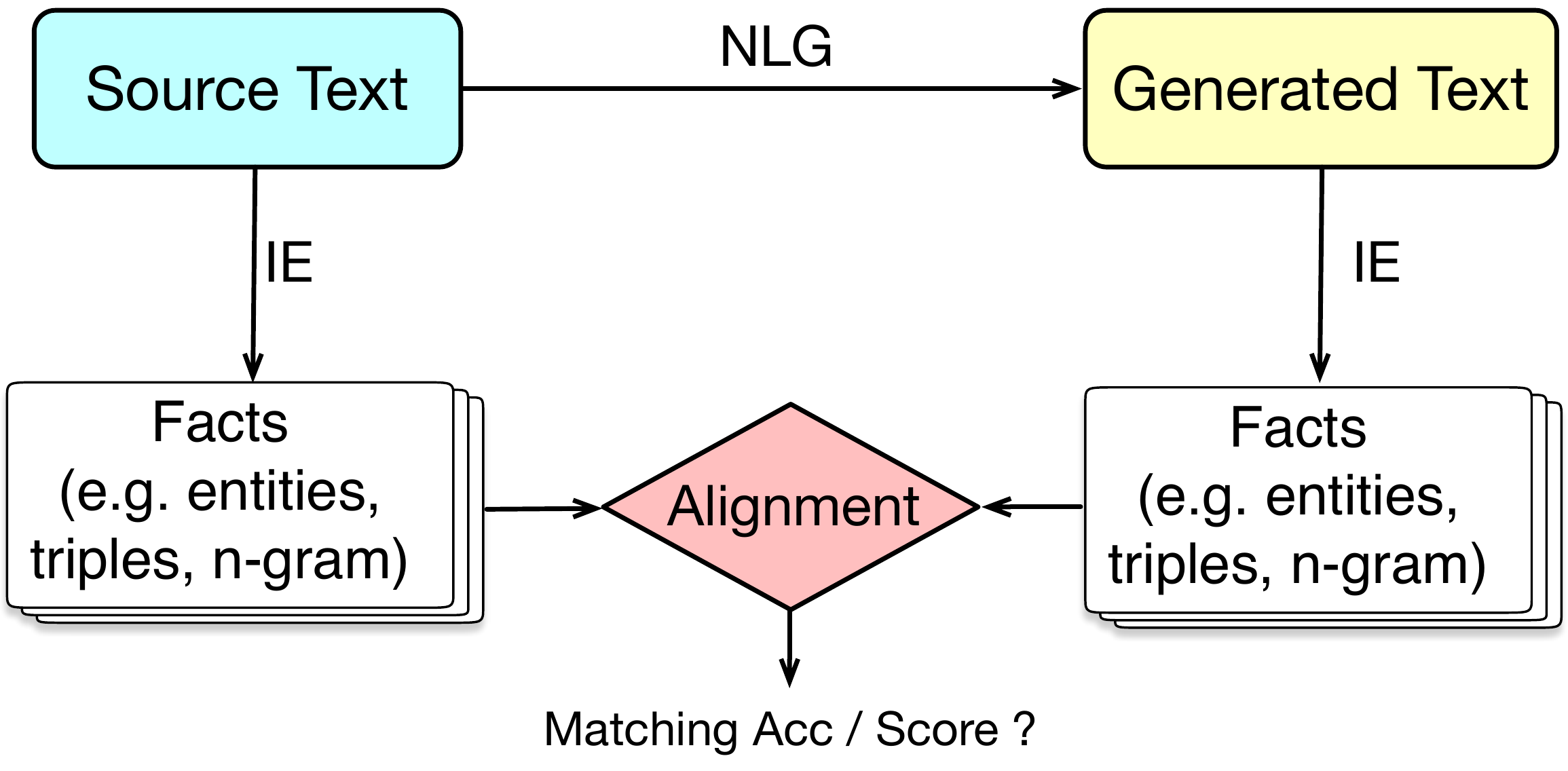}
	\caption{The framework of fact alignment-based metrics.}
	\label{fact_alignment}
\end{figure}

The most intuitive way to evaluating faithfulness is to count the fact overlap between generated text and source document, as shown in Figure~\ref{fact_alignment}. Facts can be represented in different forms, such as entities, n-grams and relation triples (subject, relation, object). 

Factual inconsistency can occur at either the entity or the relation level. At the entity level, a model generated text may contain named entities that never appeared in the source document.
At the relation level, the entities indeed exist in the source document but the relations between them are not in the source document.

\subsubsection{Entity-based}
\paragraph{EntityAlign} \citet{nan2021entity} proposed an entity-based metrics that rely on off-the shelf tools to perform Named-Entity Recognition(NER). Let $N(x)$ and $N(y)$ denote the number of named-entities in the source (input document) and target (generated text), respectively. $N(y \cap x)$ denotes the number of entities found in the generated text that can find a match in the source document. If a named entity in the generated text consists of multiple words, it is considered a match as long as any n-gram of the named-entity can be found in the source document.
The degree of faithfulness with respect to the source text is quantified: $\textbf{prec} = \mathcal{N}(y \cap x) / \mathcal{N}(y)$. 

\paragraph{SimAlign} For machine translation, \citet{sabet2020simalign} proposed to leverage multilingual word embeddings – both static and contextualized – for word alignment between source language and translated language. 

\subsubsection{Ngram-based}
\paragraph{PARENT}
For table-to-text generation task, \citet{dhingra_handling_2019} modeled facts as n-grams, and developed a metric PARENT (Precision And Recall of Entailed Ngrams from the Table) which aligns n-grams from the reference and generated texts to the semi-structured data before computing their precision and recall. When computing precision, PARENT effectively uses a union of the reference and the table, to reward correct information missing from the reference. When computing recall, it uses an intersection of the reference and the table, to ignore extra incorrect information in the reference. The union and intersection are computed with the help of an entailment model to decide if a text n-gram is entailed by the table.
The entailed precision and recall are combined into an F-score to give the PARENT metric for one instance. The system-level PARENT score for a model is the average of instance-level PARENT scores across the evaluation set.

\paragraph{PARENT-T}
PARENT-T \citep{wang2020FaithfulNeural} is a table-focused version of PARENT. When computing precision, PARENT-T considers an n-gram to be correct if it has a high probability of being entailed by the table. PARENT-T uses the word overlap model for computing entailment probability. 
For recall, PARENT-T only computes it against table to ensure that texts that mention more information from the table get higher scores. 
The system-level PARENT-T score for a model is the average of instance-level PARENT-T scores across the evaluation set.

\subsubsection{Relation-based}
\paragraph{TripleAlign}
Facts are usually represented by relation triples (subject, relation, object), where the subject has a relation to the object. To extract triples, \citet{goodrich_assessing_2019} first try to use OpenIE tool \citep{yates2007textrunner}. However, OpenIE extracts triples with an unspecified schema instead of a fixed schema. In unspecific schema extraction, relation is extracted from the text between subject and object. In fixed schema extraction, a relation is predicted from a pre-defined relations set, which could be viewed as a classification task. Unspecific schema extraction makes the extracted triples hard to compare with each other. To resolve this problem, \citet{goodrich_assessing_2019} change to use relation extraction tools with fixed schema, which helps extracted triples easier to compare.


\subsection{Other Metrics}
Recently, there are some work evaluate faithfulness of text generation from other perspectives. 

\paragraph{BARTScore}
\citet{yuan2021bartscore} conceptualize the evaluation of generated text as a text generation problem, modeled using pre-trained sequence-to-sequence models, directly evaluating text through the lens of its probability of being generated from or generating other textual inputs and outputs. The general idea is that models trained to convert the generated text to/from a reference output or the source text will achieve higher scores when the generated text is better. 
They operationalize this idea using BART \citep{lewis2019bart}, an encoder-decoder based pre-trained model, and propose a metric BARTScore with a number of variants that can be flexibly applied in an unsupervised fashion to evaluation of text from different perspectives (e.g. informativeness, fluency,or factuality).
\begin{equation}
\begin{aligned}
    BARTScore = \sum_{t=1}^{m} w_t log p(\textbf{y}_t | \textbf{y}_{<t}, \textbf{x}, \theta)
\end{aligned}
\label{eq1}
\end{equation}

To evaluate faithfulness, they propose to compute the probability from source document to hypothesis $p(y|x,\theta)$. This direction measures how likely it is that the hypothesis $y$ could be generated based on the source text $x$.

\paragraph{TokenLevelCLS}

\citet{zhou_detecting_2021} propose a general-purpose method for token level hallucination detection for conditional sequence generation tasks. Given the source input $x$, they first formulate the task of token-level hallucination detection as a sequence labeling problem where a binary label is predicted at each position of the generated text. They train a model with synthetic training data in the form of $((x,y),L_y)$ where $L_y$ are the labels at every position of $y$ that indicate if each word is a hallucinated one or not. They leverage the BART model \citep{lewis2019bart} to mapping a corrupted sentence back to the original text it was derived from, without providing it any access to the source sentence, thereby encouraging it to insert new content as needed to ensure fluency. Then, they finetune a pre-trained language model (LM) on the synthetic data to help detect the token level hallucinations in various conditional sequence generation tasks.

\paragraph{CoCo}
CoCo \citep{xie_factual_2021} is proposed to evaluate the faithfulness of summarized texts via counterfactual estimation. They point out that the effect of language prior can be blamed to cause factual inconsistency. The intuition is that when texts are generated more relying on the source document rather than the language prior, they should be more likely to be faithful w.r.t. the source documents. They adopt the probabilities of the tokens of evaluated summaries to implement the automatic evaluation metric. Specifically, given the source document $x$ and model-generated summary $y$, several key tokens $y'$ are first selected from $y$, then the source document $x$ is masked according $y'$ to produce a masked version $x'$. $x$ and $x'$ are feed into the same scoring model respectively to generate the probability of each token in $y'$, i.e., $P(y_i|x,y_{<i})$ and $P(y_i|x',y_{<i})$,
$\forall y_i \in y'$. The CoCo value is defined as:
\begin{equation}
\begin{aligned}
    CoCo = \frac{1}{|y'|} \sum_{y_i \in y'} P(y_i|x,y_{<i}) - P(y_i|x',y_{<i})
\end{aligned}
\label{eq_coco}
\end{equation}

\paragraph{ShannonScore} \citet{egan_play_2021} proposed a reference-free metric ShannonScore to evaluate the quality of generated summary via Shannon Game. The Shannon information content of event $E$ with the probability $p(E)$ of happening is defined as $I(E) = -\log p(E)$. The ShannonScore performs the Shannon Game with a language model such as GPT-2 \citep{radford2019language}. The main assumption of this metric is that if $y$ is a satisfactory summary of $x$, then $I(x|y) < I(x)$, as documents that have little to do with the summary should be much less likely than documents that are relevant to the summary after conditioning the language model. Thus they define an Information Difference metric of summary quality as:
\begin{equation}
\begin{aligned}
    ID(x,y) = I(x) - I(x|y)
\end{aligned}
\label{eq1}
\end{equation}
They further define the ShannonScore as the normalized Information Difference:
\begin{equation}
\begin{aligned}
    s(x,y) = \frac{I(x) - I(x|y)}{I(x) - I(x|x)}
\end{aligned}
\label{eq1}
\end{equation}
where $I(x|x)$ is the lower bound of $I(x|y)$.

\subsection{Meta Evaluation}
As there are a lot of new metrics proposed to evaluate the faithfulness of language generation, they perform differently on different tasks.
To verify the effectiveness of the above metrics, these work usually report the correlations between their own metrics and human-annotated factual consistency scores.
However, it is difficult to compare each metric by the correlations as the diversity of annotating settings in different works and disagreement among different annotators.
To directly compare the effectiveness of different kinds of faithfulness metrics, several work \citep{gabriel_go_2021,koto_ffci_2021} proposed benchmarks to conduct meta-evaluations of faithfulness metrics.
For example, the popular benchmarks of evaluating the faithfulness metrics for abstractive summarization include FRANK \citep{pagnoni_understanding_2021}, SUMMAc \citep{laban_summac_2021}, QAGS \citep{DBLP:conf/acl/WangCL20}, FEQA \citep{durmus_feqa_2020} and CoCo \citep{xie_factual_2021}.
Table~\ref{meta_eval} combines these benchmarks, showing the Pearson correlations between different types of faithfulness evaluation metrics and human annotations.
To facilitate reliable evaluation metrics
for grounded dialogue generation,
\citet{DBLP:journals/corr/abs-2105-00071} also proposed a benchmark BEGIN for evaluation of grounded dialogue generation systems.

\begin{sidewaystable}
\renewcommand\arraystretch{1.5}
  \caption{Meta evaluations on several popular benchmarks. The results are Pearson correlations between different types of faithfulness evaluation metrics and human annotations.}
  \label{meta_eval}
  \centering
  \begin{tabular}{l|l|l|l|l|l|l|l|l|l|l}
    \toprule[1pt]
    \multicolumn{2}{c|}{Metrics}  &  \multicolumn{2}{c|}{FRANK benchmark}  &  \multicolumn{2}{c|}{QAGS benchmark}  &  \multicolumn{2}{c|}{FEQA benchmark} & \multicolumn{3}{c}{CoCo benchmark} \\
    \multicolumn{2}{c|}{}& CNN/DM & XSum & CNN/DM & XSum & CNN/DM & XSum & CNN/DM & XSum & SUMMEVAL \\
    \midrule[1pt]
    \multirow{6}{*}{N-gram based} & ROUGE-1 & 0.12 & 0.15 & 0.29 & 0.13 & 0.12 & -0.03 & 0.29 & 0.13 & 0.20 \\
    & ROUGE-2 & 0.08 & 0.17 & 0.18 & 0.09 & 0.13 & -0.06 & 0.18 & 0.09 & 0.17 \\
    & ROUGE-L & 0.11 & 0.16 & 0.24 & 0.09 & 0.13 & -0.06 & 0.23 & 0.08 & 0.19 \\
    & BLEU & 0.08 & 0.14 & 0.21 & 0.06 & 0.12 & -0.07 & 0.18 & 0.03 & 0.11 \\
    & METEOR & 0.12 & 0.15 & 0.27 & 0.10 & - & - & 0.26 & 0.11 & 0.17 \\
    & BERTScore & 0.02 & -0.04 & 0.28 & 0.03 & 0.11 & 0.10 & 0.37 & 0.11 & 0.19 \\
    \hline
    \multirow{3}{*}{Entailment-based} & ENT & - & - & - & - & 0.03 & -0.06 & - & - & - \\
    & DAE & 0.25 & 0.04 & - & - & & - & - & - & - \\
    & FactCC & 0.36 & 0.07 & - & - & - & - & - & - & - \\
    \hline
    \multirow{3}{*}{QA-based} & FEQA & -0.01 & 0.02 & - & - & 0.32 & 0.26 & - & - & - \\
    & QAGS & 0.13 & -0.02 & 0.55 & 0.17 & - & - & 0.31 & 0.15 & 0.18 \\
    & QuestEval & - & - & - & 0.33 & -  & - & 0.49 & 0.07 & 0.37 \\
    \hline
    \multirow{2}{*}{Others} & OpenIE & 0.16 & 0.00 & - & - & 0.09 & 0.02 & - & - & - \\
    & CoCo & - & - & - & - & - & - & 0.59 & 0.24 & 0.42 \\
    \bottomrule
  \end{tabular}
\end{sidewaystable}

Table~\ref{meta_eval} shows that the meta-evaluation results in different benchmarks differ greatly. The same metric can also performs very different on different datasets. 
Overall, the QA-based metrics achieve better performance on most datasets and benchmarks.
However, the correlations with human evaluations are not more than 0.6. Therefore, the faithfulness evaluation remains an open question of exploration.

\section{Optimization Methods}
\label{methods}

A lot of optimization methods for the faithfulness problem have been proposed for different tasks, including abstractive summarization, dialogue generation, data-to-text generation and machine translation.
However, most of these methods are general for different tasks. 
They can be categorized as factual guidance, auxiliary tasks, learning methods, post-editing, constrained decoding and others.
In the following, we will organize the optimization methods for each task into the above categories to facilitate comparing and learning of different methods, as shown in Figure~\ref{content_structure}.

\subsection{Faithfulness in Abstractive Summarization}
The faithfulness problem has attracted more and more attention in abstractive summarization. A lot of mitigation methods have been studied, many of which also can be applied to other NLG tasks.
In the following, we will introduce the main types of optimization methods for abstractive summarization.


\subsubsection{Factual Guidance}
\begin{figure}
	\centering
	\includegraphics[width=4in]{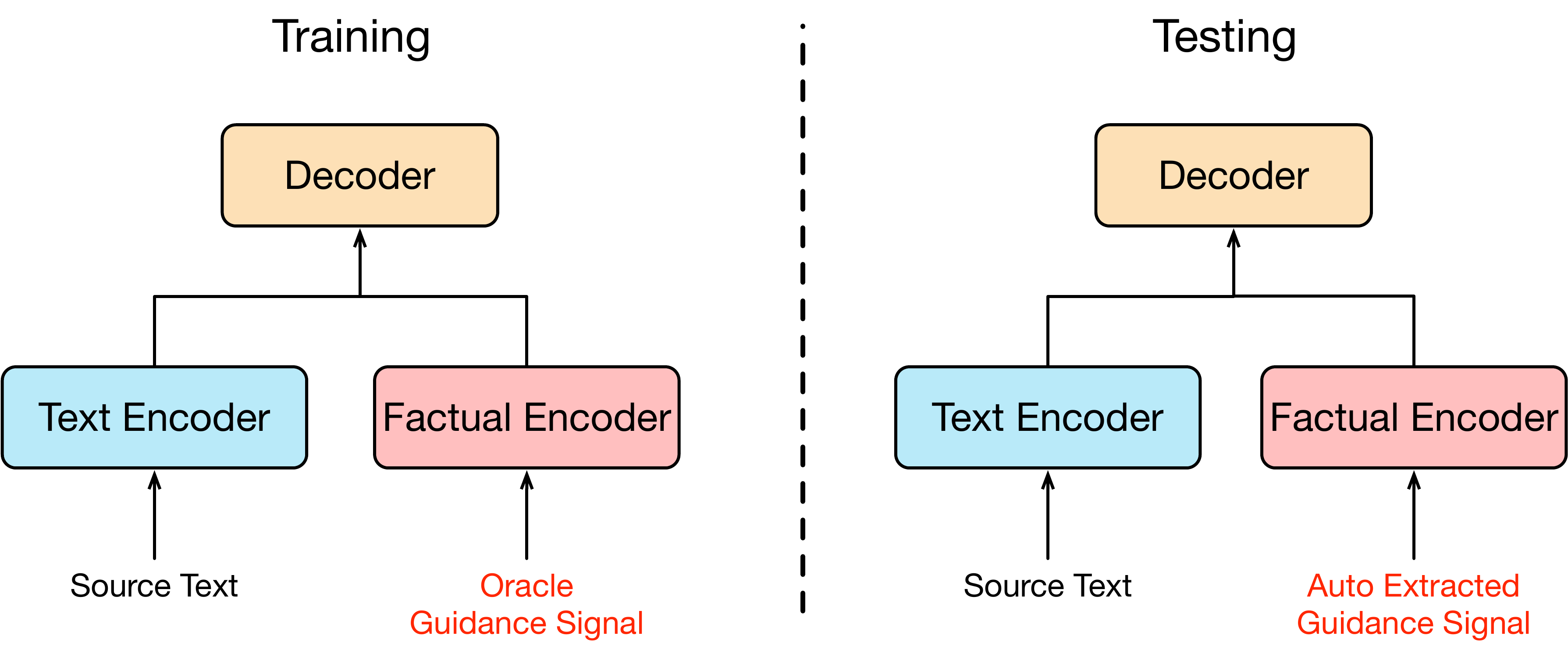}
	\caption{The framework of factual guidance. Usually, we use an oracle method to select guidance during training and use automatically extracted guidance at test time.}
	\label{factual_guidance}
\end{figure}

Factual guidance is an intuitive and effective method for boosting faithfulness and informativeness in summarization tasks. Guidance can be defined as some signals which are fed into the model as additional inputs to the source document. Within this framework, the crucial points are what kind of information we need to feed into the model and how to feed it.
Figure~\ref{factual_guidance} shows a simple seq2seq based factual guidance framework in which two encoders process origin source and extra guidance signals respectively, then a decoder generates final summaries considering the hidden states of both two encoders. Here, the guidance signals could be keywords, important sentences or other structures such as relations or semantic graphs. According to the types of guidance signals, factual encoders could be a Transformer network (for signal of sequence structure) or a Graph Attention Network (for signal of graph structure) \citet{velivckovic2017graph}. 
GSum \citet{dou2021gsum} is such a general and extensible framework that can take different kinds of external guidances as extra inputs to mitigate the unfaithful problems. We follow the basic classifications of guidance signals of GSum and then make an extension by supplying some different but effective ones. We divide guidance signals into three types: keywords, sentences and relations. 

\paragraph{Keyword Guidance}
Keywords reflect the crucial information of the source text in a simplest way. They help the summarization models to focus on the most important parts of the source text and result in less factual errors. \citet{li2018guiding} propose a Key Information Guide Network which encodes the keywords into the key information representation, to guide the process of generation. Firstly, they extract keywords from the text by using TextRank algorithm, then encode keywords by Bi-RNN network, and guide the generation process by cooperating keyword representations in both attention mechanism and the pointer mechanism. \citet{saito2020abstractive} further combine pre-trained seq2seq model with token-level saliency models called CIT, in which a saliency model (Transfomer encoder with feed-forward layer) produces a score for each token in order to select important ones which are denoted as $K$. Then a combined text $\hat{X} = concat(K, X)$ is given to the seq2seq model as the input. 

\paragraph{Sentence Guidance}
Keywords convey limited information of the source text, so some works turn to sentence-level guidance which contains more abundant information including keywords and the connections among them. \citet{cao2018retrieve} propose Re3Sum which retrieves existing summary sentences as candidate templates, and then uses an extended seq2seq framework to jointly conduct template reranking and template-aware summary generation. Specifically, both the source text $X$ and the soft template $R$ are converted into hidden states with a RNN encoder. In the Rerank module, they measure the saliency of $R$ according to its hidden state relevance to $X$. In the Rewrite module, a RNN decoder combines the hidden states of $X$ and $R$ to generate a summary $Y$. \citet{song_attractive_2020} exploit PORL-HG model following the extract-then-rewrite famework. PORL-HG firstly selects some attractive sentences from the article by an extractor, then rewrites these sentences by a seq2seq-based abstractor. The model combines the extractor with the abstractor by a reinforcement learning network which regards the popularity score and ROUGE scores as rewards to make sure generated headlines are both attractive and faithful.

\paragraph{Relation Guidance}
\citet{dou2021gsum} argue that if we utilize full sentence as guidance signals, it may contain much unnecessary and irrelevant information which is not crucial in a summary and could distract the model from focusing on the actual important parts of the source text. To address this problem, some works use relation information in the form of relational triples as factual guidance. \citet{cao_faithful_2017} leverage open information extraction and dependency parsing techniques to extract actual fact descriptions from the source text. They propose a dual-attention seq2seq framework to force the generation conditioned on both the source text and the extracted fact descriptions. \citet{huang2020knowledge} present ASGARD framework which enhances the regular document encoder with an independent graph-structured encoder which improves upon Graph Attention Networks \citep{velivckovic2017graph} to maintain the global context and local characteristics of entities. They utilize <subject, predicate, object> triples extracted by OpenIE to construct a knowledge graph, then use the hidden states of input tokens represented by RoBERTa \citep{liu2019roberta} to initialize the graph nodes. During decoding, both the representations of source tokens and graph nodes are incorporated into each generation step via cross-attention mechanism. In this way, the knowledge graph can be used as an extra factual guidance during summary generation.

Most works use OpenIE to extract relations from source documents, then represent them as graph structures to improve seq2seq models. However, these OpenIE-based graphs only contain sparse relations between partial words, which cannot cover the overall semantic meaning of the source article. \citet{wu2021bass} propose BASS which firstly introduce an unified semantic graph to enhance the performance of multi-document summarization. To construct the semantic graph, they extract phrases and their relations from sentences by a two-stage merging in which tokens are firstly merged into phrases based on dependency parsing trees, then co-referent phrases are merged into graph nodes according to co-reference chains. Finally, the model encodes graph structures both in encoding and decoding processes, by applying the graph adjacent matrix as self-attention mask and using an graph-propagate attention mechanism to guide the decoding process.


\subsubsection{Auxiliary Tasks}
\label{auxilary_tasks_summ}

\begin{figure}
	\centering
	\includegraphics[width=5in]{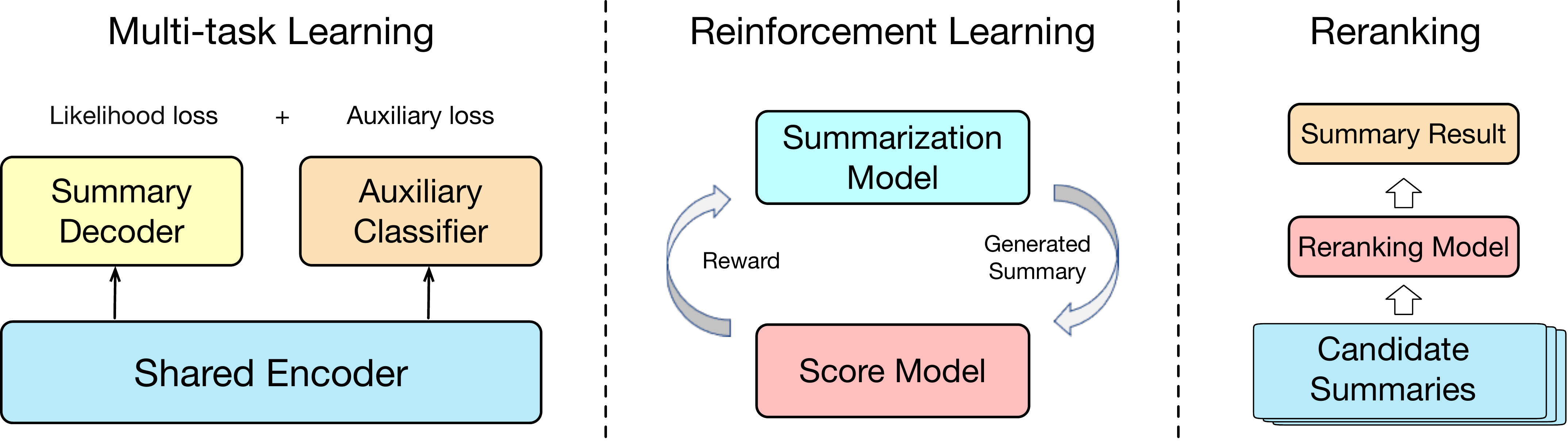}
	\caption{The framework of auxiliary task-based methods via multi-task learning, reinforcement learning and re-ranking.}
	\label{auxilary_framework}
\end{figure}

Unlike guidance methods which improve factual consistency explicitly, auxiliary task-based methods combine extra tasks which are correlative with factual correctness to boost the performance of summarization systems in an implicit way . There are three widely used frameworks that can easily involve auxiliary task into summarization task, which are reinforcement learning (RL) framework, multi-task learning framework and re-ranking framework, as shown in Figure~\ref{auxilary_framework}. In the RL framework, it is common to design a score model for generated summaries to obtain a reward which will optimize the factual consistency of summarization models. As for the multi-task framework, a task-specific layer will be stacked over the shared-weight encoder. 
In this way, the summarization model and auxiliary model share the same semantic representations but have different learning objectives. The related auxiliary task can be seen as a supplement to the summarization task and will improve the performance of the original summarization system. As for the re-ranking framework, it firstly generates several candidate summaries, then a score model based on auxiliary tasks produces a score for each candidate, and finally the best one is selected as the summary.
In the following, we will describe several common auxiliary tasks to improve faithfulness of abstractive summarization.

\paragraph{Entailment Task}
Natural Language Inference (NLI), in which a hypothesis sentence is classified as either entailed by, neutral or contradicting a premise sentence. Previous works \citep{li_ensure_2018, falke_ranking_2019, DBLP:journals/corr/abs-2005-11739, fabbri_answersumm_2021, laban_summac_2021} have proved that NLI tasks can improve faithfulness of summarization models. It can be incorporated into summarization models by multi-task learning, or acting as a RL reward, or utilized to re-ranking summary candidates.

\citet{li_ensure_2018} is the first work which incorporates entailment knowledge into abstractive summarization. They argue that a correct summary is semantic entailed by the source document. 
They propose an entailment-aware encoder under a multi-task learning framework, and an entailment-aware decoder under an RL framework with entailment rewards. In particular, they use shared weight encoders trained on both the summarization task (i.e. encoder+decoder) and the entailment task (i.e. encoder+classifier). Entailment prediction is regarded as an auxiliary task for summary generation. 
When decoding, they treat the entailment score as a special reward and combine the reward with a maximum likelihood training process by RL.

Following the idea that all information in a summary should be entailed by the source document, \citet{falke_ranking_2019} propose a re-ranking approach to select summaries with less unfaithful errors by entailment prediction models. They design a score function  mentioned in Equation \ref{RankNLI} to measure the entailment score of a generated summary $y$ given its source document $x$. The candidate summary with the highest score $\sigma(y)$ is selected as the model output after reranking.
\citet{DBLP:journals/corr/abs-2005-11739} follow this idea and make a further step by applying the adversarial NLI dataset to train the NLI model. More accurate NLI model has more potential of selecting faithful summaries.
\citet{fabbri_qafacteval_2021} propose query-based summarization model which apply NLI score
as one of the reinforcement learning rewards to improve factual coinsistency. 

\paragraph{Question answering Task}
Generating factual consistent summaries not only needs the overall understanding of source text but also the discrimination between crucial and useless parts. Thus, it is a natural way to check a summarization model's comprehension and distinction abilities by a QA model. The QA-based methods mainly calculate a QA score by measuring the overlap degree of answers extracted from source text and from the generated summaries, then use the QA score as the reward in the RL framework or the reranking framework. The key procedures of QA-based tasks are mentioned in the Section \ref{qestion_answering_task}. 
Following this idea, \citet{DBLP:conf/acl/NanSZNMNZWAX20} incorporate a QA model (Equation \ref{eq_quals}) into the seq2seq architecture by a novel contrastive learning method. They firstly produce some candidate summaries, then sort them into positive samples and negative samples according to the QA score, finally improve faithfulness of models through contrastive learning over them (specifically introduced in Section \ref{contrastive_learning_method}).

\paragraph{Other Tasks}
\citet{zhang_optimizing_2020} develop a concise framework to quantify factual correctness of a generated summary using an information extraction model. They take a structured vector $v$ to represent facts in the reference summaries. Each dimension of vector $v$ is a binary variable which describes whether an event or an entity is present or not in the text.
\begin{equation}
    v = f(y) = (v_1, \dots, v_m)
\end{equation}
Given the reference summary fact vector $v$ and generated summary fact vector $\hat{v}$, a factual accuracy score s can be computed as:
\begin{equation}
    s(\hat{v}, v) = \frac{\sum_{i=1}^m1[v_i=\hat{v_i}]}{m}
\end{equation}
Finally, they combine factual score with rouge score as reward via a reinforcement learning framework.

\citet{nan2021entity} propose a series of simple but effective entity-level methods to improve factual consistency of abstractive summarization, including data filtering, multi-task learning, and joint entity and summary generation. 
For data filtering, they first apply Spacy NER \citep{honnibal2017spacy} on reference summary to identify all named-entities. If any entities cannot find a match in the source document, they consider this sample as a noisy data and discard the sentence that contains the entity from the ground truth summary to make sure that there is no hallucination in the dataset. They also add a classification layer after the encoder of BART to identify summary-worth entities. As for decoding, they train the BART model to first generate the sequence of summary-worth entities and then the summary so that the salient named-entities can be incorporated into the cross-attention of the decoder.

\subsubsection{Learning Methods}
\begin{figure}
	\centering
	\includegraphics[width=5in]{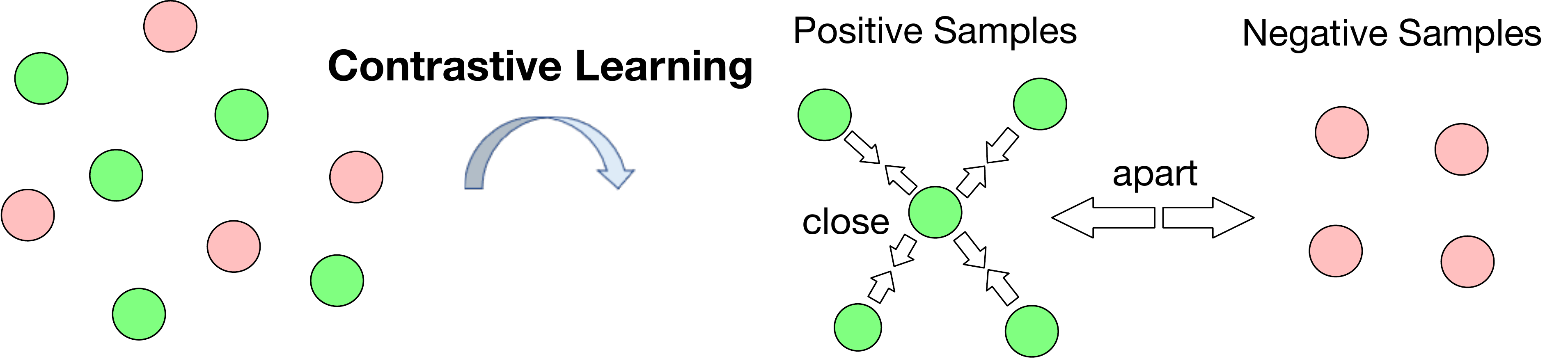}
	\caption{Contrastive learning framework.}
	\label{Contrastive_learning}
\end{figure}

\paragraph{Contrastive Learning}
\label{contrastive_learning_method}
\citet{cao_cliff_2021} observed that the commonly used maximum likelihood training method showed weak ability of distinguishing references from incorrect generations. Therefore, a potential solution is to design new learning objectives to improve the preference of factual summaries over inconsistent ones.
Contrastive learning(CL) is such a paradigm which is first proposed in visual tasks and recently utilized in many NLP tasks. The main idea of contrastive learning shown in Figure~\ref{Contrastive_learning} is to learn representations of similar samples staying close to each other, while dissimilar ones keeping away.
The key point of CL is how to generate positive and negative samples. In visual tasks, it is common to construct positive samples by rotating, resizing, distorting the origin picture, and consider other images as negative samples.
In this section, we will introduce several effective methods to construct positive and negative samples in summarization task, and how to involve them into the contrastive learning framework.

\citet{cao_cliff_2021} designe a task-specific contrastive learning formulation (CLIFF) that teaches a summarizer to expand the margin between factually consistent summaries and incorrect peers. 
CLIFF uses three methods to construct positive samples, including paraphrasing with synonym substitution, randomly replacing words, and back-translation. As for negative samples, previous works often treat other samples in the same batch as negative ones. However, \citet{cao_cliff_2021} argue that such negative samples are easy to distinguish because they are totally different from positive ones. It will be more effective to construct negative samples by making a small but crucial change based on the original references, so that the model can focus on the real important parts of the source text and enhance the ability of differentiating factual and non-factual summaries.
Following this idea, CLIFF designs four strategies to create negative samples:
\begin{itemize}
    \item \textbf{Entity swap imitates intrinsic errors:} swapping named entities in the references with other randomly selected entities of the same entity type in the source text.
    \item \textbf{Mask-and-fill with BART:} replacing each named entity in a reference with a [MASK], then let BART generates new entities.
    \item \textbf{Source-conditioned regeneration:} for each entity in the reference, feeding the text before it along with the origin source into BART, then combining the text before the entity with the generated text as a negative sample.
    \item \textbf{System generation:} selecting system generated summaries with low probability as negative samples.
\end{itemize}
After constructing positive samples (denoted as $P$) and negative samples (denoted as $N$), CLIFF optimizes the contrastive learning objective in Equation \ref{CL_loss} and combines it with typical cross-entropy loss to form the final training objective shown in Equation \ref{final_loss}, where $h_i$, $h_j$, $h_k$ are representations for summary $y_i$, $y_j$, $y_k$. $sim$ calculates the cosine similarity between summary representations. 
\begin{equation}
\begin{split}
    L_{CL} = -\frac{1}{\dbinom{\left|P\right|}{2}}\sum_{y_i,y_j \in P, y_i \neq y_j} log \frac{exp(sim(h_i, h_j) / \tau)}{\sum_{y_k \in P \cup N, y_k \neq y_i} exp(sim(h_i, h_k) / \tau)}
    \label{CL_loss}
\end{split}
\end{equation}

\begin{equation}
    L = L_{CE} + \lambda L_{CL}
    \label{final_loss}
\end{equation}

Except for entity-level replacement, \citet{liu_improving_2021} make a further step by switching the sentiment of some sentences by adding negation words or replacing opposite meaning words to generate more diverse negative samples.

\citet{liu2021co2sum} argue that previous works \citep{cao_cliff_2021, liu_improving_2021} mainly focus on entity faithfulness which is not equal to summary faithfulness. Thus, they propose a contrastive summarization framework CO2Sum and a span-level negative samples  construction method LFN based on pre-trained language model. Specifically, they delete or disturb factual fragments in sentences and observe the language model probability of predicting the context based on these sentences in an iterative way to distinguish which fragments are important to the source text. After detecting most influential factual spans, they replace the fragment in the gold summary with embedding-similar article words to construct negative samples. They involve contrastive learning in both encoder and decoder. In the encoding procedure, they apply contrastive learning between source text and summaries by making the representations of the article and the ground truth summary closer, and make that of the article and the factual inconsistent summaries apart. The CL loss $L_{Enc}$ for the encoder is similar to \citet{cao_cliff_2021}. As for the decoder, CO2Sum applies contrastive learning between summaries with a max-margin loss \citep{yang2019reducing} $L_{Dec}$ to force the model to increase the decoding probabilities of ground truth summaries while decrease the decoding probabilities of negative summaries. The margin loss $L_{Dec}$ and the final training objective $L$ are shown as following.
\begin{equation}
    L_{Dec} = max\left\{\frac{1}{R} \sum_{i \in R} (P_s(T_{neg}, i) - P_s(T_{gold}, i)) + \eta, 0 \right\}
\end{equation}
\begin{equation}
    L = L_{CE} + \lambda_{Enc} L_{Enc} + \lambda_{Dec} L_{Dec}
\end{equation}
\noindent where $R$ means replaced positions with inconsistent facts, $T_{neg}$ and $T_{neg}$ denote the negative summary and ground truth summary respectively, $P_s(T, i)$ denotes the generation probability of the $i$-th position in sequence $T$. $\lambda_{Enc}$ and $\lambda_{Dec}$ denote the loss weights of the constrastive learning objectives in the encoder side and decoder side, respectively.

Most works construct negative samples by simply replacing some non-target sequences. \citet{lee_contrastive_2021} argue that these explicit negative samples are suboptimal, since they are easily distinguishable from the correct output, especially when models are pre-trained with large corpus. Within the simple and explicit sample construction framework, models barely learn nothing. 
Thus, they propose a principled method called CLAPS to construct positive and negative samples implicitly by adding perturbations to the input sequence.  To generate a negative example, they add a small perturbation (hard sample) to the hidden representation of target sequence, then minimize its conditional likelihood. As for positive examples, they adding a large perturbations while enforcing the model to have a high conditional likelihood. 
This will yield a negative example that is very close to the original representation of target sequence in the embedding space but is largely dissimilar in the semantics, while the generated positive example is far away from the original input sequence but has the same semantic as the target sequence. It could generate hard examples which the model might be difficult to discriminate, helping it learn more meaningful representations.

\subsubsection{Post-Editing}
\begin{figure}
	\centering
	\includegraphics[width=4.5in]{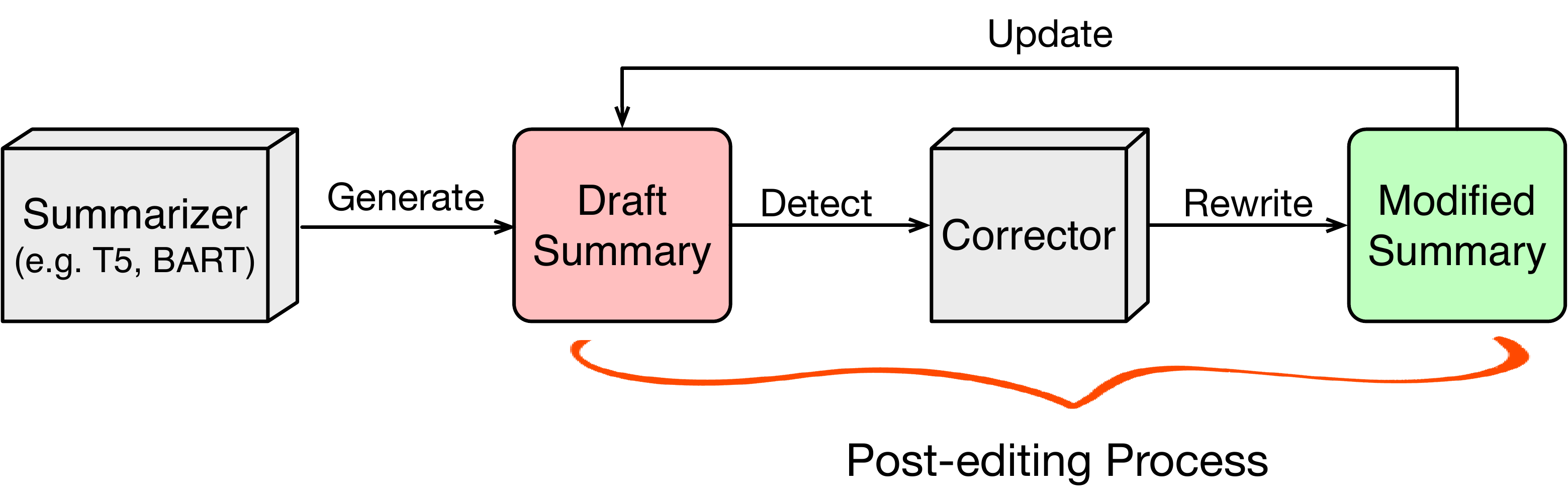}
	\caption{The post-editing framework.}
	\label{Post-editing}
\end{figure}

Above methods require modification of model structures or extra sample construction processes to improve factual consistency, which may affect the informativeness (e.g. ROUGE scores) of summary results.
Post-editing based methods improve factual consistency by adding an corrector to system-generated summaries. They consider generated summaries as drafts, and correct factual errors to form the final summaries. This process is quite similar to the human writing process, where people write a first draft, then review and edit it to make it better. Figure~\ref{Post-editing} shows the general framework of post-editing methods.

\citet{dong2020multi} propose SpanFact, a suite of two factual correction models that leverage knowledge learned from question answering models to correct system-generated summaries through span selection and correction. SpanFact takes into account entity-level corrections and make them iteratively. Specifically, assume that the system summary has $N$ entities. At time step $i$, they mask the $i$-th entity and use this masked sequence as a query to the QA model. The QA model will replace the wrong entities with the correct ones based on the source document. The corrected entity will then form an updated summary for use in the next step.
Human evaluation demonstrates that SpanFact is able to correct about 26\% unfaithful summaries, while barely destroying any otherwise correct summaries.
\citet{cao2020factual} simplify the post-editing procedures by directly training a seq2seq rewrite model on artificial unfaithful summaries as a corrector. They create a weakly-supervised training dataset based on the text transformations following \citet{kryscinski_evaluating_2019} which replace entities, numbers, numerals and pronouns in source documents with other tokens of the same type. The goal of the corrector is to generate correct summaries based on the unfaithful summaries and source documents.

As a standalone module, post-editing methods have been shown to be effective in improving the faithfulness of abstractive summarization systems while preserving their informativeness. However, it's more of an indirect solution than a fundamental solution to factual inconsistencies.

\subsubsection{Constrained Decoding}
Lexically constrained or guided decoding is a modification of beam search that enforces the inclusion of pre-specified words and phrases in the output. This is a general way to control specific tokens in the generated output without modifying the model structure or additional training data.

\citet{mao2020constrained} propose CAS (Constrained Abstractive Summarization) to improve the factual consistency of summarization systems by constructing constrained token sets during dynamic beam search decoding. It only allows the generation process to end when all constraints are met. They focus on entities and noun phrases and select these types of words that are not present in the summaries generated by the unconstrained system to form constrained sets. Therefore, the model will generate more correct and faithful tokens during the inference process, effectively improving the faithfulness of abstractive summarization.
\citet{DBLP:conf/acl/AralikatteNMRM20} introduce the Foucs Attention Mechanism (FAME) for the transformer-based seq2seq architecture. FAME combines a standard contextual representation with a dynamic source-conditioned lexical bias layer, which encourages the decoder to actively generate tokens that are faithful to the input document.

\subsubsection{Other Methods}
\citet{zhao2020reducing} propose HERMAN which learns to recognize and verify quantity entities in candidate summaries, in order to re-rank the candidate summaries to select the one whose quantity terms are supported by the original text. During the training process, they use a BiLSTM-CRF decoder as a verification model to tag sequence labels and finally predict an overall label that indicates whether the output summary is faithful to the source input or not. At the test time, the same verification model is applied to rerank the candidate summaries, then select the best one with less hallucinations.

\citet{gabriel2021discourse} proposed Co-opNet, a generator-discriminator framework to do fact-checking for text generation. In this framework, the generator outputs a series of candidate summaries. Then the discriminator scores the factuality of these summaries using one of the following objectives: the overlap between the introduction of a scientific article and the predicted evidence spans in summaries, the ordering of predicted discourse roles, the coverage of predicted discourse roles, or the likelilood of adjacency between generated sentences. The best summary is selected by combining the scores of the generator and the discriminator.

\citet{cao_inspecting_2021} propose an interesting method to detect factual errors by using the prior and posterior predicted probabilities of each token. They assume that if an entity is a factual error, giving the source should not provide more evidence for it, resulting in only small changes in the probabilities between the prior (i.e. without source) and the posterior (i.e. given source) language models. Based on this assumption, they use prior and posterior probabilities as key features of a classifier to predict the factuality of entities.

\subsection{Faithfulness in Dialogue Generation}

Recently, the area of dialogue generation has made significant progress with end-to-end neural networks and large-scale pre-training \citep{DBLP:conf/eacl/RollerDGJWLXOSB21, DBLP:conf/acl/BaoHWWW20}.
However, a long standing problem, faithfulness,  still challenges current best dialog systems and attracts an increasing amount of attention.  In general, the generated utterance  should be faithful to its history utterances \citep{DBLP:journals/corr/VinyalsL15}.
Different from other generation tasks like text summarization, various forms of dialog generation tasks include  a diversity of background or knowledge inputs, with which the generated utterances should also be consist. 
In Table \ref{tab:consistency_type}, we summarize different forms of inputs that have been studied in dialogue faithfulness.
The optimization methods for dialogue faithfulness are similar to abstractive summarization, which consists of six types of methods. Some of them can also be utilized in summarization tasks.

\begin{table}
\renewcommand\arraystretch{1.5}
  \caption{Different types of source that dialogue generation models should be faithful to in different tasks.}
  \label{tab:consistency_type}
  \centering
  \begin{tabular}{l|l}
    \toprule[1pt]
    \textbf{Source Type}     &      \textbf{Methods}  \\
    \midrule[1pt]
    \multirow{6}{*}{History Dialogue} & DialogNLI \citep{DBLP:conf/acl/WelleckWSC19},    \quad \citet{DBLP:conf/emnlp/GaoZLGBGD19}\\
    & \citet{DBLP:conf/coling/ArunBBCDHIJKMMW20}, \quad  \citet{DBLP:conf/aaai/GhazvininejadBC18} \\
    & DECODE \citep{DBLP:conf/acl/NieWBKW20}, \quad CI-ToD \citep{DBLP:conf/emnlp/QinXHCXC21}\\
    &TransferTransfo \citep{mesgar_improving_2021}, \quad  UL \citep{DBLP:conf/acl/LiRKWBCW20}\\
    &Blender \citep{DBLP:conf/eacl/RollerDGJWLXOSB21},\quad \citet{DBLP:conf/acl/BalakrishnanRUW19}\\
    &\citet{nye2021improving}, \quad \citet{DBLP:conf/emnlp/KimKK20}\\
    
    \midrule[1pt]
    
    \multirow{5}{*}{Persona Facts}& \citet{DBLP:conf/acl/LiGBSGD16},\quad \citet{DBLP:conf/acl/KielaWZDUS18}\\

    & DialogNLI \citep{DBLP:conf/acl/WelleckWSC19}, \quad DECODE \citep{DBLP:conf/acl/NieWBKW20}\\
    &KvBERT \citep{song_profile_2020}, \quad RCDG \citep{song_generating_2020}\\
    & TransferTransfo \citep{mesgar_improving_2021}, \quad UL \citep{DBLP:conf/acl/LiRKWBCW20}\\
    & GDR \citep{DBLP:conf/acl/SongWZLL20}, \quad \citet{DBLP:conf/emnlp/KimKK20}\\
    \midrule[1pt]

    Unstructured Knowledge   & \citet{DBLP:conf/acl/RashkinRT020},\quad \citet{DBLP:conf/aaai/WuGBZ0QKGHOD21}\\
    
    (e.g. Wikipedia Documents)& \citet{DBLP:conf/iclr/DinanRSFAW19}, \quad \citet{DBLP:conf/emnlp/0001PCKW21}\\ 
    \midrule[1pt]
    Structured Knowledge & KvBERT \citep{song_profile_2020},  \quad CI-ToD \citep{DBLP:conf/emnlp/QinXHCXC21}\\
    (e.g. Knowledge Graph) & NPH \citep{dziri_neural_2021}\\
    \midrule[1pt]
    User Query & CI-ToD \citep{DBLP:conf/emnlp/QinXHCXC21}\\
    (i.e. task-oriented dialogue) & \\
    \bottomrule
  \end{tabular}
\end{table}

\subsubsection{Factual Guidance}
\begin{figure}
	\centering
	\includegraphics[width=4in]{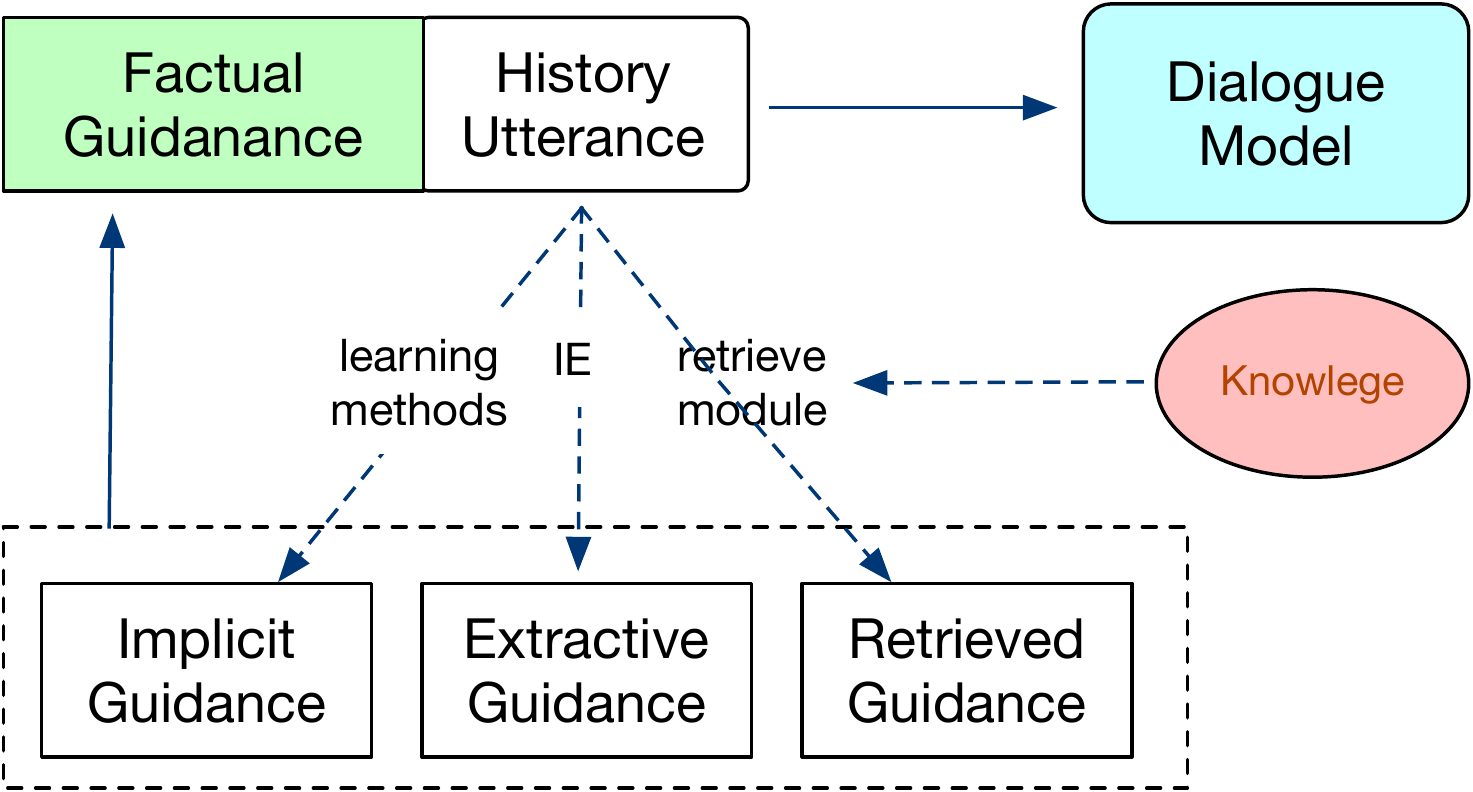}
	\caption{The framework of three types of factual guidance in dialogue generation.}
	\label{fact_guidance}
\end{figure}

Several works utilized various  guidance information to improve factual consistency of dialogues.
These methods incorporate relevant guidance information into the training or inference process of dialogue models.
As shown in Figure \ref{factual_guidance}, we categorize theses guidance into three types: implicit guidance, which is the guidance in vector representations; extracted guidance, which is the relevant textual information extracted from source inputs; retrieved guidance, which is information retrieved from open-domain knowledge.

\paragraph{Implicit Guidance}
Implicit guidance is usually  representations that are automatically learned before or during the training process of a dialogue model.
\citet{DBLP:conf/acl/LiGBSGD16} inject implicit speaker information into a LSTM-based model to improve the personality consistency.
In each generation step of their model, they fuse the embedding of  the speaker into the text encoder.
 \citet{DBLP:conf/acl/KielaWZDUS18} present the PERSONA-CHAT dataset which provides persona profiles for each speaker. 
 They also propose a memory-augmented dialogue system where persona profiles were saved and updated in memory. 
 Guided by the profile memory, the generated dialogues are more consistent in personality.
 \citet{DBLP:conf/emnlp/GaoZLGBGD19} combine dialog generation with style-transfer for a more stylized and context-relevant chatbot. 
 They fuse conversation modeling and non-parallel style transfer method by sharing a structured latent space to guide the decoding process.

\paragraph{Extractive Guidance}
Extractive guidance is often the important information extracted from the source input, which helps the model focus on the important parts of the input.
\citet{DBLP:conf/aaai/GhazvininejadBC18} propose a knowledge-grounded conversation model, which extracts factual sentences from history dialog and utilizes them as factual guidance in the decoding process. 
\citet{DBLP:conf/coling/ArunBBCDHIJKMMW20} extract tree-based meaning representations to improve the faithfulness of generated responses for task-oriented dialog systems.
The extract structural knowledge efficiently guided the model to generate correct information.
\citet{DBLP:conf/acl/RashkinRT020} utilize control codes to encourage the model to generate responses that are faithful to the provided evidence.  
They apply three types of control codes, including entailment, objective voice and lexical precision, which are calculated during data pre-processing. 
\citet{DBLP:conf/aaai/WuGBZ0QKGHOD21} further construct fine-grained control codes by using lexical phrases as factual guidance. 
Based on these phrases, the generated responses are more relevant and faithful to the input. 

\paragraph{Retrieved Guidance}
Retrieved guidance is usually from external knowledge.
\citet{DBLP:conf/iclr/DinanRSFAW19} create an open-domain dialogue dataset Wow, where each topic in the conversation is connected to Wikipedia articles.
Then they design a memory network based dialog system which is enhanced by retrieved knowledge from Wikipedia. 
Augmented by knowledge guidance, their model is able to generate more precise responses.
\citet{DBLP:conf/emnlp/0001PCKW21}  propose a retrieval-augmented neural architectures, in which dialogues are generated grounded on retrieved knowledge. 
Specially, they apply a learnable retriever and designed a fine-grained interactions between history dialogue and knowledge.

\subsubsection{Auxiliary Tasks}
Many work utilize the auxiliary task of Natural Language Inference (NLI) to improve the factual consistency of dialogue systems.
The main approaches include leveraging entailment scores to rerank candidate texts, or treating entailment as a reward for reinforcement learning, which are similar to the entailment-based methods for text summarization (described in Section~\ref{auxilary_tasks_summ}).

\paragraph{Reranking-based}
Several works apply entailment score predicted by an NLI model to the  reranking process for selecting more faithful generated text.
As discussed in Section \ref{metric_entail}, several works \citep{DBLP:conf/acl/WelleckWSC19, song_profile_2020, DBLP:conf/emnlp/QinXHCXC21, DBLP:conf/acl/NieWBKW20} propose  their factual evaluation metrics based on entailment.
They utilize entailment scores predicted by their proposed metrics in the re-ranking process to improve faithfulness of dialogue models.
For example, in \citet{DBLP:conf/acl/WelleckWSC19},  given a persona $P$, previous utterances $u_{<=t}$, and the dialogue model outputs the score of a  next-utterance candidate $s^i_{t+1}$, the new score $s_{t+1}^{re-rank}$ after incorporating NLI relation is:
\begin{equation}
    s_{t+1}^{re-rank} = s_{t+1} + \lambda s_{t+1}^{contradict}
\end{equation}
where $s_{t+1}^{contradict}$ is the highest contradiction score between $s^i_{t+1}$ and persona sentences in  $P$, hyper-parameter $\lambda$  controls the NLI model’s influence in re-ranking.

\paragraph{Reinforcement Learning} Another type of methods incorporate entailment scores as a part of the reinforcement learning (RL) rewards, similar to Figure \ref{auxilary_framework}.
\citet{song_generating_2020} propose a RL-based model RCDG for generating persona consistent dialogues. 
Similar to the architecture of GANs (Generation Adversarial Neural Networks), RCG is composed of a generator and two evaluators to estimate the quality and consistency of generated utterances, respectively.
The consistency evaluation is based on an NLI classifier to compute the entailment score.
\citet{mesgar_improving_2021} also propose an RL-based model TransferTransfo-RL for improving consistency between generated responses and personas.
Differently, TransferTransfo-RL take the advantage of Actor-Critic \citep{DBLP:conf/icml/MnihBMGLHSK16} learning approach, which also utilizes the entailment score as reward.

\subsubsection{Learning Methods}
As unfaithful generation relates to the deficiencies of  training strategy, several works improve  factual consistency of dialogue models by refining the training procedures.

\citet{DBLP:conf/acl/LiRKWBCW20} extend unlikelihood training \cite{} to address various problems in generating dialogues, including over copying, repetitions, overuse frequent words, and factual inconsistency. Besides training with common maximum likelihood estimation (MLE), they apply unlikelihood loss (UL) to alleviate these problems.
In the time step $i$ during training, given an input-output pair $(x,y)$,  a dialogue model $p_{\theta}$ and a set $\mathcal{C}$ containing sentences contradicting with $y$ which the model should avoid to generate, the  UL is defined as: 
\begin{equation}
    \mathcal{L}_{UL}(p_\theta,\mathcal{C},x,y) = -\sum_{t=1}^{T}  \sum_{y_c\in{C}}\beta(y_c)(1-p_{\theta}(y_c|x,y_{<t}))
\end{equation}
where $\beta(y_c)$ is the weighting parameter for every $y_c\in{C}$.
\citet{DBLP:conf/eacl/RollerDGJWLXOSB21} followed this method for training a large-scale chatbot.

\subsubsection{Constrained Decoding}
Some works also focus on designing decoding strategies to improve consistency. 
Especially, these works apply constrained decoding during inference.
These methods usually require the generated utterance to be semantically consistent with inputs based on certain semantic structure.

\citet{DBLP:conf/acl/BalakrishnanRUW19} apply a tree-structured meaning representations (MR) in dialog systems.
Comparing to common flat MR, which is a ﬂat list of key-value pairs, their MR is able to represent more fine-grained relations. 
Based on the proposed MR, they design a constrained decoding strategy on beam search which requires the MR of generated text  not conflicting with the input. 
Similarly, \citet{nye2021improving} propose a dual-system approach for faithful text generation, where the ``system  1'' is a common model for generation, and the ``system 2'' constrains and controls the generated sentences to be factually correct. 
The essence of this method is that it applies GPT-3 \citep{brown2020language} to parse text into clear and correct logical symbols that are easy for ``system  2'' to check. 
During decoding, ``system  2'' selects correct candidates that are faithful to the given context. 

\subsubsection{Post-Editing}
Several works focus on refining the generated dialogues without modifying the original model.
These works mainly design an extra refining module to correct the factual errors in the generated dialogues.
These models usually consist of three steps: generate, delete, and rewrite, similar to the summarization task shown in Figure~\ref{Post-editing}. In the first step, the dialogue model normally generates utterances. In the second step, the rewrite module removes the incorrect contexts in the generated utterances, and then the third step rewrites them to the correct contexts.

\citet{DBLP:conf/acl/SongWZLL20} propose a  post-editing  based dialogue model, GDR, following the three steps introduced above.
After the normal text generation procedure ``Generate'' in the first stage, GDR identifies and deletes conflict words in the second stage ``Delete''. 
Then GDR recovers the deleted words by a generation module in the last stage ``Rewrite''.
Through the three stages above, GDR refines factual errors in the generated utterance. 
\citet{dziri_neural_2021} apply the similar strategy on knowledge grounded dialogue system. 
Different from GDR, their refining module NPH needs to rewrite based on knowledge graph. 
After deleting potential incorrect entities in the generated text, NPH based on graph neural network retrieves correct entities in the grounded knowledge graph for refining.

\subsubsection{Other Methods}
\citet{DBLP:conf/emnlp/KimKK20} propose a dialogue system based on Rational Speech Act framework \citep{frank2012predicting}, which enforces dialogue agents to refrain from uttering contradiction.
The proposed model endows dialogue agents with public self-consciousness, helping them maintain consistency across each generation step by reflecting the distribution of imagined listeners across roles.

\subsection{Faithfulness in Machine Translation}
Neural machine translation (NMT) has achieved great success due to the ability to generate high-quality sentences. Compared with human translations, one of the drawbacks of current NMT is that translations are not usually faithful to the input, e.g., omitting information or generating unrelated fragments, which inevitably decreases the overall quality, especially for human readers.
The optimization methods for faithfulness in machine translation mainly include: incorporating auxiliary tasks like word alignment (\ref{auxiliary_tasks}), improving learning methods like minimum risk training (\ref{learning_methods}), utilizing constrained decoding methods like grid beam search (\ref{constrained_decoding}), etc.

\subsubsection{Auxiliary Tasks}
\label{auxiliary_tasks}
\citet{wang2020FaithfulNeural} propose a multi-task learning paradigm with two auxilary tasks, including marked language model task and word alignment task, for building a faithfulness enhanced NMT (named FENMT). On the encoder side, FENMT employs a masked language model (MLM) task \citep{devlin2018bert} to infer the input words didn’t be correctly translated. This task can enhance the ability of modeling the whole input sentence and give the decoder accurate and complete representations. On the decoder side, FENMT further uses a word alignment task to improve the alignment accuracy of the encoder decoder cross-attention to help the decoder to capture correct contextual representation. Furthermore, along with the NMT objective, an auxiliary max-margin objective based on contrastive learning is introduced in all decoding timesteps which prompts the decoder to translate fluent and faithful sentences.

To improve the ability of the decoder, \citet{tu2017neural} propose to introduce a reconstruction loss to make translation can reconstruct the input sentence. \citet{kong2019neural} propose to use a coverage difference ratio metric as a reward to train NMT. \citet{zhang2021neural,feng2020modeling,garg2019jointly} propose to introduce word alignment information in Transformer to improve translation accuracy.

\subsubsection{Learning Methods}
\label{learning_methods}

\paragraph{Minimum Risk Training}
\citet{wang_exposure_2020} hypothesise that exposure bias \citep{ranzato2015sequence}, a discrepancy between training and inference, is partially to blame for hallucinations, and that training with Minimum Risk Training, which avoids exposure bias, can mitigate this. Minimum Risk Training (MRT) is a sequence level objective that avoids this problem. Specifically, the objective function of MRT is the expected loss (risk) with respect to the posterior distribution:
\begin{equation}
\begin{aligned}
    \mathcal{R(\theta)} = \sum_{(x,y) \in D} \sum_{\widetilde{y} \in Y(x)} P(\widetilde{y}|x;\theta) \triangle(\widetilde{y}; y)
\end{aligned}
\label{eq1}
\end{equation}
\noindent in which the loss $\triangle(\widetilde{y}; y)$ indicates the discrepancy between the gold translation $y$ and the model prediction $\widetilde{y}$. Due to the intractable search space, the posterior distribution $Y(x)$ is approximated by a subspace $S(x)$ by sampling a certain number of candidate translations, and normalizing:
\begin{equation}
\begin{aligned}
    \widetilde{P}(\widetilde{y}|x;\theta,\alpha) = \frac{P(\widetilde{y}|x;\theta)^\alpha}{\sum_{y' \in \mathcal{S}(x)} P(y'|x;\theta)^\alpha}
\end{aligned}
\label{eq1}
\end{equation}
where $\alpha$ is a hyper-parameter to control the sharpness of the subspace. Random sampling was used to generate candidate translations, and the reference translation was not added to the subspace.
They find that Minimum Risk Training, which does not suffer from exposure bias, reduces the number of hallucinations substantially, and makes beam search with large beams more stable.

\paragraph{Adversarial Learning}
NMT models are still susceptible to input sentence perturbations and tend to produce hallucinatory outputs in the presence of some source perturbations \citep{lee2018hallucinations}.
For example, \citet{belinkov2017synthetic} find that NMT models can be immensely brittle to small perturbations applied to the inputs. Even if these perturbations are not strong enough to alter the meaning of an input sentence, they can nevertheless result in different and often incorrect translations. \citet{belinkov2017synthetic} and \citet{karpukhin2019training} study how to use some synthetic noise and/or natural noise. \citet{cheng2018towards} propose adversarial stability training to improve the robustness on arbitrary noise type including feature-level and word-level noise. \citet{liu2018robust} examine the homophonic noise for Chinese translation.

Formally, a set of adversarial examples $Z(x; y)$ is generated with respect to a training sample $(x; y)$ by solving an optimization problem:
\begin{equation}
\begin{aligned}
    \{ x' | \mathcal{R}(x', x) \le \epsilon, argmax_{x'} J(x',y;\theta) \}
\end{aligned}
\label{eq1}
\end{equation}
\noindent where $J(.)$ measures the possibility of a sample being adversarial, and $R(x'; x)$ captures the degree of imperceptibility for a perturbation. Although it is difficult to give a precise definition of the degree of imperceptibility $R(x'; x)$, $l_{\infty}$ is usually used to bound the perturbations in image classification \citep{goodfellow2014explaining}.

Following \citet{goodfellow2014explaining, miyato2016adversarial, ebrahimi2017hotflip}, the white-box method to generate adversarial example stightly guided by the training loss. Given a parallel sentence pair (x; y), a set of adversarial examples A(x; y) specific to the NMT model are generated by:
\begin{equation}
\begin{aligned}
    \{ x' | \mathcal{R}(x', x) \le \epsilon, argmax_{x'} - \log P(y|x';\theta) \}
\end{aligned}
\label{eq1}
\end{equation}
\noindent where they use the negative log translation probability to estimate $J(.)$. The formula constructs adversarial examples that are expected to distort the current prediction and retain semantic similarity bounded by $\mathcal{R}$.

\citet{cheng_robust_2019} propose a gradient-based adversarial learning approach, called AdvGen, to construct adversarial examples and use these examples to both attack as well as defend the NMT model to improving the robustness of NMT models, which consists of two parts: (1) attack the translation model with adversarial source examples; (2) defend the translation model with adversarial target inputs to improve its robustness against the adversarial source inputs. For the generation of adversarial inputs, they propose a gradient-based method to craft adversarial examples informed by the translation loss over the clean inputs.

\paragraph{Robust Learning on Noisy Corpus}
Corpus-level noise in NMT parallel corpora tends to produce significant hallucinatory patterns \citep{raunak2021curious}.
NMT models also have a propensity to hallucinate more frequently under out-of-domain inputs \citep{muller2019domain}.
Several techniques can be used to improve learning robustness to corpus-level noise in NMT.
\citet{kang2020improved} propose a loss truncation method to reduce the impact of noisy references in sequence-to-sequence training. \citet{li2020tilted} propose a modification of expected risk minimization (ERM), namely Tilted-ERM, to reduce the effect of outliers during training. 
Corpus-level noise filtering that incorporating heuristics or filters \citep{zhang2020parallel,junczys2018dual} to remove invalid source-target pairs is also effective in reducing NMT hallucinations.

\subsubsection{Constrained Decoding}
\label{constrained_decoding}

One interesting method is lexically constrained decoding, a modification to beam search that allows the user to specify words and phrases that must appear in the system output. Three algorithms have been proposed for this: grid beam search \citep{hokamp_lexically_2017}, constrained beam search \citep{anderson_guided_2017} and dynamic beam allocation \citep{post_fast_2018}. These papers showed that these algorithms do a good job automatically placing constraints and improving results in tasks such as simulated post-editing, domain adaptation, and caption generation.

Grid Beam Search (GBS), an algorithm which extends beam search to allow the inclusion of pre-specified lexical constraints while still taking advantage of the distribution learned from training data. The algorithm can be used with any model that generates a sequence $\widehat{y}={y_0, \ldots, y_T}$, by maximizing $p(y|x)=\prod_{t} p(y_t | x; {y_0, \ldots, y_{t-1}})$. Lexical constraints take the form of phrases or words that must be present in the output sequence. This is a very general way to incorporate additional knowledge into a model’s output without requiring any modification of the model parameters or training data, thus can improve faithfulness of translation output.

The computational complexities of grid beam search \citep{hokamp_lexically_2017} are linear and constrained beam search are exponential in the number of constraints. \citet{post_fast_2018} present a more efficient algorithm for lexically constrained decoding with a complexity of $O(1)$ in the number of constraints.

\subsubsection{Other Methods}
Some work design special mechanism for using source representation more effectively. In the RNN-based NMT, \citet{tu2016modeling} and \citet{mi2016coverage} propose a coverage mechanism to improve the accuracy of translation outputs. \citet{weng2020gret} propose to model global representation in the source side to improve the source representation. \citet{zheng2019dynamic} propose a capsule based module to control the source representation dynamically in the decoding process. \citet{feng2020modeling} propose a faithfulness part to optimize the contextual representation before feeding into the decoder. 
\citet{weng2017neural} propose a bag-of-words loss to constrain decoding process.

\subsection{Faithfulness in Data-to-Text Generation}

Data-to-text generation (or table-to-text generation) has been widely studied for decades. Recently, deep neural networks have been successfully adopted in this task. However, the problem of unfaithfulness in data-to-text methods remains a significant challenge (especially in long-form text generation). In the data-to-document generation dataset (e.g., Rotowire \citep{wiseman2017ChallengesDatatoDocument} and MLB \citep{puduppully2019DatatotextGeneration}), obvious gaps exist in the record generation (RG) precision between recent methods and faithful templated-based baselines. Moreover, the issue of generating unfaithful text remains a serious problem for noisy datasets constructed automatically (e.g., WikiBio \citep{lebret2016NeuralText} and WikiPerson \citep{wang2018DescribingKnowledge}). In WikiBio, almost two-thirds of the training instances contain unfaithful descriptions \citep{dhingra_handling_2019}.

\subsubsection{Factual Guidance}
Compared to other text generation tasks, factual information is highly structured and exists explicitly in data-to-text tasks. Therefore the two-stage method, which plans the subset of input data to be described and then generates text from the plan, is popular in this research field. If a method explicitly considers the issue of faithfulness in the plan-to-text generation step, we categorize it into factual guidance optimization methods.

\citet{wang2021SketchRefine} propose a two-stage table-to-text generation method SANA. SANA firstly constructs a skeleton using an autoregressive pointer network to select contents from the source table. SANA expands the skeleton to the final output in the second stage, considering the source table with an edit-based non-autoregressive model. Therefore, SANA could generate more faithful because the input of the second stage is directly extracted from the input table as strong factual guidance.

\citet{shen2020NeuralDatatoText} model the table-to-text task as a segment-by-segment generation precedure and every segment is generated in two-stage. Firstly, proper data records are selected as factual guidance. Secondly, text corresponding to the plan is generated by paying attention only to the selected input data records.

\subsubsection{Auxilary Tasks}
\citet{liu2021FaithfulnessOpen} extend the two-stage table-to-text generator to an augmented plan-based method. They create a pseudo training corpus for the plan-to-text phase, which covers all entities in the target description. In this way, there is no hallucinated entity in the training phase of the plan-to-text generating. Therefore the noise in the original corpus does not affect the plan-to-text generating step.

\subsubsection{Learning methods}
Neural sequence-to-sequence learning-based data-to-text models are often trained by maximum likelihood loss optimization. However, this kind of model suffers from the noise of the training data and leads to unfaithful output (also called divergence).

\citet{wang2020FaithfulNeural} extend the maximum likelihood loss of attention-based Transformer with two additional losses. The first one is a latent matching disagreement loss which measures the distance between embeddings of the input table and the output text. The second one is an entity matching optimal-transport loss to measure the entity matching of the table and the output.
Reinforcement learning (RL) based methods could directly optimize the faithfulness of data-to-text generation.

\citet{liu2019ComprehensiveDescription} propose a force attention method to encourage the model to focus on uncovered data records. Then they adopt RL for information richness to generate more faithful descriptions for the input data. \citet{rebuffel2020PARENTingModelAgnostic} propose an RL-based approach relying on the PARENT metric to reduce the issue of unfaithfulness.

\subsubsection{Constrained Decoding}
\citet{tian2019StickingFacts} propose a confident decoding method to detect and avoid unfaithful generation in the decoder. For each decoder position, a confidence score consists of attention and a dedicated language model that only gives higher scores to common words. Therefore, the confidence score could be utilized to detect the generating of a word conveying source information without paying enough attention to the source data. \citet{tian2019StickingFacts} believe that lower confidence scores indicate higher risks of generating unfaithful text. In the training time, a variational Bayes training framework is designed to ensure the model generates high confidence results. In the inference time, tokens generated with low confidence scores would be marked and skipped.

\citet{filippova2020ControlledHallucinations} treats faithful data-to-text as a controllable text generation problem. In the training corpus, a prefix measuring the amount of noise is appended to the input sequence. Although data-to-text models are trained without modification, we still categorize this method into constrained decoding because the controlling prefix guides the decoding. \citet{rebuffel2022ControllingHallucinations} extend the controllable text generation method to the fine-grained level. Word alignment labels are calculated through dependency parsing, and the labels guide the proposed weighted multi-branch neural decoder.

\subsubsection{Other Methods}
Considering that noise in the training corpus is critical to the faithfulness of data-to-text generation, pre-processing datasets is a direct and reasonable method. \citet{nie2019SimpleRecipe} propose a neural data refinement method to reduce unaligned noise from original datasets. \citet{wang2019RevisitingChallenges} observe that only 60\% of the output contents in RotoWire could be grounded to the input records. They purify and enlarge the original dataset to a new RotoWire-FG dataset.

\subsection{Faithfulness in Other NLG Tasks}
Faithfulness is a common problem in NLG tasks. There are also some researches on faithfulness for other tasks, such as image caption,  image-to-text radiology report generation and general factual language model. The methods for improving faithfulness include: factual guidance by knowledge graph, constrained decoding, and incorporating auxiliary tasks (e.g. entailment, word/entity alignment), etc.

\subsubsection{Factual Language Model}
Some work design language models that are conditioned on external, structured knowledge source to generate factual text. \citet{logan2019barack} introduce the knowledge graph language model (KGLM), a neural language model with mechanisms for selecting and copying information from an external knowledge graph. The KGLM maintains a dynamically growing local knowledge graph, a subset of the knowledge graph that contains entities that have already been mentioned in the text, and their related entities. When generating entity tokens, the model either decides to render a new entity that is absent from the local graph, thereby growing the local knowledge graph, or to render a fact from the local graph. When rendering, the model combines the standard vocabulary with tokens available in the knowledge graph, thus supporting numbers, dates, and other rare tokens.

\begin{figure}
	\centering
	\includegraphics[width=4in]{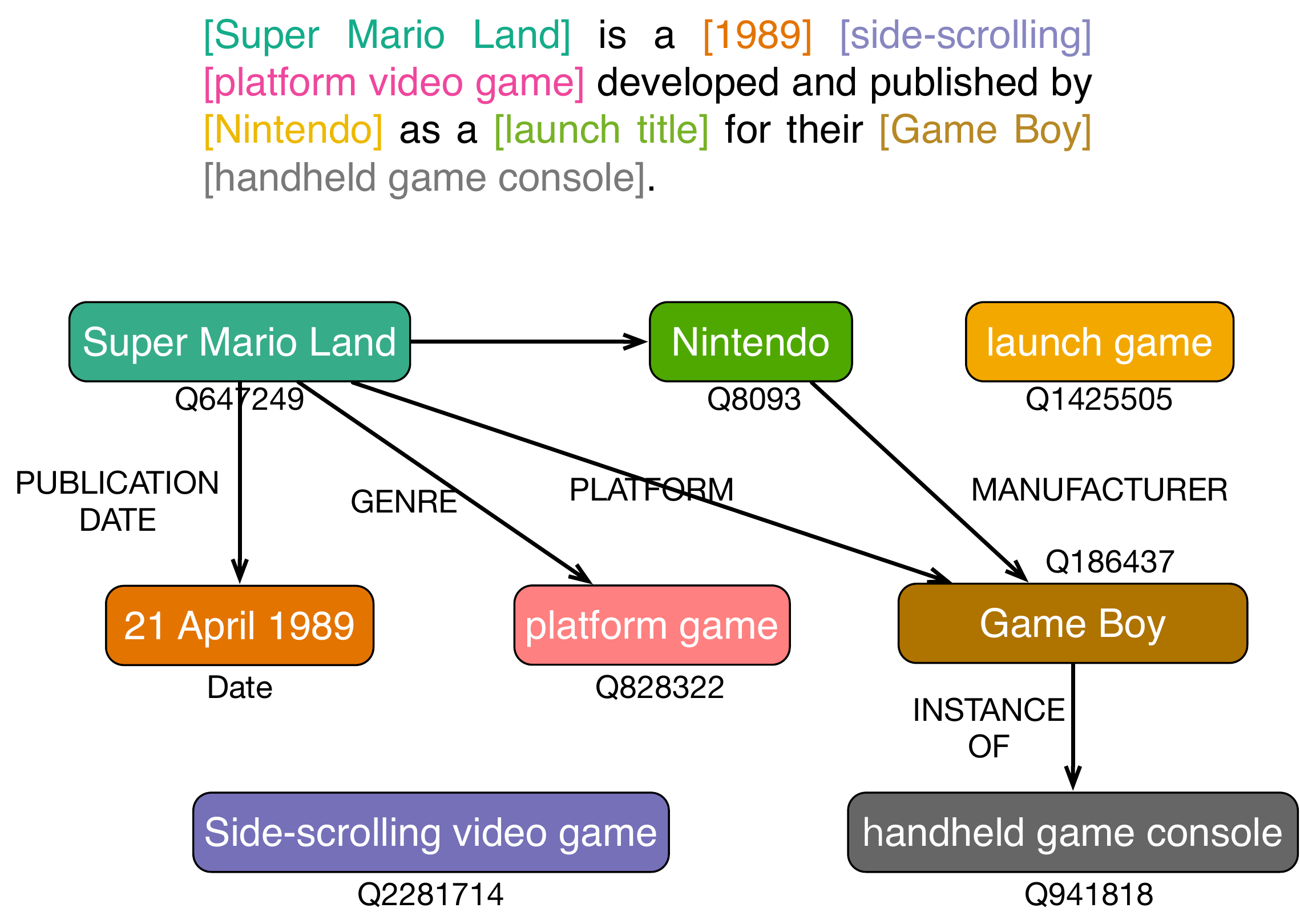}
	\caption{An illustration example of the factual language model \citep{logan2019barack}. It maintains a dynamically growing localized knowledge graph containing facts that are (possibly) conveyed in the sentence above. The graph is built by iteratively linking each detected entity to Wikidata, then adding any relations to previously mentioned entities.}
	\label{factual_lm}
\end{figure}

An example is shown in Figure~\ref{factual_lm}. Initially, the graph is empty and the model uses the entity ``Super Mario Land'' to render the first three tokens, thus adding it and its relations to the local knowledge graph. After generating the next two tokens (“is”, “a”) using the standard language model, the model selects ``Super Mario Land'' as the parent entity, ``Publication Date'' as the relation to render, and copies one of the tokens of the date entity as the token (``1989'' in this case). The factual completion capabilities of KGLM, which predicts the next word after a factual sentence (e.g., ``Barack is married to''), is significantly more accurate. KGLM is able to generate accurate facts for rare entities, and can be controlled via modifications on the knowledge graph.

\subsubsection{Factuality Detection}
Factuality detection is an important task in practical applications.  \citet{hansen2020factuality} build an ensemble learner that predicts news headline factuality using only eye-tracking measurements as they find that false headlines receive statistically significantly less visual attention than true headlines. 
\citet{meng2020gradient} propose a gradient-based adversarial training on transformer networks to the task of detecting check-worthy claims.
\citet{zhong2020neural} propose a graph-based reasoning approach utilizing factual structure of text for deepfake detection.
\citet{zellers_defending_2020} propose a controllable text generation model Grover to defend against fake news, which can generate fake news that are more trustworthy than human-written disinformation.
Counterintuitively, the best defense against Grover turns out to be Grover itself, with 92\% accuracy versus 73\% from best discriminators.
As an important factuality detection task, fake news detection has been widely explored in the literature \citep{shu2017fake,shu2019beyond,jain2018fake,reis2019supervised}.

\subsubsection{Constrained Text Generation}
Generating text under specific lexical constraints is challenging, which can help generate more faithful and factual texts. Constrained text generation broadly falls into two categories, depending on whether inclusion of specified keywords in the output is mandatory. 

In soft-constrained generation \citep{qin2019conversing,tang2019target}, keyword-text pairs are typically first constructed (sometimes along with other conditioning information), and a conditional text generation model is trained to capture their co-occurrence, so that the model learns to incorporate the constrained keywords into the generated text. While soft constrained models are easy to design, keywords are apt to be lost during generation, especially when multiple keywords must be included, or the keywords are less correlated. Soft enforcing algorithms such as attention and copy mechanisms \citep{bahdanau2014neural,gu2016incorporating} can be helpful in preserving keywords, but do not guarantee that constraints will be included in the output sentence. 

Hard-constrained generation \citep{hokamp_lexically_2017,post_fast_2018,hu2019improved,miao2019cgmh,welleck2019non}, on the other hand, requires that all the lexical constraints be present in the output sentence. This approach typically involve sophisticated design of network architectures. \citet{hokamp_lexically_2017} construct a lexical-constrained grid beam search decoding algorithm to incorporate constraints. However, \citet{hu2019improved} observe that a naive implementation of this algorithm has a high running time complexity. \citet{miao2019cgmh} introduces a sampling-based conditional generation method, where the constraints are first placed in a template, then words in a random position are either inserted, deleted or updated under a Metropolis-Hastings-like scheme. However, individually sampling each token result in slow convergence, as the joint distribution of all the tokens in a sentence is highly correlated. \citet{welleck2019non} propose a tree-based text generation scheme, where a token is first generated in an arbitrary position, and then the model recursively generates words to its left and right, yielding a binary tree. However, the constructed tree may not reflect the progressive hierarchy/granularity from high-level concepts to low-level details. Further, the time complexity of generating a sentence using this approach is O(n), like standard auto-regressive generation methods.

\citet{zhang_pointer_2020} propose a novel non-autoregressive model for hard-constrained text generation, called POINTER (PrOgressive INsertion based TransformER). Given lexical constraints, POINTER first generates high-level words (e.g., informative nouns, verbs and adjectives) that bridge the keyword constraints, then these words are used as pivoting points at which to insert details of finer granularity. This process iterates until a sentence is finally completed by adding the least informative words (typically pronouns and prepositions).

\citet{anderson_guided_2017} uses constrained beam search to force the inclusion of selected tag words in image caption, and fixed, pre-trained word embeddings to facilitate vocabulary expansion to previously unseen tag words. Constrained beam search is an approximate search algorithm capable of enforcing any constraints over resulting output sequences that can be expressed in a finite-state machine.

\subsubsection{Image Caption}

\begin{figure}
	\centering
	\includegraphics[width=5in]{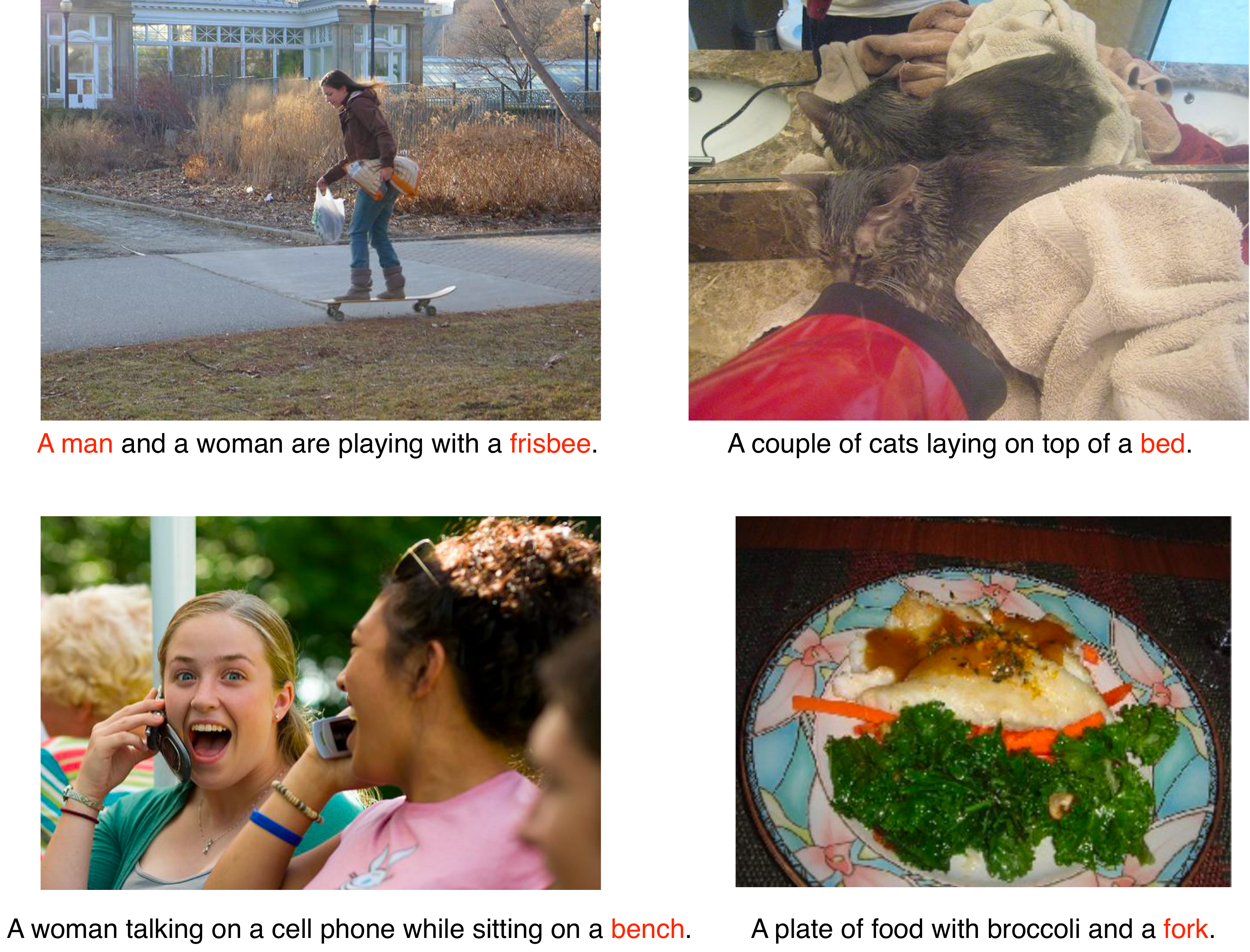}
	\caption{Several examples of image captioning with unfaithful object description.}
	\label{image_caption_example}
\end{figure}

Image caption is an important type of multimodality-to-text generation.
Image captioning models are prone to ``hallucinating'' objects that are not actually in a scene.
Several examples are shown in Figure~\ref{image_caption_example}.
The standard evaluation metrics only measure similarity to ground truth captions and cannot fully capture image relevance.
\citet{rohrbach_object_2018} propose a image relevance metric to evaluate image captioning models with veridical visual labels and assess their rate of object hallucination.
\citet{anderson_guided_2017} use constrained beam search to
force the inclusion of selected tag words
in the output, and fixed pre-trained word
embeddings to facilitate vocabulary expansion to previously unseen tag words.

Image-to-Text radiology report generation is a new type of image caption, which is an important application of natural language generation (NLG). It is to build assistive systems that take X-ray images of a patient and generate a textual report describing clinical observations in the images \citep{jing2017automatic,li2018hybrid,liu2019clinically,boag2020baselines,chen2020generating}. This is a clinically important task, offering the potential to reduce radiologists’ repetitive work and generally improve clinical communication \citep{kahn2009toward}. However, the automatic generated reports by neural models are not always factually complete or consistent. They usually face the issues of factual incompleteness and inconsistency \citep{liu2019clinically, boag2020baselines}.

\citet{miura_improving_2021} show that existing image-to-text radiology report generation systems can be substantially improved by replacing widely used NLG metrics with simple alternatives. They propose two new simple rewards that can encourage the factual completeness and consistency of the generated reports. First, they propose the Exact Entity Match Reward (factENT) which captures the completeness of a generated report by measuring its coverage of entities in the radiology domain, compared with a reference report. The goal of the reward is to better capture disease and anatomical knowledge that are encoded in the entities. Second, they propose the Entailing Entity Match Reward (factENTNLI), which extends factENT with a natural language inference (NLI) model that further considers how inferentially consistent the generated entities are with their descriptions in the reference. They add NLI to control the overestimation of disease when optimizing towards factENT. They directly optimizes these two rewards with RL, showing that the proposed approach substantially improves performance on factual consistency and completeness.

\section{Discussion}
The faithfulness problem is the most critical challenge in modern NLG. 
As described above, we have discussed evaluation metrics and optimization methods for different NLG tasks.
In the following, we discuss some of the limitations of current evaluation and optimization methods, and provide several research directions worth investigating in the general NLG.

\subsection{Fine-grained and General Evaluation}
Most existing faithfulness evaluation metrics measure the faithfulness of the generated text as a score, such as entailment score or QA matching score. They cannot distinguish between different types of factual errors and cannot locate specific error spans, which are detrimental to robust evaluation and repair.
Moreover, most of them mainly focus on specific tasks, rather than be general to all NLG tasks.
More fine-grained and general evaluation methods are needed to drive further developments in the field of NLG.

\paragraph{Intrinsic and Extrinsic}
Most of the existing evaluation metrics measure intrinsic and extrinsic factual errors as a unified metric without distinguishing them.
We argue that intrinsic and extrinsic factual errors should be evaluated separately, as they have different definition and reference.
The reference for the intrinsic error is the source text, however, the reference for the extrinsic error include the source text and world knowledge.
As the extrinsic hallucinations contains both factual and non-factual information, it is necessary to distinguish them as the non-factual extrinsic hallucinations are harmful while the factual extrinsic hallucinations are usually beneficial for NLG tasks, such as dialogue generation.
The field of fact checking is promising to help detect non-factual extrinsic hallucinations.

\paragraph{Fine-grained Error Types}
None of existing evaluation metrics can locate specific error spans and their fine-grained error types, such as predicate error, entity error, circumstance error or discourse link error.
Fine-grained error types can help post-editing methods to fix unfaithful content and also provide richer insight to the researchers.

\paragraph{General Evaluation}
Existing evaluation metrics are mainly designed for specific tasks, such as text summarization or table-to-text generation.
Although some of them can be applied to other NLG tasks, task-agnostic general metrics and evaluation benchmarks are lacking.
Although the source and output texts of different tasks come in various forms, it is worth exploring the relationship between them and propose a general and fine-grained metric to evaluate faithfulness.
Task-agnostic metrics with cross-domain robustness can help the research community to establish a unified benchmark, which is important and meaningful to help collaborate and standardize evaluation metrics for NLG tasks.

\subsection{Reasoning-based Optimization}
Most intrinsic errors are caused by misunderstanding the facts in the source context.
Besides NLU-based pre-training on large scale corpus, to help models understand the facts correctly requires reasoning over the input context or world knowledge. 

\paragraph{Numerical Reasoning}
The correctness of the numerals in the generated texts such as date, quantity and scalar are important for readers to get the correct information.
As existing models usually model numerals in the same way as textual tokens, such as splitting numerals into sub-words or byte-pairs, they are more vulnerable to numerical errors.
However, most of the existing optimization methods do not focus on the faithfulness of numerals.
Tasks with quantities such as table-to-text generation, require numerical reasoning.
Therefore, adding reasoning ability to numerical modeling is crucial for faithfulness optimization.

\paragraph{Grounded Language Representation}
The expressiveness of language generation models are vital to the faithfulness of the generated results.
A model with poor input text representation will fail to do document-level understanding and inference.
Language grounding is an active field aiming at enriching textual representations with visual information, which has been shown to improve performance on a variety of core NLP tasks
\citep{bruni2014multimodal,baroni2016grounding,kiela2017deep}.
Some recent work also propose unified-modal models for language understanding and generation UNIMO \citep{li2020unimo}.
Learning grounded language representation is a promising direction for improving the expressiveness of NLG models, thus improving the faithfulness of their generated texts.

\paragraph{Incorporating Causal Inference}
Existing language models are mostly correlation prediction models, where predictions are due to correlation rather than causal inference. Therefore, these correlational predictive models are not credible and may lead to errors in out-of-distribution or long-tailed situations. Clearly, the lack of causality is one of the main reasons for poor model generalization and faithfulness. Incorporating causal inference into language model may be the fundamental method to solve the faithfulness of NLG models.

\section{Conclusion}
In this survey, we conduct a systematic overview of the faithfulness problem across different NLG tasks. 
We organize and discuss the faithfulness analysis, evaluation metrics and optimization methods in a combined manner under a general categorization standard.
The evaluation metrics for different NLG tasks are categorized into four types: Entailment-based, QA-based, Fact-based and Others.
The optimization methods are categorized into six types: Factual Guidance, Auxiliary Task, Post-Editing, Learning Method, Constrained Decoding and Others.
All the evaluation metrics and optimization methods are discussed and compared to facilitate both task-specific and task-agnostics understanding of the faithfulness problem.
In addition, we propose potential future directions according to the challenges of this problem and the current research status of evaluation metrics and optimization methods.

\bibliographystyle{plainnat}
\bibliography{survey}

\begin{thebibliography}{205}
\providecommand{\natexlab}[1]{#1}
\providecommand{\url}[1]{\texttt{#1}}
\expandafter\ifx\csname urlstyle\endcsname\relax
  \providecommand{\doi}[1]{doi: #1}\else
  \providecommand{\doi}{doi: \begingroup \urlstyle{rm}\Url}\fi

\bibitem[Allahyari et~al.(2017)Allahyari, Pouriyeh, Assefi, Safaei, Trippe,
  Gutierrez, and Kochut]{allahyari2017text}
Mehdi Allahyari, Seyedamin Pouriyeh, Mehdi Assefi, Saeid Safaei, Elizabeth~D
  Trippe, Juan~B Gutierrez, and Krys Kochut.
\newblock Text summarization techniques: a brief survey.
\newblock \emph{arXiv preprint arXiv:1707.02268}, 2017.

\bibitem[Anderson et~al.(2017)Anderson, Fernando, Johnson, and
  Gould]{anderson_guided_2017}
Peter Anderson, Basura Fernando, Mark Johnson, and Stephen Gould.
\newblock Guided {Open} {Vocabulary} {Image} {Captioning} with {Constrained}
  {Beam} {Search}.
\newblock In \emph{Proceedings of the 2017 {Conference} on {Empirical}
  {Methods} in {Natural} {Language} {Processing}}, pages 936--945, Copenhagen,
  Denmark, 2017. Association for Computational Linguistics.
\newblock \doi{10.18653/v1/D17-1098}.
\newblock URL \url{https://aclanthology.org/D17-1098}.

\bibitem[Aralikatte et~al.(2021)Aralikatte, Narayan, Maynez, Rothe, and
  McDonald]{DBLP:conf/acl/AralikatteNMRM20}
Rahul Aralikatte, Shashi Narayan, Joshua Maynez, Sascha Rothe, and Ryan~T.
  McDonald.
\newblock Focus attention: Promoting faithfulness and diversity in
  summarization.
\newblock In Chengqing Zong, Fei Xia, Wenjie Li, and Roberto Navigli, editors,
  \emph{Proceedings of the 59th Annual Meeting of the Association for
  Computational Linguistics and the 11th International Joint Conference on
  Natural Language Processing, {ACL/IJCNLP} 2021, (Volume 1: Long Papers),
  Virtual Event, August 1-6, 2021}, pages 6078--6095. Association for
  Computational Linguistics, 2021.
\newblock \doi{10.18653/v1/2021.acl-long.474}.
\newblock URL \url{https://doi.org/10.18653/v1/2021.acl-long.474}.

\bibitem[Arun et~al.(2020)Arun, Batra, Bhardwaj, Challa, Donmez, Heidari, Inan,
  Jain, Kumar, Mei, Mohan, and White]{DBLP:conf/coling/ArunBBCDHIJKMMW20}
Ankit Arun, Soumya Batra, Vikas Bhardwaj, Ashwini Challa, Pinar Donmez, Peyman
  Heidari, Hakan Inan, Shashank Jain, Anuj Kumar, Shawn Mei, Karthik Mohan, and
  Michael White.
\newblock Best practices for data-efficient modeling in {NLG:} how to train
  production-ready neural models with less data.
\newblock In Ann Clifton and Courtney Napoles, editors, \emph{Proceedings of
  the 28th International Conference on Computational Linguistics, {COLING} 2020
  - Industry Track, Online, December 12, 2020}, pages 64--77. International
  Committee on Computational Linguistics, 2020.
\newblock \doi{10.18653/v1/2020.coling-industry.7}.
\newblock URL \url{https://doi.org/10.18653/v1/2020.coling-industry.7}.

\bibitem[Bahdanau et~al.(2014)Bahdanau, Cho, and Bengio]{bahdanau2014neural}
Dzmitry Bahdanau, Kyunghyun Cho, and Yoshua Bengio.
\newblock Neural machine translation by jointly learning to align and
  translate.
\newblock \emph{arXiv preprint arXiv:1409.0473}, 2014.

\bibitem[Balakrishnan et~al.(2019)Balakrishnan, Rao, Upasani, White, and
  Subba]{DBLP:conf/acl/BalakrishnanRUW19}
Anusha Balakrishnan, Jinfeng Rao, Kartikeya Upasani, Michael White, and Rajen
  Subba.
\newblock Constrained decoding for neural {NLG} from compositional
  representations in task-oriented dialogue.
\newblock In Anna Korhonen, David~R. Traum, and Llu{\'{\i}}s M{\`{a}}rquez,
  editors, \emph{Proceedings of the 57th Conference of the Association for
  Computational Linguistics, {ACL} 2019, Florence, Italy, July 28- August 2,
  2019, Volume 1: Long Papers}, pages 831--844. Association for Computational
  Linguistics, 2019.
\newblock \doi{10.18653/v1/p19-1080}.
\newblock URL \url{https://doi.org/10.18653/v1/p19-1080}.

\bibitem[Banerjee and Lavie(2005)]{banerjee2005meteor}
Satanjeev Banerjee and Alon Lavie.
\newblock Meteor: An automatic metric for mt evaluation with improved
  correlation with human judgments.
\newblock In \emph{Proceedings of the acl workshop on intrinsic and extrinsic
  evaluation measures for machine translation and/or summarization}, pages
  65--72, 2005.

\bibitem[Bao et~al.(2020)Bao, He, Wang, Wu, and Wang]{DBLP:conf/acl/BaoHWWW20}
Siqi Bao, Huang He, Fan Wang, Hua Wu, and Haifeng Wang.
\newblock {PLATO:} pre-trained dialogue generation model with discrete latent
  variable.
\newblock In Dan Jurafsky, Joyce Chai, Natalie Schluter, and Joel~R. Tetreault,
  editors, \emph{Proceedings of the 58th Annual Meeting of the Association for
  Computational Linguistics, {ACL} 2020, Online, July 5-10, 2020}, pages
  85--96. Association for Computational Linguistics, 2020.
\newblock \doi{10.18653/v1/2020.acl-main.9}.
\newblock URL \url{https://doi.org/10.18653/v1/2020.acl-main.9}.

\bibitem[Baroni(2016)]{baroni2016grounding}
Marco Baroni.
\newblock Grounding distributional semantics in the visual world.
\newblock \emph{Language and Linguistics Compass}, 10\penalty0 (1):\penalty0
  3--13, 2016.

\bibitem[Barrantes et~al.(2020)Barrantes, Herudek, and
  Wang]{DBLP:journals/corr/abs-2005-11739}
Mario Barrantes, Benedikt Herudek, and Richard Wang.
\newblock Adversarial {NLI} for factual correctness in text summarisation
  models.
\newblock \emph{CoRR}, abs/2005.11739, 2020.
\newblock URL \url{https://arxiv.org/abs/2005.11739}.

\bibitem[Belinkov and Bisk(2017)]{belinkov2017synthetic}
Yonatan Belinkov and Yonatan Bisk.
\newblock Synthetic and natural noise both break neural machine translation.
\newblock \emph{arXiv preprint arXiv:1711.02173}, 2017.

\bibitem[Boag et~al.(2020)Boag, Hsu, McDermott, Berner, Alesentzer, and
  Szolovits]{boag2020baselines}
William Boag, Tzu-Ming~Harry Hsu, Matthew McDermott, Gabriela Berner, Emily
  Alesentzer, and Peter Szolovits.
\newblock Baselines for chest x-ray report generation.
\newblock In \emph{Machine Learning for Health Workshop}, pages 126--140. PMLR,
  2020.

\bibitem[Brown et~al.(1983)Brown, Brown, Brown, Yule, and
  Gillian]{brown1983discourse}
Gillian Brown, Gillian~D Brown, Gillian~R Brown, George Yule, and Brown
  Gillian.
\newblock \emph{Discourse analysis}.
\newblock Cambridge university press, 1983.

\bibitem[Brown et~al.(2020)Brown, Mann, Ryder, Subbiah, Kaplan, Dhariwal,
  Neelakantan, Shyam, Sastry, Askell, et~al.]{brown2020language}
Tom Brown, Benjamin Mann, Nick Ryder, Melanie Subbiah, Jared~D Kaplan, Prafulla
  Dhariwal, Arvind Neelakantan, Pranav Shyam, Girish Sastry, Amanda Askell,
  et~al.
\newblock Language models are few-shot learners.
\newblock \emph{Advances in neural information processing systems},
  33:\penalty0 1877--1901, 2020.

\bibitem[Bruni et~al.(2014)Bruni, Tran, and Baroni]{bruni2014multimodal}
Elia Bruni, Nam-Khanh Tran, and Marco Baroni.
\newblock Multimodal distributional semantics.
\newblock \emph{Journal of artificial intelligence research}, 49:\penalty0
  1--47, 2014.

\bibitem[Cao et~al.(2020)Cao, Dong, Wu, and Cheung]{cao2020factual}
Meng Cao, Yue Dong, Jiapeng Wu, and Jackie Chi~Kit Cheung.
\newblock Factual error correction for abstractive summarization models.
\newblock In \emph{Proceedings of the 2020 Conference on Empirical Methods in
  Natural Language Processing (EMNLP)}, pages 6251--6258, 2020.

\bibitem[Cao et~al.(2021)Cao, Dong, and Cheung]{cao_inspecting_2021}
Meng Cao, Yue Dong, and Jackie Chi~Kit Cheung.
\newblock Inspecting the {Factuality} of {Hallucinated} {Entities} in
  {Abstractive} {Summarization}.
\newblock \emph{arXiv:2109.09784 [cs]}, August 2021.
\newblock URL \url{http://arxiv.org/abs/2109.09784}.
\newblock arXiv: 2109.09784.

\bibitem[Cao and Wang(2021)]{cao_cliff_2021}
Shuyang Cao and Lu~Wang.
\newblock {CLIFF}: {Contrastive} {Learning} for {Improving} {Faithfulness} and
  {Factuality} in {Abstractive} {Summarization}.
\newblock In \emph{Proceedings of the 2021 {Conference} on {Empirical}
  {Methods} in {Natural} {Language} {Processing}}, pages 6633--6649, Online and
  Punta Cana, Dominican Republic, November 2021. Association for Computational
  Linguistics.
\newblock URL \url{https://aclanthology.org/2021.emnlp-main.532}.

\bibitem[Cao et~al.(2017)Cao, Wei, Li, and Li]{cao_faithful_2017}
Ziqiang Cao, Furu Wei, Wenjie Li, and Sujian Li.
\newblock Faithful to the {Original}: {Fact} {Aware} {Neural} {Abstractive}
  {Summarization}.
\newblock \emph{arXiv:1711.04434 [cs]}, November 2017.
\newblock URL \url{http://arxiv.org/abs/1711.04434}.
\newblock arXiv: 1711.04434.

\bibitem[Cao et~al.(2018)Cao, Li, Li, and Wei]{cao2018retrieve}
Ziqiang Cao, Wenjie Li, Sujian Li, and Furu Wei.
\newblock Retrieve, rerank and rewrite: Soft template based neural
  summarization.
\newblock In \emph{Proceedings of the 56th Annual Meeting of the Association
  for Computational Linguistics (Volume 1: Long Papers)}, pages 152--161, 2018.

\bibitem[Chen et~al.(2020{\natexlab{a}})Chen, Stanovsky, Singh, and
  Gardner]{DBLP:conf/emnlp/ChenSSG20}
Anthony Chen, Gabriel Stanovsky, Sameer Singh, and Matt Gardner.
\newblock {MOCHA:} {A} dataset for training and evaluating generative reading
  comprehension metrics.
\newblock In Bonnie Webber, Trevor Cohn, Yulan He, and Yang Liu, editors,
  \emph{Proceedings of the 2020 Conference on Empirical Methods in Natural
  Language Processing, {EMNLP} 2020, Online, November 16-20, 2020}, pages
  6521--6532. Association for Computational Linguistics, 2020{\natexlab{a}}.
\newblock \doi{10.18653/v1/2020.emnlp-main.528}.
\newblock URL \url{https://doi.org/10.18653/v1/2020.emnlp-main.528}.

\bibitem[Chen et~al.(2017)Chen, Zhu, Ling, Wei, Jiang, and
  Inkpen]{DBLP:conf/acl/ChenZLWJI17}
Qian Chen, Xiaodan Zhu, Zhen{-}Hua Ling, Si~Wei, Hui Jiang, and Diana Inkpen.
\newblock Enhanced {LSTM} for natural language inference.
\newblock In Regina Barzilay and Min{-}Yen Kan, editors, \emph{Proceedings of
  the 55th Annual Meeting of the Association for Computational Linguistics,
  {ACL} 2017, Vancouver, Canada, July 30 - August 4, Volume 1: Long Papers},
  pages 1657--1668. Association for Computational Linguistics, 2017.
\newblock \doi{10.18653/v1/P17-1152}.
\newblock URL \url{https://doi.org/10.18653/v1/P17-1152}.

\bibitem[Chen et~al.(2021)Chen, Zhang, Sone, and Roth]{chen_improving_2021}
Sihao Chen, Fan Zhang, Kazoo Sone, and Dan Roth.
\newblock Improving {Faithfulness} in {Abstractive} {Summarization} with
  {Contrast} {Candidate} {Generation} and {Selection}.
\newblock In \emph{Proceedings of the 2021 {Conference} of the {North}
  {American} {Chapter} of the {Association} for {Computational} {Linguistics}:
  {Human} {Language} {Technologies}}, pages 5935--5941, Online, June 2021.
  Association for Computational Linguistics.
\newblock \doi{10.18653/v1/2021.naacl-main.475}.
\newblock URL \url{https://aclanthology.org/2021.naacl-main.475}.

\bibitem[Chen et~al.(2020{\natexlab{b}})Chen, Su, Yan, and Wang]{chen2020kgpt}
Wenhu Chen, Yu~Su, Xifeng Yan, and William~Yang Wang.
\newblock Kgpt: Knowledge-grounded pre-training for data-to-text generation.
\newblock \emph{arXiv preprint arXiv:2010.02307}, 2020{\natexlab{b}}.

\bibitem[Chen et~al.(2020{\natexlab{c}})Chen, Song, Chang, and
  Wan]{chen2020generating}
Zhihong Chen, Yan Song, Tsung-Hui Chang, and Xiang Wan.
\newblock Generating radiology reports via memory-driven transformer.
\newblock \emph{arXiv preprint arXiv:2010.16056}, 2020{\natexlab{c}}.

\bibitem[Cheng et~al.(2018)Cheng, Tu, Meng, Zhai, and Liu]{cheng2018towards}
Yong Cheng, Zhaopeng Tu, Fandong Meng, Junjie Zhai, and Yang Liu.
\newblock Towards robust neural machine translation.
\newblock \emph{arXiv preprint arXiv:1805.06130}, 2018.

\bibitem[Cheng et~al.(2019)Cheng, Jiang, and Macherey]{cheng_robust_2019}
Yong Cheng, Lu~Jiang, and Wolfgang Macherey.
\newblock Robust {Neural} {Machine} {Translation} with {Doubly} {Adversarial}
  {Inputs}.
\newblock In \emph{Proceedings of the 57th {Annual} {Meeting} of the
  {Association} for {Computational} {Linguistics}}, pages 4324--4333, Florence,
  Italy, July 2019. Association for Computational Linguistics.
\newblock \doi{10.18653/v1/P19-1425}.
\newblock URL \url{https://aclanthology.org/P19-1425}.

\bibitem[Choubey et~al.(2021)Choubey, Vig, Liu, and Rajani]{choubey_mofe_2021}
Prafulla~Kumar Choubey, Jesse Vig, Wenhao Liu, and Nazneen~Fatema Rajani.
\newblock {MoFE}: {Mixture} of {Factual} {Experts} for {Controlling}
  {Hallucinations} in {Abstractive} {Summarization}.
\newblock \emph{arXiv:2110.07166 [cs]}, October 2021.
\newblock URL \url{http://arxiv.org/abs/2110.07166}.
\newblock arXiv: 2110.07166.

\bibitem[Clark et~al.(2020)Clark, Luong, Le, and
  Manning]{DBLP:conf/iclr/ClarkLLM20}
Kevin Clark, Minh{-}Thang Luong, Quoc~V. Le, and Christopher~D. Manning.
\newblock {ELECTRA:} pre-training text encoders as discriminators rather than
  generators.
\newblock In \emph{8th International Conference on Learning Representations,
  {ICLR} 2020, Addis Ababa, Ethiopia, April 26-30, 2020}. OpenReview.net, 2020.
\newblock URL \url{https://openreview.net/forum?id=r1xMH1BtvB}.

\bibitem[Demszky et~al.(2018)Demszky, Guu, and
  Liang]{DBLP:journals/corr/abs-1809-02922}
Dorottya Demszky, Kelvin Guu, and Percy Liang.
\newblock Transforming question answering datasets into natural language
  inference datasets.
\newblock \emph{CoRR}, abs/1809.02922, 2018.
\newblock URL \url{http://arxiv.org/abs/1809.02922}.

\bibitem[Devlin et~al.(2018)Devlin, Chang, Lee, and Toutanova]{devlin2018bert}
Jacob Devlin, Ming-Wei Chang, Kenton Lee, and Kristina Toutanova.
\newblock Bert: Pre-training of deep bidirectional transformers for language
  understanding.
\newblock \emph{arXiv preprint arXiv:1810.04805}, 2018.

\bibitem[Dhingra et~al.(2019)Dhingra, Faruqui, Parikh, Chang, Das, and
  Cohen]{dhingra_handling_2019}
Bhuwan Dhingra, Manaal Faruqui, Ankur Parikh, Ming-Wei Chang, Dipanjan Das, and
  William Cohen.
\newblock Handling {Divergent} {Reference} {Texts} when {Evaluating}
  {Table}-to-{Text} {Generation}.
\newblock In \emph{Proceedings of the 57th {Annual} {Meeting} of the
  {Association} for {Computational} {Linguistics}}, pages 4884--4895, Florence,
  Italy, 2019. Association for Computational Linguistics.
\newblock \doi{10.18653/v1/P19-1483}.
\newblock URL \url{https://aclanthology.org/P19-1483}.

\bibitem[Dinan et~al.(2019)Dinan, Roller, Shuster, Fan, Auli, and
  Weston]{DBLP:conf/iclr/DinanRSFAW19}
Emily Dinan, Stephen Roller, Kurt Shuster, Angela Fan, Michael Auli, and Jason
  Weston.
\newblock Wizard of wikipedia: Knowledge-powered conversational agents.
\newblock In \emph{7th International Conference on Learning Representations,
  {ICLR} 2019, New Orleans, LA, USA, May 6-9, 2019}. OpenReview.net, 2019.
\newblock URL \url{https://openreview.net/forum?id=r1l73iRqKm}.

\bibitem[Dong et~al.(2019)Dong, Yang, Wang, Wei, Liu, Wang, Gao, Zhou, and
  Hon]{dong2019unified}
Li~Dong, Nan Yang, Wenhui Wang, Furu Wei, Xiaodong Liu, Yu~Wang, Jianfeng Gao,
  Ming Zhou, and Hsiao-Wuen Hon.
\newblock Unified language model pre-training for natural language
  understanding and generation.
\newblock \emph{Advances in Neural Information Processing Systems}, 32, 2019.

\bibitem[Dong et~al.(2020)Dong, Wang, Gan, Cheng, Cheung, and
  Liu]{dong2020multi}
Yue Dong, Shuohang Wang, Zhe Gan, Yu~Cheng, Jackie Chi~Kit Cheung, and Jingjing
  Liu.
\newblock Multi-fact correction in abstractive text summarization.
\newblock In \emph{Proceedings of the 2020 Conference on Empirical Methods in
  Natural Language Processing (EMNLP)}, pages 9320--9331, 2020.

\bibitem[Dou et~al.(2021)Dou, Liu, Hayashi, Jiang, and Neubig]{dou2021gsum}
Zi-Yi Dou, Pengfei Liu, Hiroaki Hayashi, Zhengbao Jiang, and Graham Neubig.
\newblock Gsum: A general framework for guided neural abstractive
  summarization.
\newblock In \emph{Proceedings of the 2021 Conference of the North American
  Chapter of the Association for Computational Linguistics: Human Language
  Technologies}, pages 4830--4842, 2021.

\bibitem[Du et~al.(2017)Du, Shao, and Cardie]{du2017learning}
Xinya Du, Junru Shao, and Claire Cardie.
\newblock Learning to ask: Neural question generation for reading
  comprehension.
\newblock \emph{arXiv preprint arXiv:1705.00106}, 2017.

\bibitem[Durmus et~al.(2020)Durmus, He, and Diab]{durmus_feqa_2020}
Esin Durmus, He~He, and Mona Diab.
\newblock {FEQA}: {A} {Question} {Answering} {Evaluation} {Framework} for
  {Faithfulness} {Assessment} in {Abstractive} {Summarization}.
\newblock \emph{Proceedings of the 58th Annual Meeting of the Association for
  Computational Linguistics}, pages 5055--5070, 2020.
\newblock \doi{10.18653/v1/2020.acl-main.454}.
\newblock URL \url{http://arxiv.org/abs/2005.03754}.
\newblock arXiv: 2005.03754.

\bibitem[Dziri et~al.(2021{\natexlab{a}})Dziri, Madotto, Zaïane, and
  Bose]{dziri_neural_2021}
Nouha Dziri, Andrea Madotto, Osmar Zaïane, and Avishek~Joey Bose.
\newblock Neural {Path} {Hunter}: {Reducing} {Hallucination} in {Dialogue}
  {Systems} via {Path} {Grounding}.
\newblock In \emph{Proceedings of the 2021 {Conference} on {Empirical}
  {Methods} in {Natural} {Language} {Processing}}, pages 2197--2214, Online and
  Punta Cana, Dominican Republic, November 2021{\natexlab{a}}. Association for
  Computational Linguistics.
\newblock URL \url{https://aclanthology.org/2021.emnlp-main.168}.

\bibitem[Dziri et~al.(2021{\natexlab{b}})Dziri, Rashkin, Linzen, and
  Reitter]{DBLP:journals/corr/abs-2105-00071}
Nouha Dziri, Hannah Rashkin, Tal Linzen, and David Reitter.
\newblock Evaluating groundedness in dialogue systems: The {BEGIN} benchmark.
\newblock \emph{CoRR}, abs/2105.00071, 2021{\natexlab{b}}.
\newblock URL \url{https://arxiv.org/abs/2105.00071}.

\bibitem[Ebrahimi et~al.(2017)Ebrahimi, Rao, Lowd, and
  Dou]{ebrahimi2017hotflip}
Javid Ebrahimi, Anyi Rao, Daniel Lowd, and Dejing Dou.
\newblock Hotflip: White-box adversarial examples for text classification.
\newblock \emph{arXiv preprint arXiv:1712.06751}, 2017.

\bibitem[Egan et~al.(2021)Egan, Vasilyev, and Bohannon]{egan_play_2021}
Nicholas Egan, Oleg Vasilyev, and John Bohannon.
\newblock Play the {Shannon} {Game} {With} {Language} {Models}: {A}
  {Human}-{Free} {Approach} to {Summary} {Evaluation}.
\newblock \emph{arXiv:2103.10918 [cs]}, December 2021.
\newblock URL \url{http://arxiv.org/abs/2103.10918}.
\newblock arXiv: 2103.10918.

\bibitem[Fabbri et~al.(2021{\natexlab{a}})Fabbri, Wu, Liu, and
  Xiong]{fabbri_qafacteval_2021}
Alexander~R. Fabbri, Chien-Sheng Wu, Wenhao Liu, and Caiming Xiong.
\newblock {QAFactEval}: {Improved} {QA}-{Based} {Factual} {Consistency}
  {Evaluation} for {Summarization}.
\newblock \emph{arXiv:2112.08542 [cs]}, December 2021{\natexlab{a}}.
\newblock URL \url{http://arxiv.org/abs/2112.08542}.
\newblock arXiv: 2112.08542.

\bibitem[Fabbri et~al.(2021{\natexlab{b}})Fabbri, Wu, Iyer, Li, and
  Diab]{fabbri_answersumm_2021}
Alexander~R. Fabbri, Xiaojian Wu, Srini Iyer, Haoran Li, and Mona Diab.
\newblock {AnswerSumm}: {A} {Manually}-{Curated} {Dataset} and {Pipeline} for
  {Answer} {Summarization}.
\newblock \emph{arXiv:2111.06474 [cs]}, November 2021{\natexlab{b}}.
\newblock URL \url{http://arxiv.org/abs/2111.06474}.
\newblock arXiv: 2111.06474.

\bibitem[Falke et~al.(2019)Falke, Ribeiro, Utama, Dagan, and
  Gurevych]{falke_ranking_2019}
Tobias Falke, Leonardo F.~R. Ribeiro, Prasetya~Ajie Utama, Ido Dagan, and Iryna
  Gurevych.
\newblock Ranking {Generated} {Summaries} by {Correctness}: {An} {Interesting}
  but {Challenging} {Application} for {Natural} {Language} {Inference}.
\newblock In \emph{Proceedings of the 57th {Annual} {Meeting} of the
  {Association} for {Computational} {Linguistics}}, pages 2214--2220, Florence,
  Italy, July 2019. Association for Computational Linguistics.
\newblock \doi{10.18653/v1/P19-1213}.
\newblock URL \url{https://aclanthology.org/P19-1213}.

\bibitem[Feng et~al.(2020)Feng, Xie, Gu, Shao, Zhang, Yang, and
  Yu]{feng2020modeling}
Yang Feng, Wanying Xie, Shuhao Gu, Chenze Shao, Wen Zhang, Zhengxin Yang, and
  Dong Yu.
\newblock Modeling fluency and faithfulness for diverse neural machine
  translation.
\newblock In \emph{Proceedings of the AAAI Conference on Artificial
  Intelligence}, volume~34, pages 59--66, 2020.

\bibitem[Filippova(2020)]{filippova2020ControlledHallucinations}
Katja Filippova.
\newblock Controlled {{Hallucinations}}: Learning to {{Generate Faithfully}}
  from {{Noisy Data}}.
\newblock In \emph{Findings of the {{Association}} for {{Computational
  Linguistics}}: {{EMNLP}} 2020}, pages 864--870, {Online}, 2020. {Association
  for Computational Linguistics}.
\newblock \doi{10.18653/v1/2020.findings-emnlp.76}.

\bibitem[Fillmore et~al.(1976)]{fillmore1976frame}
Charles~J Fillmore et~al.
\newblock Frame semantics and the nature of language.
\newblock In \emph{Annals of the New York Academy of Sciences: Conference on
  the origin and development of language and speech}, volume 280, pages 20--32.
  New York, 1976.

\bibitem[Fleiss(1971)]{fleiss1971measuring}
Joseph~L Fleiss.
\newblock Measuring nominal scale agreement among many raters.
\newblock \emph{Psychological bulletin}, 76\penalty0 (5):\penalty0 378, 1971.

\bibitem[Frank and Goodman(2012)]{frank2012predicting}
Michael~C Frank and Noah~D Goodman.
\newblock Predicting pragmatic reasoning in language games.
\newblock \emph{Science}, 336\penalty0 (6084):\penalty0 998--998, 2012.

\bibitem[Gabriel et~al.(2021{\natexlab{a}})Gabriel, Bosselut, Da, Holtzman,
  Buys, Lo, Celikyilmaz, and Choi]{gabriel2021discourse}
Saadia Gabriel, Antoine Bosselut, Jeff Da, Ari Holtzman, Jan Buys, Kyle Lo,
  Asli Celikyilmaz, and Yejin Choi.
\newblock Discourse understanding and factual consistency in abstractive
  summarization.
\newblock In \emph{Proceedings of the 16th Conference of the European Chapter
  of the Association for Computational Linguistics: Main Volume}, pages
  435--447, 2021{\natexlab{a}}.

\bibitem[Gabriel et~al.(2021{\natexlab{b}})Gabriel, Celikyilmaz, Jha, Choi, and
  Gao]{gabriel_go_2021}
Saadia Gabriel, Asli Celikyilmaz, Rahul Jha, Yejin Choi, and Jianfeng Gao.
\newblock {GO} {FIGURE}: {A} {Meta} {Evaluation} of {Factuality} in
  {Summarization}.
\newblock In \emph{Findings of the {Association} for {Computational}
  {Linguistics}: {ACL}-{IJCNLP} 2021}, pages 478--487, Online, August
  2021{\natexlab{b}}. Association for Computational Linguistics.
\newblock \doi{10.18653/v1/2021.findings-acl.42}.
\newblock URL \url{https://aclanthology.org/2021.findings-acl.42}.

\bibitem[Gao et~al.(2019)Gao, Zhang, Lee, Galley, Brockett, Gao, and
  Dolan]{DBLP:conf/emnlp/GaoZLGBGD19}
Xiang Gao, Yizhe Zhang, Sungjin Lee, Michel Galley, Chris Brockett, Jianfeng
  Gao, and Bill Dolan.
\newblock Structuring latent spaces for stylized response generation.
\newblock In Kentaro Inui, Jing Jiang, Vincent Ng, and Xiaojun Wan, editors,
  \emph{Proceedings of the 2019 Conference on Empirical Methods in Natural
  Language Processing and the 9th International Joint Conference on Natural
  Language Processing, {EMNLP-IJCNLP} 2019, Hong Kong, China, November 3-7,
  2019}, pages 1814--1823. Association for Computational Linguistics, 2019.
\newblock \doi{10.18653/v1/D19-1190}.
\newblock URL \url{https://doi.org/10.18653/v1/D19-1190}.

\bibitem[Garg et~al.(2019)Garg, Peitz, Nallasamy, and Paulik]{garg2019jointly}
Sarthak Garg, Stephan Peitz, Udhyakumar Nallasamy, and Matthias Paulik.
\newblock Jointly learning to align and translate with transformer models.
\newblock \emph{arXiv preprint arXiv:1909.02074}, 2019.

\bibitem[Gatt and Krahmer(2018)]{gatt2018survey}
Albert Gatt and Emiel Krahmer.
\newblock Survey of the state of the art in natural language generation: Core
  tasks, applications and evaluation.
\newblock \emph{Journal of Artificial Intelligence Research}, 61:\penalty0
  65--170, 2018.

\bibitem[Gehrmann et~al.(2021)Gehrmann, Adewumi, Aggarwal, Ammanamanchi,
  Anuoluwapo, Bosselut, Chandu, Clinciu, Das, Dhole, Du, Durmus, Dusek, Emezue,
  Gangal, Garbacea, Hashimoto, Hou, Jernite, Jhamtani, Ji, Jolly, Kumar,
  Ladhak, Madaan, Maddela, Mahajan, Mahamood, Majumder, Martins,
  McMillan{-}Major, Mille, van Miltenburg, Nadeem, Narayan, Nikolaev,
  Niyongabo, Osei, Parikh, Perez{-}Beltrachini, Rao, Raunak, Rodriguez,
  Santhanam, Sedoc, Sellam, Shaikh, Shimorina, Cabezudo, Strobelt, Subramani,
  Xu, Yang, Yerukola, and Zhou]{DBLP:journals/corr/abs-2102-01672}
Sebastian Gehrmann, Tosin~P. Adewumi, Karmanya Aggarwal, Pawan~Sasanka
  Ammanamanchi, Aremu Anuoluwapo, Antoine Bosselut, Khyathi~Raghavi Chandu,
  Miruna{-}Adriana Clinciu, Dipanjan Das, Kaustubh~D. Dhole, Wanyu Du, Esin
  Durmus, Ondrej Dusek, Chris Emezue, Varun Gangal, Cristina Garbacea,
  Tatsunori Hashimoto, Yufang Hou, Yacine Jernite, Harsh Jhamtani, Yangfeng Ji,
  Shailza Jolly, Dhruv Kumar, Faisal Ladhak, Aman Madaan, Mounica Maddela,
  Khyati Mahajan, Saad Mahamood, Bodhisattwa~Prasad Majumder, Pedro~Henrique
  Martins, Angelina McMillan{-}Major, Simon Mille, Emiel van Miltenburg, Moin
  Nadeem, Shashi Narayan, Vitaly Nikolaev, Rubungo~Andre Niyongabo, Salomey
  Osei, Ankur~P. Parikh, Laura Perez{-}Beltrachini, Niranjan~Ramesh Rao, Vikas
  Raunak, Juan~Diego Rodriguez, Sashank Santhanam, Jo{\~{a}}o Sedoc, Thibault
  Sellam, Samira Shaikh, Anastasia Shimorina, Marco Antonio~Sobrevilla
  Cabezudo, Hendrik Strobelt, Nishant Subramani, Wei Xu, Diyi Yang, Akhila
  Yerukola, and Jiawei Zhou.
\newblock The {GEM} benchmark: Natural language generation, its evaluation and
  metrics.
\newblock \emph{CoRR}, abs/2102.01672, 2021.
\newblock URL \url{https://arxiv.org/abs/2102.01672}.

\bibitem[Ghazvininejad et~al.(2018)Ghazvininejad, Brockett, Chang, Dolan, Gao,
  Yih, and Galley]{DBLP:conf/aaai/GhazvininejadBC18}
Marjan Ghazvininejad, Chris Brockett, Ming{-}Wei Chang, Bill Dolan, Jianfeng
  Gao, Wen{-}tau Yih, and Michel Galley.
\newblock A knowledge-grounded neural conversation model.
\newblock In Sheila~A. McIlraith and Kilian~Q. Weinberger, editors,
  \emph{Proceedings of the Thirty-Second {AAAI} Conference on Artificial
  Intelligence, (AAAI-18), the 30th innovative Applications of Artificial
  Intelligence (IAAI-18), and the 8th {AAAI} Symposium on Educational Advances
  in Artificial Intelligence (EAAI-18), New Orleans, Louisiana, USA, February
  2-7, 2018}, pages 5110--5117. {AAAI} Press, 2018.
\newblock URL
  \url{https://www.aaai.org/ocs/index.php/AAAI/AAAI18/paper/view/16710}.

\bibitem[Goodfellow et~al.(2014)Goodfellow, Shlens, and
  Szegedy]{goodfellow2014explaining}
Ian~J Goodfellow, Jonathon Shlens, and Christian Szegedy.
\newblock Explaining and harnessing adversarial examples.
\newblock \emph{arXiv preprint arXiv:1412.6572}, 2014.

\bibitem[Goodrich et~al.(2019)Goodrich, Rao, Saleh, and
  Liu]{goodrich_assessing_2019}
Ben Goodrich, Vinay Rao, Mohammad Saleh, and Peter~J. Liu.
\newblock Assessing {The} {Factual} {Accuracy} of {Generated} {Text}.
\newblock \emph{Proceedings of the 25th ACM SIGKDD International Conference on
  Knowledge Discovery \& Data Mining}, pages 166--175, July 2019.
\newblock \doi{10.1145/3292500.3330955}.
\newblock URL \url{http://arxiv.org/abs/1905.13322}.
\newblock arXiv: 1905.13322.

\bibitem[Goyal and Durrett(2020)]{goyal_evaluating_2020}
Tanya Goyal and Greg Durrett.
\newblock Evaluating {Factuality} in {Generation} with {Dependency}-level
  {Entailment}.
\newblock In \emph{Findings of the {Association} for {Computational}
  {Linguistics}: {EMNLP} 2020}, pages 3592--3603, Online, November 2020.
  Association for Computational Linguistics.
\newblock \doi{10.18653/v1/2020.findings-emnlp.322}.
\newblock URL \url{https://aclanthology.org/2020.findings-emnlp.322}.

\bibitem[Goyal and Durrett(2021)]{goyal_annotating_2021}
Tanya Goyal and Greg Durrett.
\newblock Annotating and {Modeling} {Fine}-grained {Factuality} in
  {Summarization}.
\newblock In \emph{Proceedings of the 2021 {Conference} of the {North}
  {American} {Chapter} of the {Association} for {Computational} {Linguistics}:
  {Human} {Language} {Technologies}}, pages 1449--1462, Online, June 2021.
  Association for Computational Linguistics.
\newblock \doi{10.18653/v1/2021.naacl-main.114}.
\newblock URL \url{https://aclanthology.org/2021.naacl-main.114}.

\bibitem[Graves(2013)]{graves2013generating}
Alex Graves.
\newblock Generating sequences with recurrent neural networks.
\newblock \emph{arXiv preprint arXiv:1308.0850}, 2013.

\bibitem[Gu et~al.(2016)Gu, Lu, Li, and Li]{gu2016incorporating}
Jiatao Gu, Zhengdong Lu, Hang Li, and Victor~OK Li.
\newblock Incorporating copying mechanism in sequence-to-sequence learning.
\newblock \emph{arXiv preprint arXiv:1603.06393}, 2016.

\bibitem[Gupta et~al.(2021)Gupta, Wu, Liu, and
  Xiong]{DBLP:journals/corr/abs-2110-08222}
Prakhar Gupta, Chien{-}Sheng Wu, Wenhao Liu, and Caiming Xiong.
\newblock Dialfact: {A} benchmark for fact-checking in dialogue.
\newblock \emph{CoRR}, abs/2110.08222, 2021.
\newblock URL \url{https://arxiv.org/abs/2110.08222}.

\bibitem[Hadash et~al.(2018)Hadash, Kermany, Carmeli, Lavi, Kour, and
  Jacovi]{hadash2018estimate}
Guy Hadash, Einat Kermany, Boaz Carmeli, Ofer Lavi, George Kour, and Alon
  Jacovi.
\newblock Estimate and replace: A novel approach to integrating deep neural
  networks with existing applications.
\newblock \emph{arXiv preprint arXiv:1804.09028}, 2018.

\bibitem[Hansen et~al.(2020)Hansen, Hansen, Simonsen, Larsen, Alstrup, and
  Lioma]{hansen2020factuality}
Christian Hansen, Casper Hansen, Jakob~Grue Simonsen, Birger Larsen, Stephen
  Alstrup, and Christina Lioma.
\newblock Factuality checking in news headlines with eye tracking.
\newblock In \emph{Proceedings of the 43rd International ACM SIGIR Conference
  on Research and Development in Information Retrieval}, pages 2013--2016,
  2020.

\bibitem[Hokamp and Liu(2017)]{hokamp_lexically_2017}
Chris Hokamp and Qun Liu.
\newblock Lexically {Constrained} {Decoding} for {Sequence} {Generation}
  {Using} {Grid} {Beam} {Search}.
\newblock \emph{arXiv:1704.07138 [cs]}, May 2017.
\newblock URL \url{http://arxiv.org/abs/1704.07138}.
\newblock arXiv: 1704.07138.

\bibitem[Honnibal and Montani(2017)]{honnibal2017spacy}
Matthew Honnibal and Ines Montani.
\newblock spacy 2: Natural language understanding with bloom embeddings,
  convolutional neural networks and incremental parsing.
\newblock \emph{To appear}, 7\penalty0 (1):\penalty0 411--420, 2017.

\bibitem[Honovich et~al.(2021)Honovich, Choshen, Aharoni, Neeman, Szpektor, and
  Abend]{DBLP:conf/emnlp/HonovichCANSA21}
Or~Honovich, Leshem Choshen, Roee Aharoni, Ella Neeman, Idan Szpektor, and Omri
  Abend.
\newblock {\textdollar}q{\^{}}2{\textdollar}: Evaluating factual consistency in
  knowledge-grounded dialogues via question generation and question answering.
\newblock In Marie{-}Francine Moens, Xuanjing Huang, Lucia Specia, and
  Scott~Wen{-}tau Yih, editors, \emph{Proceedings of the 2021 Conference on
  Empirical Methods in Natural Language Processing, {EMNLP} 2021, Virtual Event
  / Punta Cana, Dominican Republic, 7-11 November, 2021}, pages 7856--7870.
  Association for Computational Linguistics, 2021.
\newblock \doi{10.18653/v1/2021.emnlp-main.619}.
\newblock URL \url{https://doi.org/10.18653/v1/2021.emnlp-main.619}.

\bibitem[Hu et~al.(2019)Hu, Khayrallah, Culkin, Xia, Chen, Post, and
  Van~Durme]{hu2019improved}
J~Edward Hu, Huda Khayrallah, Ryan Culkin, Patrick Xia, Tongfei Chen, Matt
  Post, and Benjamin Van~Durme.
\newblock Improved lexically constrained decoding for translation and
  monolingual rewriting.
\newblock In \emph{Proceedings of the 2019 Conference of the North American
  Chapter of the Association for Computational Linguistics: Human Language
  Technologies, Volume 1 (Long and Short Papers)}, pages 839--850, 2019.

\bibitem[Huang et~al.(2020)Huang, Wu, and Wang]{huang2020knowledge}
Luyang Huang, Lingfei Wu, and Lu~Wang.
\newblock Knowledge graph-augmented abstractive summarization with
  semantic-driven cloze reward.
\newblock In \emph{Proceedings of the 58th Annual Meeting of the Association
  for Computational Linguistics}, pages 5094--5107, 2020.

\bibitem[Huang et~al.(2016)Huang, Ferraro, Mostafazadeh, Misra, Agrawal,
  Devlin, Girshick, He, Kohli, Batra, et~al.]{huang2016visual}
Ting-Hao Huang, Francis Ferraro, Nasrin Mostafazadeh, Ishan Misra, Aishwarya
  Agrawal, Jacob Devlin, Ross Girshick, Xiaodong He, Pushmeet Kohli, Dhruv
  Batra, et~al.
\newblock Visual storytelling.
\newblock In \emph{Proceedings of the 2016 Conference of the North American
  Chapter of the Association for Computational Linguistics: Human Language
  Technologies}, pages 1233--1239, 2016.

\bibitem[Jain and Kasbe(2018)]{jain2018fake}
Akshay Jain and Amey Kasbe.
\newblock Fake news detection.
\newblock In \emph{2018 IEEE International Students' Conference on Electrical,
  Electronics and Computer Science (SCEECS)}, pages 1--5. IEEE, 2018.

\bibitem[Jing et~al.(2017)Jing, Xie, and Xing]{jing2017automatic}
Baoyu Jing, Pengtao Xie, and Eric Xing.
\newblock On the automatic generation of medical imaging reports.
\newblock \emph{arXiv preprint arXiv:1711.08195}, 2017.

\bibitem[Junczys-Dowmunt(2018)]{junczys2018dual}
Marcin Junczys-Dowmunt.
\newblock Dual conditional cross-entropy filtering of noisy parallel corpora.
\newblock \emph{arXiv preprint arXiv:1809.00197}, 2018.

\bibitem[Kahn~Jr et~al.(2009)Kahn~Jr, Langlotz, Burnside, Carrino, Channin,
  Hovsepian, and Rubin]{kahn2009toward}
Charles~E Kahn~Jr, Curtis~P Langlotz, Elizabeth~S Burnside, John~A Carrino,
  David~S Channin, David~M Hovsepian, and Daniel~L Rubin.
\newblock Toward best practices in radiology reporting.
\newblock \emph{Radiology}, 252\penalty0 (3):\penalty0 852--856, 2009.

\bibitem[Kalchbrenner et~al.(2014)Kalchbrenner, Grefenstette, and
  Blunsom]{kalchbrenner2014convolutional}
Nal Kalchbrenner, Edward Grefenstette, and Phil Blunsom.
\newblock A convolutional neural network for modelling sentences.
\newblock \emph{arXiv preprint arXiv:1404.2188}, 2014.

\bibitem[Kang and Hashimoto(2020)]{kang2020improved}
Daniel Kang and Tatsunori Hashimoto.
\newblock Improved natural language generation via loss truncation.
\newblock \emph{arXiv preprint arXiv:2004.14589}, 2020.

\bibitem[Karpukhin et~al.(2019)Karpukhin, Levy, Eisenstein, and
  Ghazvininejad]{karpukhin2019training}
Vladimir Karpukhin, Omer Levy, Jacob Eisenstein, and Marjan Ghazvininejad.
\newblock Training on synthetic noise improves robustness to natural noise in
  machine translation.
\newblock \emph{arXiv preprint arXiv:1902.01509}, 2019.

\bibitem[Ke et~al.(2021)Ke, Ji, Ran, Cui, Wang, Song, Zhu, and
  Huang]{ke2021jointgt}
Pei Ke, Haozhe Ji, Yu~Ran, Xin Cui, Liwei Wang, Linfeng Song, Xiaoyan Zhu, and
  Minlie Huang.
\newblock Jointgt: Graph-text joint representation learning for text generation
  from knowledge graphs.
\newblock \emph{arXiv preprint arXiv:2106.10502}, 2021.

\bibitem[Kiela(2017)]{kiela2017deep}
Douwe Kiela.
\newblock Deep embodiment: grounding semantics in perceptual modalities.
\newblock Technical report, University of Cambridge, Computer Laboratory, 2017.

\bibitem[Kim et~al.(2020)Kim, Kim, and Kim]{DBLP:conf/emnlp/KimKK20}
Hyunwoo Kim, Byeongchang Kim, and Gunhee Kim.
\newblock Will {I} sound like me? improving persona consistency in dialogues
  through pragmatic self-consciousness.
\newblock In Bonnie Webber, Trevor Cohn, Yulan He, and Yang Liu, editors,
  \emph{Proceedings of the 2020 Conference on Empirical Methods in Natural
  Language Processing, {EMNLP} 2020, Online, November 16-20, 2020}, pages
  904--916. Association for Computational Linguistics, 2020.
\newblock \doi{10.18653/v1/2020.emnlp-main.65}.
\newblock URL \url{https://doi.org/10.18653/v1/2020.emnlp-main.65}.

\bibitem[Koehn(2009)]{koehn2009statistical}
Philipp Koehn.
\newblock \emph{Statistical machine translation}.
\newblock Cambridge University Press, 2009.

\bibitem[Kong et~al.(2019)Kong, Tu, Shi, Hovy, and Zhang]{kong2019neural}
Xiang Kong, Zhaopeng Tu, Shuming Shi, Eduard Hovy, and Tong Zhang.
\newblock Neural machine translation with adequacy-oriented learning.
\newblock In \emph{Proceedings of the AAAI Conference on Artificial
  Intelligence}, volume~33, pages 6618--6625, 2019.

\bibitem[Koto et~al.(2021)Koto, Baldwin, and Lau]{koto_ffci_2021}
Fajri Koto, Timothy Baldwin, and Jey~Han Lau.
\newblock {FFCI}: {A} {Framework} for {Interpretable} {Automatic} {Evaluation}
  of {Summarization}.
\newblock \emph{arXiv:2011.13662 [cs]}, July 2021.
\newblock URL \url{http://arxiv.org/abs/2011.13662}.
\newblock arXiv: 2011.13662.

\bibitem[Kryściński et~al.(2019)Kryściński, McCann, Xiong, and
  Socher]{kryscinski_evaluating_2019}
Wojciech Kryściński, Bryan McCann, Caiming Xiong, and Richard Socher.
\newblock Evaluating the {Factual} {Consistency} of {Abstractive} {Text}
  {Summarization}.
\newblock \emph{arXiv:1910.12840 [cs]}, October 2019.
\newblock URL \url{http://arxiv.org/abs/1910.12840}.
\newblock arXiv: 1910.12840.

\bibitem[Laban et~al.(2021{\natexlab{a}})Laban, Schnabel, Bennett, and
  Hearst]{DBLP:journals/corr/abs-2111-09525}
Philippe Laban, Tobias Schnabel, Paul~N. Bennett, and Marti~A. Hearst.
\newblock Summac: Re-visiting nli-based models for inconsistency detection in
  summarization.
\newblock \emph{CoRR}, abs/2111.09525, 2021{\natexlab{a}}.
\newblock URL \url{https://arxiv.org/abs/2111.09525}.

\bibitem[Laban et~al.(2021{\natexlab{b}})Laban, Schnabel, Bennett, and
  Hearst]{laban_summac_2021}
Philippe Laban, Tobias Schnabel, Paul~N. Bennett, and Marti~A. Hearst.
\newblock {SummaC}: {Re}-{Visiting} {NLI}-based {Models} for {Inconsistency}
  {Detection} in {Summarization}.
\newblock \emph{arXiv:2111.09525 [cs]}, November 2021{\natexlab{b}}.
\newblock URL \url{http://arxiv.org/abs/2111.09525}.
\newblock arXiv: 2111.09525.

\bibitem[Lebret et~al.(2016)Lebret, Grangier, and Auli]{lebret2016NeuralText}
R{\'e}mi Lebret, David Grangier, and Michael Auli.
\newblock Neural {{Text Generation}} from {{Structured Data}} with
  {{Application}} to the {{Biography Domain}}.
\newblock In \emph{Proceedings of the 2016 {{Conference}} on {{Empirical
  Methods}} in {{Natural Language Processing}}}, pages 1203--1213, {Austin,
  Texas}, 2016. {Association for Computational Linguistics}.
\newblock \doi{10.18653/v1/D16-1128}.

\bibitem[Lee et~al.(2018)Lee, Firat, Agarwal, Fannjiang, and
  Sussillo]{lee2018hallucinations}
Katherine Lee, Orhan Firat, Ashish Agarwal, Clara Fannjiang, and David
  Sussillo.
\newblock Hallucinations in neural machine translation.
\newblock 2018.

\bibitem[Lee et~al.(2021)Lee, Lee, and Hwang]{lee_contrastive_2021}
Seanie Lee, Dong~Bok Lee, and Sung~Ju Hwang.
\newblock {CONTRASTIVE} {LEARNING} {WITH} {ADVERSARIAL} {PER}- {TURBATIONS}
  {FOR} {CONDITIONAL} {TEXT} {GENERATION}.
\newblock page~25, 2021.

\bibitem[Lewis et~al.(2019)Lewis, Liu, Goyal, Ghazvininejad, Mohamed, Levy,
  Stoyanov, and Zettlemoyer]{lewis2019bart}
Mike Lewis, Yinhan Liu, Naman Goyal, Marjan Ghazvininejad, Abdelrahman Mohamed,
  Omer Levy, Ves Stoyanov, and Luke Zettlemoyer.
\newblock Bart: Denoising sequence-to-sequence pre-training for natural
  language generation, translation, and comprehension.
\newblock \emph{arXiv preprint arXiv:1910.13461}, 2019.

\bibitem[Li et~al.(2018{\natexlab{a}})Li, Xu, Li, and Gao]{li2018guiding}
Chenliang Li, Weiran Xu, Si~Li, and Sheng Gao.
\newblock Guiding generation for abstractive text summarization based on key
  information guide network.
\newblock In \emph{Proceedings of the 2018 Conference of the North American
  Chapter of the Association for Computational Linguistics: Human Language
  Technologies, Volume 2 (Short Papers)}, pages 55--60, 2018{\natexlab{a}}.

\bibitem[Li et~al.(2018{\natexlab{b}})Li, Zhu, Zhang, and Zong]{li_ensure_2018}
Haoran Li, Junnan Zhu, Jiajun Zhang, and Chengqing Zong.
\newblock Ensure the {Correctness} of the {Summary}: {Incorporate} {Entailment}
  {Knowledge} into {Abstractive} {Sentence} {Summarization}.
\newblock In \emph{Proceedings of the 27th {International} {Conference} on
  {Computational} {Linguistics}}, pages 1430--1441, Santa Fe, New Mexico, USA,
  August 2018{\natexlab{b}}. Association for Computational Linguistics.
\newblock URL \url{https://aclanthology.org/C18-1121}.

\bibitem[Li et~al.(2016{\natexlab{a}})Li, Galley, Brockett, Spithourakis, Gao,
  and Dolan]{DBLP:conf/acl/LiGBSGD16}
Jiwei Li, Michel Galley, Chris Brockett, Georgios~P. Spithourakis, Jianfeng
  Gao, and William~B. Dolan.
\newblock A persona-based neural conversation model.
\newblock In \emph{Proceedings of the 54th Annual Meeting of the Association
  for Computational Linguistics, {ACL} 2016, August 7-12, 2016, Berlin,
  Germany, Volume 1: Long Papers}. The Association for Computer Linguistics,
  2016{\natexlab{a}}.
\newblock \doi{10.18653/v1/p16-1094}.
\newblock URL \url{https://doi.org/10.18653/v1/p16-1094}.

\bibitem[Li et~al.(2016{\natexlab{b}})Li, Monroe, Ritter, Galley, Gao, and
  Jurafsky]{li2016deep}
Jiwei Li, Will Monroe, Alan Ritter, Michel Galley, Jianfeng Gao, and Dan
  Jurafsky.
\newblock Deep reinforcement learning for dialogue generation.
\newblock \emph{arXiv preprint arXiv:1606.01541}, 2016{\natexlab{b}}.

\bibitem[Li et~al.(2020{\natexlab{a}})Li, Roller, Kulikov, Welleck, Boureau,
  Cho, and Weston]{DBLP:conf/acl/LiRKWBCW20}
Margaret Li, Stephen Roller, Ilia Kulikov, Sean Welleck, Y{-}Lan Boureau,
  Kyunghyun Cho, and Jason Weston.
\newblock Don't say that! making inconsistent dialogue unlikely with
  unlikelihood training.
\newblock In Dan Jurafsky, Joyce Chai, Natalie Schluter, and Joel~R. Tetreault,
  editors, \emph{Proceedings of the 58th Annual Meeting of the Association for
  Computational Linguistics, {ACL} 2020, Online, July 5-10, 2020}, pages
  4715--4728. Association for Computational Linguistics, 2020{\natexlab{a}}.
\newblock \doi{10.18653/v1/2020.acl-main.428}.
\newblock URL \url{https://doi.org/10.18653/v1/2020.acl-main.428}.

\bibitem[Li et~al.(2020{\natexlab{b}})Li, Beirami, Sanjabi, and
  Smith]{li2020tilted}
Tian Li, Ahmad Beirami, Maziar Sanjabi, and Virginia Smith.
\newblock Tilted empirical risk minimization.
\newblock \emph{arXiv preprint arXiv:2007.01162}, 2020{\natexlab{b}}.

\bibitem[Li et~al.(2020{\natexlab{c}})Li, Gao, Niu, Xiao, Liu, Liu, Wu, and
  Wang]{li2020unimo}
Wei Li, Can Gao, Guocheng Niu, Xinyan Xiao, Hao Liu, Jiachen Liu, Hua Wu, and
  Haifeng Wang.
\newblock Unimo: Towards unified-modal understanding and generation via
  cross-modal contrastive learning.
\newblock \emph{arXiv preprint arXiv:2012.15409}, 2020{\natexlab{c}}.

\bibitem[Li et~al.(2018{\natexlab{c}})Li, Liang, Hu, and Xing]{li2018hybrid}
Yuan Li, Xiaodan Liang, Zhiting Hu, and Eric~P Xing.
\newblock Hybrid retrieval-generation reinforced agent for medical image report
  generation.
\newblock \emph{Advances in neural information processing systems}, 31,
  2018{\natexlab{c}}.

\bibitem[Li et~al.(2017)Li, Jiang, Shang, and Li]{li2017paraphrase}
Zichao Li, Xin Jiang, Lifeng Shang, and Hang Li.
\newblock Paraphrase generation with deep reinforcement learning.
\newblock \emph{arXiv preprint arXiv:1711.00279}, 2017.

\bibitem[Lin(2004)]{lin2004rouge}
Chin-Yew Lin.
\newblock Rouge: A package for automatic evaluation of summaries.
\newblock In \emph{Text summarization branches out}, pages 74--81, 2004.

\bibitem[Liu et~al.(2019{\natexlab{a}})Liu, Hsu, McDermott, Boag, Weng,
  Szolovits, and Ghassemi]{liu2019clinically}
Guanxiong Liu, Tzu-Ming~Harry Hsu, Matthew McDermott, Willie Boag, Wei-Hung
  Weng, Peter Szolovits, and Marzyeh Ghassemi.
\newblock Clinically accurate chest x-ray report generation.
\newblock In \emph{Machine Learning for Healthcare Conference}, pages 249--269.
  PMLR, 2019{\natexlab{a}}.

\bibitem[Liu et~al.(2018{\natexlab{a}})Liu, Ma, Huang, Xiong, and
  He]{liu2018robust}
Hairong Liu, Mingbo Ma, Liang Huang, Hao Xiong, and Zhongjun He.
\newblock Robust neural machine translation with joint textual and phonetic
  embedding.
\newblock \emph{arXiv preprint arXiv:1810.06729}, 2018{\natexlab{a}}.

\bibitem[Liu et~al.(2018{\natexlab{b}})Liu, Wang, Sha, Chang, and
  Sui]{liu2018table}
Tianyu Liu, Kexiang Wang, Lei Sha, Baobao Chang, and Zhifang Sui.
\newblock Table-to-text generation by structure-aware seq2seq learning.
\newblock In \emph{Thirty-Second AAAI Conference on Artificial Intelligence},
  2018{\natexlab{b}}.

\bibitem[Liu et~al.(2019{\natexlab{b}})Liu, Luo, Yang, Wu, Chang, and
  Sui]{liu2019ComprehensiveDescription}
Tianyu Liu, Fuli Luo, Pengcheng Yang, Wei Wu, Baobao Chang, and Zhifang Sui.
\newblock Towards {{Comprehensive Description Generation}} from {{Factual
  Attribute}}-value {{Tables}}.
\newblock In \emph{Proceedings of the 57th {{Annual Meeting}} of the
  {{Association}} for {{Computational Linguistics}}}, pages 5985--5996,
  {Florence, Italy}, 2019{\natexlab{b}}. {Association for Computational
  Linguistics}.
\newblock \doi{10.18653/v1/P19-1600}.

\bibitem[Liu et~al.(2021{\natexlab{a}})Liu, Zheng, Chang, and
  Sui]{liu2021FaithfulnessOpen}
Tianyu Liu, Xin Zheng, Baobao Chang, and Zhifang Sui.
\newblock Towards {{Faithfulness}} in {{Open Domain Table}}-to-text
  {{Generation}} from an {{Entity}}-centric {{View}}.
\newblock In \emph{Proceedings of the {{AAAI Conference}} on {{Artificial
  Intelligence}}}, volume~35, pages 13415--13423, May 2021{\natexlab{a}}.

\bibitem[Liu et~al.(2021{\natexlab{b}})Liu, Wu, Mu, Li, Chen, and
  Nie]{liu2021co2sum}
Wei Liu, Huanqin Wu, Wenjing Mu, Zhen Li, Tao Chen, and Dan Nie.
\newblock Co2sum: Contrastive learning for factual-consistent abstractive
  summarization.
\newblock \emph{arXiv preprint arXiv:2112.01147}, 2021{\natexlab{b}}.

\bibitem[Liu and Lapata(2019)]{liu2019text}
Yang Liu and Mirella Lapata.
\newblock Text summarization with pretrained encoders.
\newblock \emph{arXiv preprint arXiv:1908.08345}, 2019.

\bibitem[Liu et~al.(2021{\natexlab{c}})Liu, Sun, and Gao]{liu_improving_2021}
Yang Liu, Yifei Sun, and Vincent Gao.
\newblock Improving {Factual} {Consistency} of {Abstractive} {Summarization} on
  {Customer} {Feedback}.
\newblock \emph{arXiv:2106.16188 [cs]}, June 2021{\natexlab{c}}.
\newblock URL \url{http://arxiv.org/abs/2106.16188}.
\newblock arXiv: 2106.16188.

\bibitem[Liu et~al.(2019{\natexlab{c}})Liu, Ott, Goyal, Du, Joshi, Chen, Levy,
  Lewis, Zettlemoyer, and Stoyanov]{liu2019roberta}
Yinhan Liu, Myle Ott, Naman Goyal, Jingfei Du, Mandar Joshi, Danqi Chen, Omer
  Levy, Mike Lewis, Luke Zettlemoyer, and Veselin Stoyanov.
\newblock Roberta: A robustly optimized bert pretraining approach.
\newblock \emph{arXiv preprint arXiv:1907.11692}, 2019{\natexlab{c}}.

\bibitem[Liu et~al.(2020)Liu, Gu, Goyal, Li, Edunov, Ghazvininejad, Lewis, and
  Zettlemoyer]{liu2020multilingual}
Yinhan Liu, Jiatao Gu, Naman Goyal, Xian Li, Sergey Edunov, Marjan
  Ghazvininejad, Mike Lewis, and Luke Zettlemoyer.
\newblock Multilingual denoising pre-training for neural machine translation.
\newblock \emph{Transactions of the Association for Computational Linguistics},
  8:\penalty0 726--742, 2020.

\bibitem[Logan~IV et~al.(2019)Logan~IV, Liu, Peters, Gardner, and
  Singh]{logan2019barack}
Robert~L Logan~IV, Nelson~F Liu, Matthew~E Peters, Matt Gardner, and Sameer
  Singh.
\newblock Barack's wife hillary: Using knowledge-graphs for fact-aware language
  modeling.
\newblock \emph{arXiv preprint arXiv:1906.07241}, 2019.

\bibitem[Longpre et~al.(2021)Longpre, Perisetla, Chen, Ramesh, DuBois, and
  Singh]{longpre2021entity}
Shayne Longpre, Kartik Perisetla, Anthony Chen, Nikhil Ramesh, Chris DuBois,
  and Sameer Singh.
\newblock Entity-based knowledge conflicts in question answering.
\newblock \emph{arXiv preprint arXiv:2109.05052}, 2021.

\bibitem[Ma et~al.(2002)Ma, Lu, Zhang, and Li]{ma2002user}
Yu-Fei Ma, Lie Lu, Hong-Jiang Zhang, and Mingjing Li.
\newblock A user attention model for video summarization.
\newblock In \emph{Proceedings of the tenth ACM international conference on
  Multimedia}, pages 533--542, 2002.

\bibitem[Mao et~al.(2020)Mao, Ren, Ji, and Han]{mao2020constrained}
Yuning Mao, Xiang Ren, Heng Ji, and Jiawei Han.
\newblock Constrained abstractive summarization: Preserving factual consistency
  with constrained generation.
\newblock \emph{arXiv preprint arXiv:2010.12723}, 2020.

\bibitem[Maynez et~al.(2020)Maynez, Narayan, Bohnet, and
  McDonald]{maynez_faithfulness_2020}
Joshua Maynez, Shashi Narayan, Bernd Bohnet, and Ryan McDonald.
\newblock On {Faithfulness} and {Factuality} in {Abstractive} {Summarization}.
\newblock In \emph{Proceedings of the 58th {Annual} {Meeting} of the
  {Association} for {Computational} {Linguistics}}, pages 1906--1919, Online,
  July 2020. Association for Computational Linguistics.
\newblock \doi{10.18653/v1/2020.acl-main.173}.
\newblock URL \url{https://www.aclweb.org/anthology/2020.acl-main.173}.

\bibitem[Meng et~al.(2020)Meng, Jimenez, Arslan, Devasier, Obembe, and
  Li]{meng2020gradient}
Kevin Meng, Damian Jimenez, Fatma Arslan, Jacob~Daniel Devasier, Daniel Obembe,
  and Chengkai Li.
\newblock Gradient-based adversarial training on transformer networks for
  detecting check-worthy factual claims.
\newblock \emph{arXiv preprint arXiv:2002.07725}, 2020.

\bibitem[Mesgar et~al.(2021)Mesgar, Simpson, and
  Gurevych]{mesgar_improving_2021}
Mohsen Mesgar, Edwin Simpson, and Iryna Gurevych.
\newblock Improving {Factual} {Consistency} {Between} a {Response} and
  {Persona} {Facts}.
\newblock In \emph{Proceedings of the 16th {Conference} of the {European}
  {Chapter} of the {Association} for {Computational} {Linguistics}: {Main}
  {Volume}}, pages 549--562, Online, April 2021. Association for Computational
  Linguistics.
\newblock \doi{10.18653/v1/2021.eacl-main.44}.
\newblock URL \url{https://aclanthology.org/2021.eacl-main.44}.

\bibitem[Mi et~al.(2016)Mi, Sankaran, Wang, and Ittycheriah]{mi2016coverage}
Haitao Mi, Baskaran Sankaran, Zhiguo Wang, and Abe Ittycheriah.
\newblock Coverage embedding models for neural machine translation.
\newblock \emph{arXiv preprint arXiv:1605.03148}, 2016.

\bibitem[Miao et~al.(2019)Miao, Zhou, Mou, Yan, and Li]{miao2019cgmh}
Ning Miao, Hao Zhou, Lili Mou, Rui Yan, and Lei Li.
\newblock Cgmh: Constrained sentence generation by metropolis-hastings
  sampling.
\newblock In \emph{Proceedings of the AAAI Conference on Artificial
  Intelligence}, volume~33, pages 6834--6842, 2019.

\bibitem[Mishra et~al.(2021)Mishra, Patel, Vijayakumar, Li, Kapanipathi, and
  Talamadupula]{mishra_looking_2021}
Anshuman Mishra, Dhruvesh Patel, Aparna Vijayakumar, Xiang~Lorraine Li, Pavan
  Kapanipathi, and Kartik Talamadupula.
\newblock Looking {Beyond} {Sentence}-{Level} {Natural} {Language} {Inference}
  for {Question} {Answering} and {Text} {Summarization}.
\newblock In \emph{Proceedings of the 2021 {Conference} of the {North}
  {American} {Chapter} of the {Association} for {Computational} {Linguistics}:
  {Human} {Language} {Technologies}}, pages 1322--1336, Online, 2021.
  Association for Computational Linguistics.
\newblock \doi{10.18653/v1/2021.naacl-main.104}.
\newblock URL \url{https://aclanthology.org/2021.naacl-main.104}.

\bibitem[Miura et~al.(2021)Miura, Zhang, Tsai, Langlotz, and
  Jurafsky]{miura_improving_2021}
Yasuhide Miura, Yuhao Zhang, Emily~Bao Tsai, Curtis~P. Langlotz, and Dan
  Jurafsky.
\newblock Improving {Factual} {Completeness} and {Consistency} of
  {Image}-to-{Text} {Radiology} {Report} {Generation}.
\newblock \emph{arXiv:2010.10042 [cs]}, April 2021.
\newblock URL \url{http://arxiv.org/abs/2010.10042}.
\newblock arXiv: 2010.10042.

\bibitem[Miyato et~al.(2016)Miyato, Dai, and Goodfellow]{miyato2016adversarial}
Takeru Miyato, Andrew~M Dai, and Ian Goodfellow.
\newblock Adversarial training methods for semi-supervised text classification.
\newblock \emph{arXiv preprint arXiv:1605.07725}, 2016.

\bibitem[Mnih et~al.(2016)Mnih, Badia, Mirza, Graves, Lillicrap, Harley,
  Silver, and Kavukcuoglu]{DBLP:conf/icml/MnihBMGLHSK16}
Volodymyr Mnih, Adri{\`{a}}~Puigdom{\`{e}}nech Badia, Mehdi Mirza, Alex Graves,
  Timothy~P. Lillicrap, Tim Harley, David Silver, and Koray Kavukcuoglu.
\newblock Asynchronous methods for deep reinforcement learning.
\newblock In Maria{-}Florina Balcan and Kilian~Q. Weinberger, editors,
  \emph{Proceedings of the 33nd International Conference on Machine Learning,
  {ICML} 2016, New York City, NY, USA, June 19-24, 2016}, volume~48 of
  \emph{{JMLR} Workshop and Conference Proceedings}, pages 1928--1937.
  JMLR.org, 2016.
\newblock URL \url{http://proceedings.mlr.press/v48/mniha16.html}.

\bibitem[M{\"u}ller et~al.(2019)M{\"u}ller, Rios, and
  Sennrich]{muller2019domain}
Mathias M{\"u}ller, Annette Rios, and Rico Sennrich.
\newblock Domain robustness in neural machine translation.
\newblock \emph{arXiv preprint arXiv:1911.03109}, 2019.

\bibitem[Nan et~al.(2021{\natexlab{a}})Nan, dos Santos, Zhu, Ng, McKeown,
  Nallapati, Zhang, Wang, Arnold, and Xiang]{DBLP:conf/acl/NanSZNMNZWAX20}
Feng Nan, C{\'{\i}}cero~Nogueira dos Santos, Henghui Zhu, Patrick Ng,
  Kathleen~R. McKeown, Ramesh Nallapati, Dejiao Zhang, Zhiguo Wang, Andrew~O.
  Arnold, and Bing Xiang.
\newblock Improving factual consistency of abstractive summarization via
  question answering.
\newblock In Chengqing Zong, Fei Xia, Wenjie Li, and Roberto Navigli, editors,
  \emph{Proceedings of the 59th Annual Meeting of the Association for
  Computational Linguistics and the 11th International Joint Conference on
  Natural Language Processing, {ACL/IJCNLP} 2021, (Volume 1: Long Papers),
  Virtual Event, August 1-6, 2021}, pages 6881--6894. Association for
  Computational Linguistics, 2021{\natexlab{a}}.
\newblock \doi{10.18653/v1/2021.acl-long.536}.
\newblock URL \url{https://doi.org/10.18653/v1/2021.acl-long.536}.

\bibitem[Nan et~al.(2021{\natexlab{b}})Nan, Nallapati, Wang, dos Santos, Zhu,
  Zhang, Mckeown, and Xiang]{nan2021entity}
Feng Nan, Ramesh Nallapati, Zhiguo Wang, Cicero dos Santos, Henghui Zhu, Dejiao
  Zhang, Kathleen Mckeown, and Bing Xiang.
\newblock Entity-level factual consistency of abstractive text summarization.
\newblock In \emph{Proceedings of the 16th Conference of the European Chapter
  of the Association for Computational Linguistics: Main Volume}, pages
  2727--2733, 2021{\natexlab{b}}.

\bibitem[Nie et~al.(2019)Nie, Yao, Wang, Pan, and Lin]{nie2019SimpleRecipe}
Feng Nie, Jin-Ge Yao, Jinpeng Wang, Rong Pan, and Chin-Yew Lin.
\newblock A {{Simple Recipe}} towards {{Reducing Hallucination}} in {{Neural
  Surface Realisation}}.
\newblock In \emph{Proceedings of the 57th {{Annual Meeting}} of the
  {{Association}} for {{Computational Linguistics}}}, pages 2673--2679,
  {Florence, Italy}, 2019. {Association for Computational Linguistics}.
\newblock \doi{10.18653/v1/P19-1256}.

\bibitem[Nie et~al.(2021)Nie, Williamson, Bansal, Kiela, and
  Weston]{DBLP:conf/acl/NieWBKW20}
Yixin Nie, Mary Williamson, Mohit Bansal, Douwe Kiela, and Jason Weston.
\newblock I like fish, especially dolphins: Addressing contradictions in
  dialogue modeling.
\newblock In Chengqing Zong, Fei Xia, Wenjie Li, and Roberto Navigli, editors,
  \emph{Proceedings of the 59th Annual Meeting of the Association for
  Computational Linguistics and the 11th International Joint Conference on
  Natural Language Processing, {ACL/IJCNLP} 2021, (Volume 1: Long Papers),
  Virtual Event, August 1-6, 2021}, pages 1699--1713. Association for
  Computational Linguistics, 2021.
\newblock \doi{10.18653/v1/2021.acl-long.134}.
\newblock URL \url{https://doi.org/10.18653/v1/2021.acl-long.134}.

\bibitem[Nye et~al.(2021)Nye, Tessler, Tenenbaum, and Lake]{nye2021improving}
Maxwell Nye, Michael Tessler, Josh Tenenbaum, and Brenden~M Lake.
\newblock Improving coherence and consistency in neural sequence models with
  dual-system, neuro-symbolic reasoning.
\newblock \emph{Advances in Neural Information Processing Systems}, 34, 2021.

\bibitem[Pagnoni et~al.(2021)Pagnoni, Balachandran, and
  Tsvetkov]{pagnoni_understanding_2021}
Artidoro Pagnoni, Vidhisha Balachandran, and Yulia Tsvetkov.
\newblock Understanding {Factuality} in {Abstractive} {Summarization} with
  {FRANK}: {A} {Benchmark} for {Factuality} {Metrics}.
\newblock In \emph{Proceedings of the 2021 {Conference} of the {North}
  {American} {Chapter} of the {Association} for {Computational} {Linguistics}:
  {Human} {Language} {Technologies}}, pages 4812--4829, Online, June 2021.
  Association for Computational Linguistics.
\newblock URL \url{https://www.aclweb.org/anthology/2021.naacl-main.383}.

\bibitem[Palmer et~al.(2005)Palmer, Gildea, and
  Kingsbury]{palmer2005proposition}
Martha Palmer, Daniel Gildea, and Paul Kingsbury.
\newblock The proposition bank: An annotated corpus of semantic roles.
\newblock \emph{Computational linguistics}, 31\penalty0 (1):\penalty0 71--106,
  2005.

\bibitem[Papineni et~al.(2002)Papineni, Roukos, Ward, and
  Zhu]{papineni2002bleu}
Kishore Papineni, Salim Roukos, Todd Ward, and Wei-Jing Zhu.
\newblock Bleu: a method for automatic evaluation of machine translation.
\newblock In \emph{Proceedings of the 40th annual meeting of the Association
  for Computational Linguistics}, pages 311--318, 2002.

\bibitem[Post and Vilar(2018)]{post_fast_2018}
Matt Post and David Vilar.
\newblock Fast {Lexically} {Constrained} {Decoding} with {Dynamic} {Beam}
  {Allocation} for {Neural} {Machine} {Translation}.
\newblock In \emph{Proceedings of the 2018 {Conference} of the {North}
  {American} {Chapter} of the {Association} for {Computational} {Linguistics}:
  {Human} {Language} {Technologies}, {Volume} 1 ({Long} {Papers})}, pages
  1314--1324, New Orleans, Louisiana, June 2018. Association for Computational
  Linguistics.
\newblock \doi{10.18653/v1/N18-1119}.
\newblock URL \url{https://aclanthology.org/N18-1119}.

\bibitem[Puduppully et~al.(2019)Puduppully, Dong, and
  Lapata]{puduppully2019DatatotextGeneration}
Ratish Puduppully, Li~Dong, and Mirella Lapata.
\newblock Data-to-text {{Generation}} with {{Entity Modeling}}.
\newblock In \emph{Proceedings of the 57th {{Annual Meeting}} of the
  {{Association}} for {{Computational Linguistics}}}, pages 2023--2035,
  {Florence, Italy}, 2019. {Association for Computational Linguistics}.
\newblock \doi{10.18653/v1/P19-1195}.

\bibitem[Qin et~al.(2019)Qin, Galley, Brockett, Liu, Gao, Dolan, Choi, and
  Gao]{qin2019conversing}
Lianhui Qin, Michel Galley, Chris Brockett, Xiaodong Liu, Xiang Gao, Bill
  Dolan, Yejin Choi, and Jianfeng Gao.
\newblock Conversing by reading: Contentful neural conversation with on-demand
  machine reading.
\newblock \emph{arXiv preprint arXiv:1906.02738}, 2019.

\bibitem[Qin et~al.(2021)Qin, Xie, Huang, Chen, Xu, and
  Che]{DBLP:conf/emnlp/QinXHCXC21}
Libo Qin, Tianbao Xie, Shijue Huang, Qiguang Chen, Xiao Xu, and Wanxiang Che.
\newblock Don't be contradicted with anything! ci-tod: Towards benchmarking
  consistency for task-oriented dialogue system.
\newblock In Marie{-}Francine Moens, Xuanjing Huang, Lucia Specia, and
  Scott~Wen{-}tau Yih, editors, \emph{Proceedings of the 2021 Conference on
  Empirical Methods in Natural Language Processing, {EMNLP} 2021, Virtual Event
  / Punta Cana, Dominican Republic, 7-11 November, 2021}, pages 2357--2367.
  Association for Computational Linguistics, 2021.
\newblock \doi{10.18653/v1/2021.emnlp-main.182}.
\newblock URL \url{https://doi.org/10.18653/v1/2021.emnlp-main.182}.

\bibitem[Radford et~al.(2019)Radford, Wu, Child, Luan, Amodei, Sutskever,
  et~al.]{radford2019language}
Alec Radford, Jeffrey Wu, Rewon Child, David Luan, Dario Amodei, Ilya
  Sutskever, et~al.
\newblock Language models are unsupervised multitask learners.
\newblock \emph{OpenAI blog}, 1\penalty0 (8):\penalty0 9, 2019.

\bibitem[Raffel et~al.(2019)Raffel, Shazeer, Roberts, Lee, Narang, Matena,
  Zhou, Li, and Liu]{raffel2019exploring}
Colin Raffel, Noam Shazeer, Adam Roberts, Katherine Lee, Sharan Narang, Michael
  Matena, Yanqi Zhou, Wei Li, and Peter~J Liu.
\newblock Exploring the limits of transfer learning with a unified text-to-text
  transformer.
\newblock \emph{arXiv preprint arXiv:1910.10683}, 2019.

\bibitem[Ranzato et~al.(2015)Ranzato, Chopra, Auli, and
  Zaremba]{ranzato2015sequence}
Marc'Aurelio Ranzato, Sumit Chopra, Michael Auli, and Wojciech Zaremba.
\newblock Sequence level training with recurrent neural networks.
\newblock \emph{arXiv preprint arXiv:1511.06732}, 2015.

\bibitem[Rashkin et~al.(2021)Rashkin, Reitter, Tomar, and
  Das]{DBLP:conf/acl/RashkinRT020}
Hannah Rashkin, David Reitter, Gaurav~Singh Tomar, and Dipanjan Das.
\newblock Increasing faithfulness in knowledge-grounded dialogue with
  controllable features.
\newblock In Chengqing Zong, Fei Xia, Wenjie Li, and Roberto Navigli, editors,
  \emph{Proceedings of the 59th Annual Meeting of the Association for
  Computational Linguistics and the 11th International Joint Conference on
  Natural Language Processing, {ACL/IJCNLP} 2021, (Volume 1: Long Papers),
  Virtual Event, August 1-6, 2021}, pages 704--718. Association for
  Computational Linguistics, 2021.
\newblock \doi{10.18653/v1/2021.acl-long.58}.
\newblock URL \url{https://doi.org/10.18653/v1/2021.acl-long.58}.

\bibitem[Raunak et~al.(2021)Raunak, Menezes, and
  Junczys-Dowmunt]{raunak2021curious}
Vikas Raunak, Arul Menezes, and Marcin Junczys-Dowmunt.
\newblock The curious case of hallucinations in neural machine translation.
\newblock \emph{arXiv preprint arXiv:2104.06683}, 2021.

\bibitem[Rebuffel et~al.(2020)Rebuffel, Soulier, Scoutheeten, and
  Gallinari]{rebuffel2020PARENTingModelAgnostic}
Clement Rebuffel, Laure Soulier, Geoffrey Scoutheeten, and Patrick Gallinari.
\newblock {{PARENTing}} via {{Model}}-{{Agnostic Reinforcement Learning}} to
  {{Correct Pathological Behaviors}} in {{Data}}-to-{{Text Generation}}.
\newblock In \emph{Proceedings of the 13th {{International Conference}} on
  {{Natural Language Generation}}}, pages 120--130, {Dublin, Ireland}, 2020.
  {Association for Computational Linguistics}.

\bibitem[Rebuffel et~al.(2022)Rebuffel, Roberti, Soulier, Scoutheeten,
  Cancelliere, and Gallinari]{rebuffel2022ControllingHallucinations}
Clement Rebuffel, Marco Roberti, Laure Soulier, Geoffrey Scoutheeten, Rossella
  Cancelliere, and Patrick Gallinari.
\newblock Controlling hallucinations at word level in data-to-text generation.
\newblock \emph{Data Mining and Knowledge Discovery}, 36\penalty0 (1):\penalty0
  318--354, 2022.
\newblock ISSN 1573-756X.
\newblock \doi{10.1007/s10618-021-00801-4}.

\bibitem[Reis et~al.(2019)Reis, Correia, Murai, Veloso, and
  Benevenuto]{reis2019supervised}
Julio~CS Reis, Andr{\'e} Correia, Fabr{\'\i}cio Murai, Adriano Veloso, and
  Fabr{\'\i}cio Benevenuto.
\newblock Supervised learning for fake news detection.
\newblock \emph{IEEE Intelligent Systems}, 34\penalty0 (2):\penalty0 76--81,
  2019.

\bibitem[Rohrbach et~al.(2018)Rohrbach, Hendricks, Burns, Darrell, and
  Saenko]{rohrbach_object_2018}
Anna Rohrbach, Lisa~Anne Hendricks, Kaylee Burns, Trevor Darrell, and Kate
  Saenko.
\newblock Object {Hallucination} in {Image} {Captioning}.
\newblock In \emph{Proceedings of the 2018 {Conference} on {Empirical}
  {Methods} in {Natural} {Language} {Processing}}, pages 4035--4045, Brussels,
  Belgium, October 2018. Association for Computational Linguistics.
\newblock \doi{10.18653/v1/D18-1437}.
\newblock URL \url{https://aclanthology.org/D18-1437}.

\bibitem[Roller et~al.(2021)Roller, Dinan, Goyal, Ju, Williamson, Liu, Xu, Ott,
  Smith, Boureau, and Weston]{DBLP:conf/eacl/RollerDGJWLXOSB21}
Stephen Roller, Emily Dinan, Naman Goyal, Da~Ju, Mary Williamson, Yinhan Liu,
  Jing Xu, Myle Ott, Eric~Michael Smith, Y{-}Lan Boureau, and Jason Weston.
\newblock Recipes for building an open-domain chatbot.
\newblock In Paola Merlo, J{\"{o}}rg Tiedemann, and Reut Tsarfaty, editors,
  \emph{Proceedings of the 16th Conference of the European Chapter of the
  Association for Computational Linguistics: Main Volume, {EACL} 2021, Online,
  April 19 - 23, 2021}, pages 300--325. Association for Computational
  Linguistics, 2021.
\newblock \doi{10.18653/v1/2021.eacl-main.24}.
\newblock URL \url{https://doi.org/10.18653/v1/2021.eacl-main.24}.

\bibitem[Sabet et~al.(2020)Sabet, Dufter, Yvon, and
  Sch{\"u}tze]{sabet2020simalign}
Masoud~Jalili Sabet, Philipp Dufter, Fran{\c{c}}ois Yvon, and Hinrich
  Sch{\"u}tze.
\newblock Simalign: High quality word alignments without parallel training data
  using static and contextualized embeddings.
\newblock \emph{arXiv preprint arXiv:2004.08728}, 2020.

\bibitem[Saito et~al.(2020)Saito, Nishida, Nishida, and
  Tomita]{saito2020abstractive}
Itsumi Saito, Kyosuke Nishida, Kosuke Nishida, and Junji Tomita.
\newblock Abstractive summarization with combination of pre-trained
  sequence-to-sequence and saliency models.
\newblock \emph{arXiv preprint arXiv:2003.13028}, 2020.

\bibitem[Scialom et~al.(2021)Scialom, Dray, Gallinari, Lamprier, Piwowarski,
  Staiano, and Wang]{scialom_questeval_2021}
Thomas Scialom, Paul-Alexis Dray, Patrick Gallinari, Sylvain Lamprier, Benjamin
  Piwowarski, Jacopo Staiano, and Alex Wang.
\newblock {QuestEval}: {Summarization} {Asks} for {Fact}-based {Evaluation}.
\newblock \emph{arXiv:2103.12693 [cs]}, April 2021.
\newblock URL \url{http://arxiv.org/abs/2103.12693}.
\newblock arXiv: 2103.12693.

\bibitem[Shen et~al.(2020)Shen, Chang, Su, Niu, and
  Klakow]{shen2020NeuralDatatoText}
Xiaoyu Shen, Ernie Chang, Hui Su, Cheng Niu, and Dietrich Klakow.
\newblock Neural {{Data}}-to-{{Text Generation}} via {{Jointly Learning}} the
  {{Segmentation}} and {{Correspondence}}.
\newblock In \emph{Proceedings of the 58th {{Annual Meeting}} of the
  {{Association}} for {{Computational Linguistics}}}, pages 7155--7165,
  {Online}, 2020. {Association for Computational Linguistics}.
\newblock \doi{10.18653/v1/2020.acl-main.641}.

\bibitem[Shu et~al.(2017)Shu, Sliva, Wang, Tang, and Liu]{shu2017fake}
Kai Shu, Amy Sliva, Suhang Wang, Jiliang Tang, and Huan Liu.
\newblock Fake news detection on social media: A data mining perspective.
\newblock \emph{ACM SIGKDD explorations newsletter}, 19\penalty0 (1):\penalty0
  22--36, 2017.

\bibitem[Shu et~al.(2019)Shu, Wang, and Liu]{shu2019beyond}
Kai Shu, Suhang Wang, and Huan Liu.
\newblock Beyond news contents: The role of social context for fake news
  detection.
\newblock In \emph{Proceedings of the twelfth ACM international conference on
  web search and data mining}, pages 312--320, 2019.

\bibitem[Shuster et~al.(2021)Shuster, Poff, Chen, Kiela, and
  Weston]{DBLP:conf/emnlp/0001PCKW21}
Kurt Shuster, Spencer Poff, Moya Chen, Douwe Kiela, and Jason Weston.
\newblock Retrieval augmentation reduces hallucination in conversation.
\newblock In Marie{-}Francine Moens, Xuanjing Huang, Lucia Specia, and
  Scott~Wen{-}tau Yih, editors, \emph{Findings of the Association for
  Computational Linguistics: {EMNLP} 2021, Virtual Event / Punta Cana,
  Dominican Republic, 16-20 November, 2021}, pages 3784--3803. Association for
  Computational Linguistics, 2021.
\newblock \doi{10.18653/v1/2021.findings-emnlp.320}.
\newblock URL \url{https://doi.org/10.18653/v1/2021.findings-emnlp.320}.

\bibitem[Song et~al.(2020{\natexlab{a}})Song, Wang, Zhang, Zhao, Liu, and
  Liu]{song_profile_2020}
Haoyu Song, Yan Wang, Wei-Nan Zhang, Zhengyu Zhao, Ting Liu, and Xiaojiang Liu.
\newblock Profile {Consistency} {Identification} for {Open}-domain {Dialogue}
  {Agents}.
\newblock In \emph{Proceedings of the 2020 {Conference} on {Empirical}
  {Methods} in {Natural} {Language} {Processing} ({EMNLP})}, pages 6651--6662,
  Online, November 2020{\natexlab{a}}. Association for Computational
  Linguistics.
\newblock \doi{10.18653/v1/2020.emnlp-main.539}.
\newblock URL \url{https://aclanthology.org/2020.emnlp-main.539}.

\bibitem[Song et~al.(2020{\natexlab{b}})Song, Wang, Zhang, Liu, and
  Liu]{DBLP:conf/acl/SongWZLL20}
Haoyu Song, Yan Wang, Weinan Zhang, Xiaojiang Liu, and Ting Liu.
\newblock Generate, delete and rewrite: {A} three-stage framework for improving
  persona consistency of dialogue generation.
\newblock In Dan Jurafsky, Joyce Chai, Natalie Schluter, and Joel~R. Tetreault,
  editors, \emph{Proceedings of the 58th Annual Meeting of the Association for
  Computational Linguistics, {ACL} 2020, Online, July 5-10, 2020}, pages
  5821--5831. Association for Computational Linguistics, 2020{\natexlab{b}}.
\newblock \doi{10.18653/v1/2020.acl-main.516}.
\newblock URL \url{https://doi.org/10.18653/v1/2020.acl-main.516}.

\bibitem[Song et~al.(2020{\natexlab{c}})Song, Zhang, Hu, and
  Liu]{song_generating_2020}
Haoyu Song, Wei-Nan Zhang, Jingwen Hu, and Ting Liu.
\newblock Generating {Persona} {Consistent} {Dialogues} by {Exploiting}
  {Natural} {Language} {Inference}.
\newblock \emph{Proceedings of the AAAI Conference on Artificial Intelligence},
  34\penalty0 (05):\penalty0 8878--8885, April 2020{\natexlab{c}}.
\newblock ISSN 2374-3468, 2159-5399.
\newblock \doi{10.1609/aaai.v34i05.6417}.
\newblock URL \url{http://arxiv.org/abs/1911.05889}.
\newblock arXiv: 1911.05889.

\bibitem[Song et~al.(2018)Song, Zhang, Wang, and Gildea]{song2018graph}
Linfeng Song, Yue Zhang, Zhiguo Wang, and Daniel Gildea.
\newblock A graph-to-sequence model for amr-to-text generation.
\newblock \emph{arXiv preprint arXiv:1805.02473}, 2018.

\bibitem[Song et~al.(2020{\natexlab{d}})Song, Shuai, Yeh, Wu, Ku, and
  Peng]{song_attractive_2020}
Yun-Zhu Song, Hong-Han Shuai, Sung-Lin Yeh, Yi-Lun Wu, Lun-Wei Ku, and Wen-Chih
  Peng.
\newblock Attractive or {Faithful}? {Popularity}-{Reinforced} {Learning} for
  {Inspired} {Headline} {Generation}.
\newblock \emph{arXiv:2002.02095 [cs]}, February 2020{\natexlab{d}}.
\newblock URL \url{http://arxiv.org/abs/2002.02095}.
\newblock arXiv: 2002.02095.

\bibitem[Tang et~al.(2019)Tang, Zhao, Xiong, Liang, Xing, and
  Hu]{tang2019target}
Jianheng Tang, Tiancheng Zhao, Chenyan Xiong, Xiaodan Liang, Eric~P Xing, and
  Zhiting Hu.
\newblock Target-guided open-domain conversation.
\newblock \emph{arXiv preprint arXiv:1905.11553}, 2019.

\bibitem[Tang et~al.(2021)Tang, Fabbri, Mao, Adams, Wang, Li, Mehdad, and
  Radev]{tang2021investigating}
Xiangru Tang, Alexander~R Fabbri, Ziming Mao, Griffin Adams, Borui Wang, Haoran
  Li, Yashar Mehdad, and Dragomir Radev.
\newblock Investigating crowdsourcing protocols for evaluating the factual
  consistency of summaries.
\newblock \emph{arXiv preprint arXiv:2109.09195}, 2021.

\bibitem[Tian et~al.(2019)Tian, Narayan, Sellam, and
  Parikh]{tian2019StickingFacts}
Ran Tian, Shashi Narayan, Thibault Sellam, and Ankur~P. Parikh.
\newblock Sticking to the {{Facts}}: Confident {{Decoding}} for {{Faithful
  Data}}-to-{{Text Generation}}.
\newblock October 2019.

\bibitem[Tu et~al.(2016)Tu, Lu, Liu, Liu, and Li]{tu2016modeling}
Zhaopeng Tu, Zhengdong Lu, Yang Liu, Xiaohua Liu, and Hang Li.
\newblock Modeling coverage for neural machine translation.
\newblock \emph{arXiv preprint arXiv:1601.04811}, 2016.

\bibitem[Tu et~al.(2017)Tu, Liu, Shang, Liu, and Li]{tu2017neural}
Zhaopeng Tu, Yang Liu, Lifeng Shang, Xiaohua Liu, and Hang Li.
\newblock Neural machine translation with reconstruction.
\newblock In \emph{Thirty-First AAAI Conference on Artificial Intelligence},
  2017.

\bibitem[Vaswani et~al.(2017)Vaswani, Shazeer, Parmar, Uszkoreit, Jones, Gomez,
  Kaiser, and Polosukhin]{vaswani2017attention}
Ashish Vaswani, Noam Shazeer, Niki Parmar, Jakob Uszkoreit, Llion Jones,
  Aidan~N Gomez, {\L}ukasz Kaiser, and Illia Polosukhin.
\newblock Attention is all you need.
\newblock \emph{Advances in neural information processing systems}, 30, 2017.

\bibitem[Veli{\v{c}}kovi{\'c} et~al.(2017)Veli{\v{c}}kovi{\'c}, Cucurull,
  Casanova, Romero, Lio, and Bengio]{velivckovic2017graph}
Petar Veli{\v{c}}kovi{\'c}, Guillem Cucurull, Arantxa Casanova, Adriana Romero,
  Pietro Lio, and Yoshua Bengio.
\newblock Graph attention networks.
\newblock \emph{arXiv preprint arXiv:1710.10903}, 2017.

\bibitem[Vinyals and Le(2015)]{DBLP:journals/corr/VinyalsL15}
Oriol Vinyals and Quoc~V. Le.
\newblock A neural conversational model.
\newblock \emph{CoRR}, abs/1506.05869, 2015.
\newblock URL \url{http://arxiv.org/abs/1506.05869}.

\bibitem[Vinyals et~al.(2015)Vinyals, Toshev, Bengio, and
  Erhan]{vinyals2015show}
Oriol Vinyals, Alexander Toshev, Samy Bengio, and Dumitru Erhan.
\newblock Show and tell: A neural image caption generator.
\newblock In \emph{Proceedings of the IEEE conference on computer vision and
  pattern recognition}, pages 3156--3164, 2015.

\bibitem[Wang et~al.(2020{\natexlab{a}})Wang, Cho, and
  Lewis]{DBLP:conf/acl/WangCL20}
Alex Wang, Kyunghyun Cho, and Mike Lewis.
\newblock Asking and answering questions to evaluate the factual consistency of
  summaries.
\newblock In Dan Jurafsky, Joyce Chai, Natalie Schluter, and Joel~R. Tetreault,
  editors, \emph{Proceedings of the 58th Annual Meeting of the Association for
  Computational Linguistics, {ACL} 2020, Online, July 5-10, 2020}, pages
  5008--5020. Association for Computational Linguistics, 2020{\natexlab{a}}.
\newblock \doi{10.18653/v1/2020.acl-main.450}.
\newblock URL \url{https://doi.org/10.18653/v1/2020.acl-main.450}.

\bibitem[Wang and Sennrich(2020)]{wang_exposure_2020}
Chaojun Wang and Rico Sennrich.
\newblock On {Exposure} {Bias}, {Hallucination} and {Domain} {Shift} in
  {Neural} {Machine} {Translation}.
\newblock In \emph{Proceedings of the 58th {Annual} {Meeting} of the
  {Association} for {Computational} {Linguistics}}, pages 3544--3552, Online,
  July 2020. Association for Computational Linguistics.
\newblock \doi{10.18653/v1/2020.acl-main.326}.
\newblock URL \url{https://aclanthology.org/2020.acl-main.326}.

\bibitem[Wang(2019)]{wang2019RevisitingChallenges}
Hongmin Wang.
\newblock Revisiting {{Challenges}} in {{Data}}-to-{{Text Generation}} with
  {{Fact Grounding}}.
\newblock In \emph{Proceedings of the 12th {{International Conference}} on
  {{Natural Language Generation}}}, pages 311--322, {Tokyo, Japan}, October
  2019. {Association for Computational Linguistics}.
\newblock \doi{10.18653/v1/W19-8639}.

\bibitem[Wang et~al.(2021)Wang, Lin, Yang, Zhou, Zhang, Zhou, and
  Yang]{wang2021SketchRefine}
Peng Wang, Junyang Lin, An~Yang, Chang Zhou, Yichang Zhang, Jingren Zhou, and
  Hongxia Yang.
\newblock Sketch and {{Refine}}: Towards {{Faithful}} and {{Informative
  Table}}-to-{{Text Generation}}.
\newblock In \emph{Findings of the {{Association}} for {{Computational
  Linguistics}}: {{ACL}}-{{IJCNLP}} 2021}, pages 4831--4843, {Online}, 2021.
  {Association for Computational Linguistics}.
\newblock \doi{10.18653/v1/2021.findings-acl.427}.

\bibitem[Wang et~al.(2018)Wang, Pan, Huang, Zhang, Jiang, Ji, and
  Knight]{wang2018DescribingKnowledge}
Qingyun Wang, Xiaoman Pan, Lifu Huang, Boliang Zhang, Zhiying Jiang, Heng Ji,
  and Kevin Knight.
\newblock Describing a {{Knowledge Base}}.
\newblock In \emph{Proceedings of the 11th {{International Conference}} on
  {{Natural Language Generation}}}, pages 10--21, {Tilburg University, The
  Netherlands}, 2018. {Association for Computational Linguistics}.
\newblock \doi{10.18653/v1/W18-6502}.

\bibitem[Wang et~al.(2020{\natexlab{b}})Wang, Wang, An, Yu, and
  Chen]{wang2020FaithfulNeural}
Zhenyi Wang, Xiaoyang Wang, Bang An, Dong Yu, and Changyou Chen.
\newblock Towards {{Faithful Neural Table}}-to-{{Text Generation}} with
  {{Content}}-{{Matching Constraints}}.
\newblock In \emph{Proceedings of the 58th {{Annual Meeting}} of the
  {{Association}} for {{Computational Linguistics}}}, pages 1072--1086,
  {Online}, 2020{\natexlab{b}}. {Association for Computational Linguistics}.
\newblock \doi{10.18653/v1/2020.acl-main.101}.

\bibitem[Welleck et~al.(2019{\natexlab{a}})Welleck, Brantley, Iii, and
  Cho]{welleck2019non}
Sean Welleck, Kiant{\'e} Brantley, Hal~Daum{\'e} Iii, and Kyunghyun Cho.
\newblock Non-monotonic sequential text generation.
\newblock In \emph{International Conference on Machine Learning}, pages
  6716--6726. PMLR, 2019{\natexlab{a}}.

\bibitem[Welleck et~al.(2019{\natexlab{b}})Welleck, Weston, Szlam, and
  Cho]{DBLP:conf/acl/WelleckWSC19}
Sean Welleck, Jason Weston, Arthur Szlam, and Kyunghyun Cho.
\newblock Dialogue natural language inference.
\newblock In Anna Korhonen, David~R. Traum, and Llu{\'{\i}}s M{\`{a}}rquez,
  editors, \emph{Proceedings of the 57th Conference of the Association for
  Computational Linguistics, {ACL} 2019, Florence, Italy, July 28- August 2,
  2019, Volume 1: Long Papers}, pages 3731--3741. Association for Computational
  Linguistics, 2019{\natexlab{b}}.
\newblock \doi{10.18653/v1/p19-1363}.
\newblock URL \url{https://doi.org/10.18653/v1/p19-1363}.

\bibitem[Welleck et~al.(2019{\natexlab{c}})Welleck, Weston, Szlam, and
  Cho]{welleck_dialogue_2019}
Sean Welleck, Jason Weston, Arthur Szlam, and Kyunghyun Cho.
\newblock Dialogue {Natural} {Language} {Inference}.
\newblock In \emph{Proceedings of the 57th {Annual} {Meeting} of the
  {Association} for {Computational} {Linguistics}}, pages 3731--3741, Florence,
  Italy, July 2019{\natexlab{c}}. Association for Computational Linguistics.
\newblock \doi{10.18653/v1/P19-1363}.
\newblock URL \url{https://aclanthology.org/P19-1363}.

\bibitem[Weng et~al.(2017)Weng, Huang, Zheng, Dai, and Chen]{weng2017neural}
Rongxiang Weng, Shujian Huang, Zaixiang Zheng, Xinyu Dai, and Jiajun Chen.
\newblock Neural machine translation with word predictions.
\newblock \emph{arXiv preprint arXiv:1708.01771}, 2017.

\bibitem[Weng et~al.(2020{\natexlab{a}})Weng, Wei, Huang, Yu, Bing, Luo, and
  Chen]{weng2020gret}
Rongxiang Weng, Haoran Wei, Shujian Huang, Heng Yu, Lidong Bing, Weihua Luo,
  and Jiajun Chen.
\newblock Gret: Global representation enhanced transformer.
\newblock In \emph{Proceedings of the AAAI Conference on Artificial
  Intelligence}, volume~34, pages 9258--9265, 2020{\natexlab{a}}.

\bibitem[Weng et~al.(2020{\natexlab{b}})Weng, Yu, Wei, and
  Luo]{weng_towards_2020}
Rongxiang Weng, Heng Yu, Xiangpeng Wei, and Weihua Luo.
\newblock Towards {Enhancing} {Faithfulness} for {Neural} {Machine}
  {Translation}.
\newblock In \emph{Proceedings of the 2020 {Conference} on {Empirical}
  {Methods} in {Natural} {Language} {Processing} ({EMNLP})}, pages 2675--2684,
  Online, November 2020{\natexlab{b}}. Association for Computational
  Linguistics.
\newblock \doi{10.18653/v1/2020.emnlp-main.212}.
\newblock URL \url{https://www.aclweb.org/anthology/2020.emnlp-main.212}.

\bibitem[Williams et~al.(2018)Williams, Nangia, and
  Bowman]{DBLP:conf/naacl/WilliamsNB18}
Adina Williams, Nikita Nangia, and Samuel~R. Bowman.
\newblock A broad-coverage challenge corpus for sentence understanding through
  inference.
\newblock In Marilyn~A. Walker, Heng Ji, and Amanda Stent, editors,
  \emph{Proceedings of the 2018 Conference of the North American Chapter of the
  Association for Computational Linguistics: Human Language Technologies,
  {NAACL-HLT} 2018, New Orleans, Louisiana, USA, June 1-6, 2018, Volume 1 (Long
  Papers)}, pages 1112--1122. Association for Computational Linguistics, 2018.
\newblock \doi{10.18653/v1/n18-1101}.
\newblock URL \url{https://doi.org/10.18653/v1/n18-1101}.

\bibitem[Williams and Zipser(1989)]{williams1989learning}
Ronald~J Williams and David Zipser.
\newblock A learning algorithm for continually running fully recurrent neural
  networks.
\newblock \emph{Neural computation}, 1\penalty0 (2):\penalty0 270--280, 1989.

\bibitem[Wiseman et~al.(2017)Wiseman, Shieber, and
  Rush]{wiseman2017ChallengesDatatoDocument}
Sam Wiseman, Stuart Shieber, and Alexander Rush.
\newblock Challenges in {{Data}}-to-{{Document Generation}}.
\newblock In \emph{Proceedings of the 2017 {{Conference}} on {{Empirical
  Methods}} in {{Natural Language Processing}}}, pages 2253--2263, {Copenhagen,
  Denmark}, 2017. {Association for Computational Linguistics}.
\newblock \doi{10.18653/v1/D17-1239}.

\bibitem[Wu et~al.(2021{\natexlab{a}})Wu, Li, Xiao, Liu, Cao, Li, Wu, and
  Wang]{wu2021bass}
Wenhao Wu, Wei Li, Xinyan Xiao, Jiachen Liu, Ziqiang Cao, Sujian Li, Hua Wu,
  and Haifeng Wang.
\newblock Bass: Boosting abstractive summarization with unified semantic graph.
\newblock In \emph{Proceedings of the 59th Annual Meeting of the Association
  for Computational Linguistics and the 11th International Joint Conference on
  Natural Language Processing (Volume 1: Long Papers)}, pages 6052--6067,
  2021{\natexlab{a}}.

\bibitem[Wu et~al.(2021{\natexlab{b}})Wu, Galley, Brockett, Zhang, Gao, Quirk,
  Koncel{-}Kedziorski, Gao, Hajishirzi, Ostendorf, and
  Dolan]{DBLP:conf/aaai/WuGBZ0QKGHOD21}
Zeqiu Wu, Michel Galley, Chris Brockett, Yizhe Zhang, Xiang Gao, Chris Quirk,
  Rik Koncel{-}Kedziorski, Jianfeng Gao, Hannaneh Hajishirzi, Mari Ostendorf,
  and Bill Dolan.
\newblock A controllable model of grounded response generation.
\newblock In \emph{Thirty-Fifth {AAAI} Conference on Artificial Intelligence,
  {AAAI} 2021, Thirty-Third Conference on Innovative Applications of Artificial
  Intelligence, {IAAI} 2021, The Eleventh Symposium on Educational Advances in
  Artificial Intelligence, {EAAI} 2021, Virtual Event, February 2-9, 2021},
  pages 14085--14093. {AAAI} Press, 2021{\natexlab{b}}.
\newblock URL \url{https://ojs.aaai.org/index.php/AAAI/article/view/17658}.

\bibitem[Xie et~al.(2021)Xie, Sun, Deng, Li, and Ding]{xie_factual_2021}
Yuexiang Xie, Fei Sun, Yang Deng, Yaliang Li, and Bolin Ding.
\newblock Factual {Consistency} {Evaluation} for {Text} {Summarization} via
  {Counterfactual} {Estimation}.
\newblock \emph{arXiv:2108.13134 [cs]}, September 2021.
\newblock URL \url{http://arxiv.org/abs/2108.13134}.
\newblock arXiv: 2108.13134.

\bibitem[Yang et~al.(2019)Yang, Cheng, Liu, and Sun]{yang2019reducing}
Zonghan Yang, Yong Cheng, Yang Liu, and Maosong Sun.
\newblock Reducing word omission errors in neural machine translation: A
  contrastive learning approach.
\newblock In \emph{Proceedings of the 57th Annual Meeting of the Association
  for Computational Linguistics}, pages 6191--6196, 2019.

\bibitem[Yates et~al.(2007)Yates, Banko, Broadhead, Cafarella, Etzioni, and
  Soderland]{yates2007textrunner}
Alexander Yates, Michele Banko, Matthew Broadhead, Michael~J Cafarella, Oren
  Etzioni, and Stephen Soderland.
\newblock Textrunner: open information extraction on the web.
\newblock In \emph{Proceedings of Human Language Technologies: The Annual
  Conference of the North American Chapter of the Association for Computational
  Linguistics (NAACL-HLT)}, pages 25--26, 2007.

\bibitem[Yuan et~al.(2021)Yuan, Neubig, and Liu]{yuan2021bartscore}
Weizhe Yuan, Graham Neubig, and Pengfei Liu.
\newblock Bartscore: Evaluating generated text as text generation.
\newblock \emph{Advances in Neural Information Processing Systems}, 34, 2021.

\bibitem[Zaremba et~al.(2014)Zaremba, Sutskever, and
  Vinyals]{zaremba2014recurrent}
Wojciech Zaremba, Ilya Sutskever, and Oriol Vinyals.
\newblock Recurrent neural network regularization.
\newblock \emph{arXiv preprint arXiv:1409.2329}, 2014.

\bibitem[Zellers et~al.(2020)Zellers, Holtzman, Rashkin, Bisk, Farhadi,
  Roesner, and Choi]{zellers_defending_2020}
Rowan Zellers, Ari Holtzman, Hannah Rashkin, Yonatan Bisk, Ali Farhadi,
  Franziska Roesner, and Yejin Choi.
\newblock Defending {Against} {Neural} {Fake} {News}.
\newblock \emph{arXiv:1905.12616 [cs]}, December 2020.
\newblock URL \url{http://arxiv.org/abs/1905.12616}.
\newblock arXiv: 1905.12616.

\bibitem[Zhang et~al.(2020{\natexlab{a}})Zhang, Nagesh, and
  Knight]{zhang2020parallel}
Boliang Zhang, Ajay Nagesh, and Kevin Knight.
\newblock Parallel corpus filtering via pre-trained language models.
\newblock \emph{arXiv preprint arXiv:2005.06166}, 2020{\natexlab{a}}.

\bibitem[Zhang et~al.(2021)Zhang, Luan, Sun, Zhai, Xu, and
  Liu]{zhang2021neural}
Jiacheng Zhang, Huanbo Luan, Maosong Sun, Feifei Zhai, Jingfang Xu, and Yang
  Liu.
\newblock Neural machine translation with explicit phrase alignment.
\newblock \emph{IEEE/ACM Transactions on Audio, Speech, and Language
  Processing}, 29:\penalty0 1001--1010, 2021.

\bibitem[Zhang et~al.(2020{\natexlab{b}})Zhang, Zhao, Saleh, and
  Liu]{zhang2020pegasus}
Jingqing Zhang, Yao Zhao, Mohammad Saleh, and Peter Liu.
\newblock Pegasus: Pre-training with extracted gap-sentences for abstractive
  summarization.
\newblock In \emph{International Conference on Machine Learning}, pages
  11328--11339. PMLR, 2020{\natexlab{b}}.

\bibitem[Zhang et~al.(2018)Zhang, Dinan, Urbanek, Szlam, Kiela, and
  Weston]{DBLP:conf/acl/KielaWZDUS18}
Saizheng Zhang, Emily Dinan, Jack Urbanek, Arthur Szlam, Douwe Kiela, and Jason
  Weston.
\newblock Personalizing dialogue agents: {I} have a dog, do you have pets too?
\newblock In Iryna Gurevych and Yusuke Miyao, editors, \emph{Proceedings of the
  56th Annual Meeting of the Association for Computational Linguistics, {ACL}
  2018, Melbourne, Australia, July 15-20, 2018, Volume 1: Long Papers}, pages
  2204--2213. Association for Computational Linguistics, 2018.
\newblock \doi{10.18653/v1/P18-1205}.
\newblock URL \url{https://aclanthology.org/P18-1205/}.

\bibitem[Zhang and Ghorbani(2020)]{zhang2020overview}
Xichen Zhang and Ali~A Ghorbani.
\newblock An overview of online fake news: Characterization, detection, and
  discussion.
\newblock \emph{Information Processing \& Management}, 57\penalty0
  (2):\penalty0 102025, 2020.

\bibitem[Zhang et~al.(2019)Zhang, Sun, Galley, Chen, Brockett, Gao, Gao, Liu,
  and Dolan]{zhang2019dialogpt}
Yizhe Zhang, Siqi Sun, Michel Galley, Yen-Chun Chen, Chris Brockett, Xiang Gao,
  Jianfeng Gao, Jingjing Liu, and Bill Dolan.
\newblock Dialogpt: Large-scale generative pre-training for conversational
  response generation.
\newblock \emph{arXiv preprint arXiv:1911.00536}, 2019.

\bibitem[Zhang et~al.(2020{\natexlab{c}})Zhang, Wang, Li, Gan, Brockett, and
  Dolan]{zhang_pointer_2020}
Yizhe Zhang, Guoyin Wang, Chunyuan Li, Zhe Gan, Chris Brockett, and Bill Dolan.
\newblock {POINTER}: {Constrained} {Progressive} {Text} {Generation} via
  {Insertion}-based {Generative} {Pre}-training.
\newblock \emph{arXiv:2005.00558 [cs]}, September 2020{\natexlab{c}}.
\newblock URL \url{http://arxiv.org/abs/2005.00558}.
\newblock arXiv: 2005.00558.

\bibitem[Zhang et~al.(2020{\natexlab{d}})Zhang, Merck, Tsai, Manning, and
  Langlotz]{zhang_optimizing_2020}
Yuhao Zhang, Derek Merck, Emily~Bao Tsai, Christopher~D. Manning, and Curtis~P.
  Langlotz.
\newblock Optimizing the {Factual} {Correctness} of a {Summary}: {A} {Study} of
  {Summarizing} {Radiology} {Reports}.
\newblock \emph{arXiv:1911.02541 [cs]}, April 2020{\natexlab{d}}.
\newblock URL \url{http://arxiv.org/abs/1911.02541}.
\newblock arXiv: 1911.02541.

\bibitem[Zhao et~al.(2020{\natexlab{a}})Zhao, Cohen, and
  Webber]{zhao2020reducing}
Zheng Zhao, Shay~B Cohen, and Bonnie Webber.
\newblock Reducing quantity hallucinations in abstractive summarization.
\newblock In \emph{Findings of the Association for Computational Linguistics:
  EMNLP 2020}, pages 2237--2249, 2020{\natexlab{a}}.

\bibitem[Zhao et~al.(2020{\natexlab{b}})Zhao, Cohen, and
  Webber]{zhao_reducing_2020}
Zheng Zhao, Shay~B. Cohen, and Bonnie Webber.
\newblock Reducing {Quantity} {Hallucinations} in {Abstractive}
  {Summarization}.
\newblock In \emph{Findings of the {Association} for {Computational}
  {Linguistics}: {EMNLP} 2020}, pages 2237--2249, Online, 2020{\natexlab{b}}.
  Association for Computational Linguistics.
\newblock \doi{10.18653/v1/2020.findings-emnlp.203}.
\newblock URL \url{https://aclanthology.org/2020.findings-emnlp.203}.

\bibitem[Zheng et~al.(2019)Zheng, Huang, Tu, Dai, and Chen]{zheng2019dynamic}
Zaixiang Zheng, Shujian Huang, Zhaopeng Tu, Xin-Yu Dai, and Jiajun Chen.
\newblock Dynamic past and future for neural machine translation.
\newblock \emph{arXiv preprint arXiv:1904.09646}, 2019.

\bibitem[Zhong et~al.(2020)Zhong, Tang, Xu, Wang, Duan, Zhou, Wang, and
  Yin]{zhong2020neural}
Wanjun Zhong, Duyu Tang, Zenan Xu, Ruize Wang, Nan Duan, Ming Zhou, Jiahai
  Wang, and Jian Yin.
\newblock Neural deepfake detection with factual structure of text.
\newblock \emph{arXiv preprint arXiv:2010.07475}, 2020.

\bibitem[Zhou et~al.(2021)Zhou, Neubig, Gu, Diab, Guzmán, Zettlemoyer, and
  Ghazvininejad]{zhou_detecting_2021}
Chunting Zhou, Graham Neubig, Jiatao Gu, Mona Diab, Francisco Guzmán, Luke
  Zettlemoyer, and Marjan Ghazvininejad.
\newblock Detecting {Hallucinated} {Content} in {Conditional} {Neural}
  {Sequence} {Generation}.
\newblock In \emph{Findings of the {Association} for {Computational}
  {Linguistics}: {ACL}-{IJCNLP} 2021}, pages 1393--1404, Online, August 2021.
  Association for Computational Linguistics.
\newblock \doi{10.18653/v1/2021.findings-acl.120}.
\newblock URL \url{https://aclanthology.org/2021.findings-acl.120}.

\end{thebibliography}
\end{document}